\documentclass[10pt,journal,compsoc]{IEEEtran}
\usepackage{tikz}
\usetikzlibrary{shapes, arrows.meta, positioning}

\ifCLASSOPTIONcompsoc
  \usepackage[nocompress]{cite}
\else
  \usepackage{cite}
\fi

\ifCLASSINFOpdf
\else
\fi
\usepackage{amsmath}

\usepackage{array}
\usepackage{url}

\usepackage{soul}
\usepackage[hidelinks]{hyperref}
\usepackage[utf8]{inputenc}
\usepackage{graphicx}
\usepackage{amsthm}
\usepackage{booktabs}

\usepackage{amsfonts}
\usepackage{xcolor}
\usepackage{multirow}
\usepackage{makecell}
\usepackage{amssymb}
\usepackage{fontawesome}

\usepackage{subfigure}

\usepackage{tcolorbox}
\tcbuselibrary{skins, breakable, theorems}

\newtheorem{theorem}{Theorem}
\newtheorem{definition}{Definition}
\newtheorem{assumption}{Assumption}

\DeclareMathOperator*{\argmax}{arg\,max}
\DeclareMathOperator*{\argmin}{arg\,min}

\newcommand{\etal}{\emph{et~al.}}
\newcommand{\eg}{\emph{e.g.},\/~}

\newcommand{\ie}{\emph{i.e.},\/~}
\newcommand{\wrt}{\emph{w.r.t.}\/~}

\newcommand\figref[1]{Fig.~\ref{#1}}

\newcommand\tabref[1]{Table~\ref{#1}}
\newcommand\secref[1]{Section~\ref{#1}}
\newcommand\equref[1]{Equation~(\ref{#1})}
\newcommand\defref[1]{Definition~\ref{#1}}
\newcommand\assref[1]{Assumption~\ref{#1}}
\newcommand\appref[1]{Appendix~\ref{#1}}

\newcommand{\eat}[1]{}
\newcommand{\hao}[1]{{\color{blue}Hao: {#1}}}
\newcommand{\wz}[1]{{\color[rgb]{0.35,0.60,0.25}wz: {#1}}}
\newcommand\beftext[1]{{\color[rgb]{0.5,0.5,0.5}{BEFORE:#1}}}
\newcommand{\TODO}[1]{{\color{red}TODO: {#1}}}
\newcommand{\rev}[1]{{\color{purple}#1}} 

\begin{document}
%

\title{When Graph Neural Network Meets Causality: Opportunities, Methodologies and An Outlook}

\author{Wenzhao~Jiang,~\IEEEmembership{}
        Hao~Liu,~\IEEEmembership{Senior Member,~IEEE}
        and~Hui~Xiong,~\IEEEmembership{Fellow,~IEEE}
\IEEEcompsocitemizethanks{\IEEEcompsocthanksitem 
Wenzhao Jiang is with the Artificial Intelligence Thrust, The Hong Kong University of Science and Technology (Guangzhou), Guangzhou, PRC. E-mail: wjiang431@connect.hkust-gz.edu.cn
\IEEEcompsocthanksitem Hao Liu (corresponding author) and Hui Xiong are with the Artificial Intelligence Thrust, The Hong Kong University of Science and Technology (Guangzhou), Guangzhou, PRC and the Department of Computer Science and Engineering, The Hong Kong University of Science and Technology, Hong Kong SAR, PRC. E-mail: \{liuh,xionghui\}@ust.hk
}

}

\markboth{Journal of \LaTeX\ Class Files,~Vol.~14, No.~8, August~2015}%
{Shell \MakeLowercase{\textit{et al.}}: When Graph Neural Network Meets Causality: Trustworthiness, Progress and Outlook}

\IEEEtitleabstractindextext{%
\begin{abstract}
    Graph Neural Networks (GNNs) have emerged as powerful representation learning tools for capturing complex dependencies within diverse graph-structured data. Despite their success in a wide range of graph mining tasks, GNNs have raised serious concerns regarding their trustworthiness, including susceptibility to distribution shift, biases towards certain populations, and lack of explainability.
    Recently, integrating causal learning techniques into GNNs has sparked numerous ground-breaking studies since many GNN trustworthiness issues can be alleviated by capturing the underlying data causality rather than superficial correlations. 
    In this survey, we comprehensively review recent research efforts on Causality-Inspired GNNs (CIGNNs). 
    Specifically, we first employ causal tools to analyze the primary trustworthiness risks of existing GNNs, underscoring the necessity for GNNs to comprehend the causal mechanisms within graph data.
    Moreover, we introduce a taxonomy of CIGNNs based on the type of causal learning capability they are equipped with, \ie{causal reasoning and causal representation learning.}
    Besides, we systematically introduce typical methods within each category and discuss how they mitigate trustworthiness risks.
    Finally, we summarize useful resources and discuss several future directions, hoping to shed light on new research opportunities in this emerging field. The representative papers, along with open-source data and codes, are available in \url{https://github.com/usail-hkust/Causality-Inspired-GNNs}.
\end{abstract}

}

\maketitle

\IEEEdisplaynontitleabstractindextext

%
\IEEEpeerreviewmaketitle

\section{Introduction}
\IEEEPARstart{G}{raph-structured} data is prevalent in real-world domains, including social networks, traffic networks, and molecular networks. 
\eat{
hence various Graph Representation Learning (GRL) methods have been developed over the years to handle such difficulties~\cite{gnnsurvey2021tnnls,gnnbook2022springer}. Among them, Graph Neural Networks (GNNs) achieve SOTA performance and have been applied to various graph mining applications, 
}
Traditional deep learning models designed for Euclidean data often fall short when it comes to modeling non-tabular graph data.
As a result, numerous Graph Neural Networks (GNNs) have been proposed over the years~\cite{gnnsurvey2021tnnls}, achieving state-of-the-art performance across various graph mining applications.
In short, GNNs map input graphs into a set of node representations, which are iteratively updated based on information from neighboring nodes. These updates rely on functions trained directly under the supervision of the graph itself or downstream tasks. 
By capturing both local feature and global graph structural information, GNNs preserve abundant knowledge in low-dimensional representations, greatly benefiting a wide range of downstream applications, \eg bioinformatics~\cite{gnn4bio2022}, recommender systems~\cite{gnn4rs2023acm}, knowledge representation~\cite{gnn4kg2017tkde,uukg2023nips}, talent analysis~\cite{talent2022kdd} and urban computing~\cite{mugrep2021kdd,semiparking2022tkde,airpred2023tkde}.


As GNNs continue to gain traction, concerns have emerged regarding their trustworthiness~\cite{TrustGNNsurvey2022,TwGL2022,tgnn2022peijian}. 
\eat{particularly in terms of Out-Of-Distribution (OOD) generalizability, fairness, and explainability.
This survey focuses on these three key aspects, highlighting their significance in establishing trustworthy GNNs.}
First, many GNNs exhibit poor \emph{Out-Of-Distribution (OOD) generalizability} due to their susceptibility to distribution shifts between training and testing graphs. In particular, distribution shifts on graphs can occur at both the attribute level (\eg node features) and the topology level (\eg node degree), posing additional challenges for OOD generalization~\cite{oodgraphsurvey2022,ASTFA2022nips}.
Second, GNNs are prone to generating \emph{unfair} representations, resulting in biased outcomes towards certain sample groups~\cite{fairness4gnn2022survey,gear2022wsdm}. 
Third, the poor \emph{explainability} induced by the black-box nature of information propagation in GNNs raises concerns about their reliability. It also hinders developers from diagnosing and addressing the model's performance shortcomings~\cite{xgnnsurvey2022tpami,gnnexplainer2019nips}.
\eat{\TODO{Actually, it is wired in a survey that the study on casual based explainable GNN is missing/rare. Like point 1\&2, you may emphasize the importance of explainability. All in all, you haven't yet mentioned casual here.}
\wz{Agree. I deleted the sentence mentioning causal, and further added an example to emphasize the importance of explainability.
}}
Such issues undermine the trustworthiness of GNNs and become particularly concerning when applying GNNs to high-stakes applications, such as fraud detection~\cite{gnn4anomaly2021} and criminal justice~\cite{nifty2021uai}. 
It is essential to address the aforementioned challenges to broaden GNNs' application spectrum.

Causal learning, an established field focusing on uncovering causal mechanisms from observational data~\cite{causality2009pearl,causal_data_survey2020,stat2causal2022ICM}, has recently been identified as a pivotal tool for creating trustworthy AI~\cite{SRAIsurvey2021ijcai,causal_tai_survey2023}.
Traditional AI systems primarily operate at the first level of the 'ladder of causality'~\cite{seventools2019pearl}, which involves recognizing patterns and correlations in data. By incorporating causal learning, these systems advance beyond mere \emph{association} to understand the causal effects of potential \emph{interventions} and, at the highest rung, hypothesize about unseen \emph{counterfactuals}.
By climbing this ladder, AI systems gain a deeper understanding of the inherent data generation process, which is crucial for making reliable predictions and responsible decisions and earning human trust in variational and unpredictable environments.
As a result, causal learning is attracting growing interest among graph learning researchers for constructing Trustworthy Graph Neural Networks~(TGNNs).
These Causality-Inspired GNNs~(CIGNNs) are envisioned to operate accurately and reliably in vital applications based on an in-depth comprehension of the causality inherent in graph-structured data. 
For example, 
Causal learning can empower GNNs to generalize user behavior predictions across different social media platforms by learning underlying data generation processes, rather than superficial correlations~\cite{EERM2022iclr}.
GNNs aware of interventional causal effects can mitigate unfairness in recommendation systems by discerning and disregarding spurious correlations between sensitive attributes and user preferences~\cite{causalrecsurvey2022}.
Understanding causal mechanisms leading to loan default risks contributes to more interpretable justifications for loan decisions~\cite{inl2023tois}.

Though potentially promising, developing CIGNNs presents three primary challenges.
\eat{\TODO{It is wired saying still exists massive challenges in a survey paper, which let the audience feel existing studies did nothing. Instead, you may want to say existing efforts focus on addressing what kind of challenges. Pay attention if the summarized challenges are well aligned with the whole paper. More importantly, what do you want to express by talking about the challenges in this direction? A survey can address these challenges?}
\wz{Indeed, I actually want to list special challenges on causal4GNN that have been explored in existing literature. I have revised the first sentence.

By further pointing out challenges, I hope to let audience feel this direction is potentially promising while still challenging. Then later in section 4, we can show how existing works attempt to solve these challenges.

As for the alignment, challenge 1,2 are aligned with section 4, and challenge 3 is aligned with section 5-7. Although challenge 1,2 seem not rigorously match the taxonomy in section 4, they correspond to the main contributions of each introduced method. For each method, it should first give a solution to challenge 1, and then figure out challenge 2.}
}
First, the high-dimensional and non-Euclidean nature of graph-structured data make the causal relations among graph components (e.g., nodes, edges, or subgraphs) extremely complex. It is challenging to specify causal variables of interest from graph data, 
clarify causal relations among these variables conditioning on certain domain knowledge, 
\eat{\hao{what do you mean by ``if there is prior knowledge''?}
\wz{Rephrased it. It means domain knowledge, for example, domain knowledge might help clarify the causal relations between different parts of a molecule, or between a typical structure and a property of molecule.}
\hao{I don't get it. Why emphasise domain knowledge? What if no domain knowledge?}}
and choose suitable causal learning methods to obtain causal knowledge for improving trustworthiness in downstream applications~\cite{debiasedgnn2022tnnls,gear2022wsdm,gem2021icml}.
Second, incorporating causal knowledge into GNNs poses challenges on redesigning GNN architectures and training algorithms to accommodate causal relations
\cite{DIR2022iclr,nifty2021uai,orphicx2022cvpr}.
Third, evaluating CIGNNs is challenging as the data causality may vary across applications or be inaccessible, which necessitates tailored evaluation benchmarks and metrics~\cite{good2022nips,gear2022wsdm,gce_survey2022}.
Rapid progress in CIGNNs has provided valuable insights towards tackling the above challenges. 
However, certain issues in this field may impede further progress.
First, there is a lack of in-depth comparison of the pros and cons of causal techniques leveraged to improve a specific GNN trustworthiness risk. Besides, there is a notable absence of analysis on the similarities and differences in the application practices of causal techniques across varied trustworthiness risks.
A comprehensive review is required to distil the fundamental principles of existing efforts in developing and evaluating CIGNNs and unleashing their full potential.
\eat{To this end, we present a comprehensive survey of recent advancements in CIGNNs within a unified taxonomy, offering insights into their commonalities, advantages and potential impact on the field of graph learning. To the best of our knowledge, this is the first attempt to systematically review the existing CIGNNs.}
To this end, we commence the first effort to systematically survey recent advancements in CIGNNs within a unified taxonomy, offering insights into their commonalities, advantages and potential impact on the field of graph learning.
The main contributions of this survey are detailed below.
\begin{itemize}
    \eat{\item To the best of our knowledge, this is the first attempt to systematically survey the existing CIGNNs. \TODO{One question is, if CIGNNs naturally trustworthy? if so, CIGNNs is enough, no need T.}
    \wz{Indeed, CIGNNs are naturally trustworthy. I have unified all places. }}
    \item We analyze the rationale behind different trustworthiness risks of GNNs through the lens of causality, underscoring the importance of gaining a deeper understanding of the inherent causal mechanisms in graph data. From a causal view, we provide valuable insights into the development of generalizable, fair and interpretable graph learning solutions.
    \item 
    \eat{To facilitate a better understanding of the commonalities and specialities of various CIGNNs, we propose a novel taxonomy based on the type of causal learning task they are enabled to handle, including causal reasoning and causal representation learning. Within each category, we delve into representative methods and their specific contributions towards improving trustworthiness, offering a comprehensive overview of the state-of-the-art in this emerging field.}
    We innovatively categorize existing CIGNNs by identifying their essential causal learning capability. In each category, we delve into representative methodologies and highlight their impacts on GNN trustworthiness. This novel taxonomy facilitates an integrated understanding of the intricate links between causal learning and GNN trustworthiness, out of numerous seemingly isolated studies.
    \item We systematically compile an overview of open-source benchmarks, data synthesis strategies, commonly employed evaluation metrics, as well as available open-source codes and packages. This compilation aims to enable easier exploration of causality-inspired ideas in developing TGNNs and encourage their practical implementation in various downstream applications.
    \item We discuss several future directions to motivate the development of this promising field.
\end{itemize}

\noindent 
\textbf{Connections to Existing Surveys.}
There are a few surveys paying attention to the combination of causality and graph data. Specifically, Ma~\etal~\cite{graph4causal2022aimagazine} reviewed existing works of causal reasoning on graph-structured data. Job~\etal~\cite{explorecausalgnn2023} surveyed emerging GNNs methods for solving different causal learning tasks.
Nevertheless, both of them failed to systematically discuss how causal learning benefits the development of more trustworthy GNNs, which, as discussed above, has the potential to advance the practicality and reliability of graph learning methods across a wider array of real-world problems.
Guo~\etal~\cite{cf4graph2023survey} summarized recent works about graph counterfactual learning, which only covers one important type of causal reasoning capability that might boost GNNs' trustworthiness. In addition, counterfactuals are often hard to identify in real-world scenarios with far more complex data generation and collection processes~\cite{seventools2019pearl}.
In contrast, we comprehensively review existing CIGNNs, distilling and categorizing a variety of techniques for empowering GNNs with abilities to handle different causal learning tasks. 
We also provide in-depth analyses of these ideas to unveil the nuanced connection between causal learning abilities and the improved GNN trustworthiness.

Several surveys have discussed different aspects of GNN trustworthiness, including generalizability, fairness, explainability and privacy~\cite{TrustGNNsurvey2022,tgnn2022peijian,TwGL2022,oodgraphsurvey2022,fairness4gnn2022survey,xgnnsurvey2022tpami,gce_survey2022}.
\eat{However, they mainly focused on a specific trustworthiness aspect and briefed several representative causality-based approaches, failing to provide a comprehensive examination of the advantages and commonalities of TGNNs from a causality-centric perspective.}
Differently, we provide an up-to-date survey of CIGNNs with illustrations on how causal learning can enhance GNN trustworthiness from the aspects of OOD generalizability, fairness, and explainability.
Moreover, we systematically extract and discuss the advantages and commonalities of existing CIGNNs, illuminating future integration of causal learning to reinforce these trustworthiness risks of GNNs as well as other underexplored aspects such as privacy.
Given the essential role of causal learning in boosting trustworthiness, we believe that 
our survey contributes a significant and unique perspective within GNN literature.

\eat{\TODO{This is a bad summarization. You also consider these three aspects. Are you trying to say existing survey focus on one aspect but you simultaneously focus on three? This is a bad point.}
\wz{Indeed, I have given a more clear summarization above.}}

\noindent 
\textbf{Intended Audiences.}
The survey targets two main groups of audiences, 
(i) researchers seeking insights into the rationale behind enhancing GNNs with causal learning abilities for future research endeavors, and (ii) practitioners interested in applying CIGNNs to improve trustworthiness in vital real-world applications, where generalizability across diverse data sources, fairness for different individuals or model explainability is highly demanded.
\eat{\TODO{Is empowering GNNs to conduct causal learning and causal-inspired GNN equivalent? Sounds not.}
\wz{I regard them equivalent. In fact, although causal-inspired GNNs are designed for graph mining tasks, their superiority over vanilla GNNs lie in that they are empowered to conduct causal learning, i.e. learning the causal mechanism of data. Also, the categorization in section 4 is based on which types of causal learning capacity the GNNs are empowered.}}

\noindent
\textbf{Survey Structure.}
The rest of this survey is organized as follows. Section 2 introduces preliminaries of GNN and causal learning. In Section 3, we analyze the trustworthiness risks of GNNs from a causal perspective. Based on the analysis, 
Section 4 introduces existing CIGNNs categorized under our proposed taxonomy, including discussions on representative methodologies and their impact on trustworthiness.
Benchmark datasets, evaluation metrics, and open-source codes and packages for conducting CIGNN research are respectively summarized in Section~5, 6, and 7. Section~8 concludes the survey and discusses future directions.

\section{Preliminaries}
In this section, we present preliminary knowledge of graph neural networks and causal learning.
Important notations are summarized in \appref{app:notations}.

\subsection{Graph Neural Networks}

The Graph Neural Network (GNN) has achieved state-of-the-art performance on various tasks when dealing with graph-structured data~\cite{gnnsurvey2021tnnls}.
The key idea of GNN is to map nodes into low-dimensional representations that simultaneously preserve structural and contextual knowledge.
Overall, existing GNNs can be partitioned into spatial-based and spectral-based categories.
We denote a graph as $\mathcal{G}=(\mathcal{V},\mathcal{E})$, where $\mathcal{V}$ is a set of nodes and $\mathcal{E}\in \mathcal{V} \times \mathcal{V}$ is a set of edges. Let $\mathbf{X}$ and $\mathbf{A}$ denote the graph's node attribute matrix and adjacency matrix. 
Following a message-passing scheme~\cite{mpnn2017icml}, spatial-based GNNs obtain node representations by iteratively transforming and aggregating node features and neighboring information,
\begin{align}
    \mathbf{a}_u^{(l+1)} &= \operatorname{AGGREGATE}(\{\mathbf{h}_v^{(l)} : v \in \mathcal{N}_u \}), \\
    \mathbf{h}_u^{(l+1)} &= \operatorname{COMBINE}(\mathbf{h}_u^{(l)} , \mathbf{a}_u^{(l+1)}),
\end{align}
where $\mathbf{h}_u^{(l)}$ is the representation of node $u$ output by the $l$-th GNN layer, $\mathbf{h}_u^{(0)}=\mathbf{x}_u$, $\mathcal{N}_u$ is the neighborhood of node $u$, $\operatorname{AGGREGATE(\cdot)}$ aggregates information from the neighbors of each node and $\operatorname{COMBINE(\cdot,\cdot)}$ updates the node representations by combining the aggregated information with the current node representations.
By stacking $k$ GNN layers, one can capture higher-order dependencies between the target node and its $k$-hop neighbors. 
\eat{\TODO{Also mention spectral based as an alternative.}
\wz{Added below.}
There is another branch of GNNs built upon spectral graph theory~\cite{spectral_before_1,spectral_before_2,powerfulspectral2022icml}. }
As an alternative paradigm, spectral-based GNNs~\cite{spectral_before_1,spectral_before_2,powerfulspectral2022icml} regard node representation matrix $\mathbf{H} \in \mathbb{R}^{|\mathcal{V}| \times d}$ as set of $d$-dimensional graph signals, and manage to modulate their frequencies in spectral domain~\cite{spectral}.
To achieve this goal, a graph convolution operator is defined, which consists of three key steps, (i) transforming graph signals into spectral domain via Graph Fourier Transform (GFT),
\begin{align}
    \hat{\mathbf{H}}^{(l)} = \mathbf{U}^T \mathbf{H}^{(l)},
\end{align} 
where $\mathbf{U}$ is a complete set of orthonormal eigenvectors of graph $\mathcal{G}$'s corresponding Laplacian matrix, (ii) modulate Fourier coefficients $\hat{\mathbf{H}}^{(l)}$ as $g(\mathbf{\Lambda}) \hat{\mathbf{H}}^{(l)},$ where $g(\cdot)$ is a learnable graph filter and $\mathbf{\Lambda} = \text{diag}(\lambda_1, ..., \lambda_{|\mathcal{V}|})$ is the eigenvalues of the corresponding Laplacian matrix, (iii) applying inverse GFT to transform filtered Fourier coefficients back to spatial domain and obtain the reconstructed signals 
\begin{align}
    \mathbf{H}^{(l+1)} = \mathbf{U} g(\mathbf{\Lambda}) \hat{\mathbf{H}}^{(l)} = \mathbf{U} g(\mathbf{\Lambda}) \mathbf{U}^T \mathbf{H}^{(l)}.
\end{align}


Once the node representations are obtained, they can be used for downstream tasks by incorporating a predictor $w(\cdot)$, \eg a Multi-Layer Perceptron (MLP), to map the representations into label space.
\eat{\TODO{node representation maybe followed with a readout function?}
\wz{
1) It's true for graph-level tasks. Here I only aim to present the high-level connection between representation and downstream tasks. The readout will be mentioned in the introduction of graph-level tasks.
2) I changed the order between 'The entire model ... solving downstream tasks.' and 'The downstream tasks of GNNs ... predict graph labels~\cite{diffpool2018nips,sapooling2019icml}'.}}
The downstream tasks of GNNs can be roughly categorized into node-level, edge-level, and graph-level.
For node-level tasks, such as node classification and regression, the node representations can be directly fed into the downstream predictor to output predicted node labels. 
For edge-level tasks, such as link prediction, the representations of both nodes in each node pair serve as the input of the predictor to derive predictions. 
For graph-level tasks, such as graph classification, graph pooling is commonly required to further aggregate node representations into a unified graph representation $\mathbf{h}^{\mathcal{G}}$ to predict graph labels~\cite{diffpool2018nips,sapooling2019icml}.
\eat{
}
The entire model can be trained with downstream labels in an end-to-end way~\cite{gnnsurvey2021tnnls} or in a pretrain-finetune fashion~\cite{pretraingnn2022}. 
Compared with other graph learning approaches such as network embedding~\cite{netembed2019tkde}, GNN preserves both contextual and structural information under the supervision of task-specific signals, which are generally more effective in solving downstream tasks~\cite{gnnsurvey2021tnnls}.

\subsection{Causal Learning} \label{subsec:causal_learning}
\eat{
In general, machine learning tasks can be categorized as either predictive or descriptive. However, there is also a growing interest in understanding causality, where one can imagine modifying variables and rerunning the data-generating process. These types of questions can take two related forms: (1) quantifying the impact of changing a specific variable on other variables (features or the label), and (2) identifying which variables, when modified, can influence the value of another variable. These questions are commonly referred to as causal effect estimation and causal discovery, respectively. However, there is also a growing interest in understanding causality, where one can imagine modifying variables and rerunning the data generation process.
}
Causal learning investigates the cause-and-effect relations between variables to enable robust predictions and informed decisions in various real-world situations~\cite{causality2009pearl,stat2causal2022ICM}.
\eat{\TODO{1. scientific methods is unclear, 2. if the statistical approach, where is learning?}
\wz{
Revised as above. I deleted "using rigorous scientific methods and statistical analyses" due to two reasons: (i) various types of statistical, ML-based and DL-based methods are proposed for each fundamental task of causal learning, so I feel it hard to use several concise terms to summarize all these methods, only '';
(ii) I checked some papers and found most of them do not summarize the methods when first introducing causal learning.}}
Overall, there are three fundamental tasks in causal learning: 
(i) \emph{causal reasoning}, (ii) \emph{causal discovery} and (iii) \emph{causal representation learning}~\cite{stat2causal2022ICM}.
In this part, we begin with the two cornerstone causal learning frameworks, \ie \emph{potential outcome framework} and \emph{structural causal model}, and 
elaborate on how they enable a consistent formulation of fundamental causal learning tasks.
\eat{\TODO{assumptions of what?}
\wz{assumptions on the underlying data generation process}
\TODO{when you talk about consistent formulation, are you talking about how to formulate fundamental causal tasks (2.2.2)? I don't get the point.}
\wz{Formulation here means 'accurate and rigorous expression', so we can not only formulate tasks, but also prior knowledge and assumptions. The enriched sentence is currently masked because I actually did not talk detailedly on the assumptions later. Therefore, I think only mentioning 'a consistent formulation of fundamental causal learning tasks' is enough, which indeed connects to 2.2.2.}}
We also introduce some basics of \emph{causal identification} in causal learning tasks.

\subsubsection{Causal Learning Frameworks}
\eat{\TODO{You have used causal learning, causal inference, and causal modeling in different places but without proper justification.}
\wz{Causal inference, and causal modeling have been replaced with causal learning.}}
\textbf{Potential Outcome Framework (POF).}
The POF raised by Rubin~\etal~\cite{potential_outcome} proposes a concept of \emph{potential outcome} to describe the counterfactual outcomes under varied treatments.
\begin{definition}[Potential Outcome]
     A potential outcome $Y_i(t)$ for an individual $i$ is defined as the outcome that would be observed if the individual were assigned a specific treatment $t$. 
\end{definition}
\noindent
Here, $Y(t)$ is a random variable that represents the outcome of interest, \eg{health status, income, test score}. The well-definedness of potential outcomes can typically be established by making 
SUTVA and consistency assumption~\cite{stat2causal2022ICM}.

\noindent
\textbf{Structural Causal Model~(SCM).}
The SCM framework proposed by Pearl~\cite{causality2009pearl} enables a rigorous description of the underlying causal mechanisms of complex systems.
\begin{definition}[Structural Causal Model]
    An SCM $\mathcal{M} = (\mathcal{X}, \mathcal{U}, \mathcal{F}, P_\mathcal{U})$ consists of: (i) a set $\mathcal{X}$ of endogenous variables; (ii) a set $\mathcal{U}$ of exogenous variables which have no causal parents or direct causes, and follow a joint distribution $P_\mathcal{U}$; (iii) a set $\mathcal{F}$ of deterministic functions computing each $X_i \in \mathcal{X}$ from its causal parents, $\mathcal{PA}_i \subset \mathcal{X} \setminus {X_i}$ and the corresponding $U_i \in \mathcal{U}$ via the structural equations $\{X_i := f_i(\mathcal{PA}_i,U_i) \}_{i=1}^n.$
\end{definition}

An SCM naturally induces a directed \textit{causal graph}, representing endogenous variables and their causal relations.
Usually, we regard causal graphs as Directed Acyclic Graphs (DAGs)~\cite{stat2causal2022ICM}.
\eat{
}
There are three typical DAGs, (i) \emph{fork}, $T \leftarrow X \rightarrow Y,$ (ii) \emph{chain}, $T \rightarrow X \rightarrow Y,$ and (iii) \emph{immorality}, $T \rightarrow X \leftarrow Y.$ Node $X$ serves as the \textit{confounder}, \textit{mediator} and \textit{collider}, respectively. In each DAG, $X$ could lead to a spurious correlation between $T$ and $Y,$ hindering the estimation of causal relations~\cite{causality2009pearl}.
Notably, the wide existence of the three DAGs in graphs hinders correlation-based GNNs from capturing causality, which will be elucidated in \secref{sec:trust_risk}.
\eat{
}

\eat{\TODO{I feel a gap between SCM, DAG, and three example graphs. Can everyone know \textit{confounder}, \textit{mediator}, \textit{collider} well without explaination or reference?}
\wz{added examples and references.}}

Under SCMs, we use \emph{do-operator} $do(X:=x)$ or $do(x)$ to signify performing an intervention on variable $X$ by assigning it a value $x,$ which leads to the intervened SCM $\mathcal{M}^{do(X := x)}$ with structural equation of $X$ replaced by $X := x.$
The \emph{interventional distribution} of $Y$ under $\mathcal{M}^{do(X := x)}$ is denoted as $P(Y|do(x)).$
Intuitively, $P(Y | do(x))$ reflects the causal effect of $X$ on $Y$. Several methods estimate certain interventional distributions on graphs to construct TGNNs, which will be detailed in \secref{sec:causalgnn}.

\eat{
The SCM also entails a quantitative formulation of causal intervention, which is widely known as \textit{do-operator}~\cite{causality2009pearl}.
\begin{definition}[Do-operator]
    The do-operator, denoted as $do(X:=x)$ or $do(x)$ for brevity, signifies that an intervention is performed on the variable $X$ by assigning it a value of $x$.
\end{definition}
\noindent With the do-operator, we can further formulate the data distribution after intervention in SCM.
\begin{definition}[Interventional Distribution]
    The interventional distribution $P(Y | do(X:= x))$, or $P(Y|do(x))$ for brevity, denotes the distribution of any variable $Y$ generated under the intervened SCM $\mathcal{M}^{do(X := x)}$, whose structural equation \wrt{$X$} is replaced with $X := x.$
\end{definition}
\noindent In particular, if we interpret $Y$ as the label and $X$ as the graph features, then $P(Y | do(x))$ actually reflects the causal effect of $X$ on $Y$. 
Several methods reviewed in this paper are aimed at estimating this type of distribution on graph data in order to facilitate GNNs to learn causality and improve their trustworthiness, which will be detailed in \secref{sec:causalgnn}.
}

With the foundational concepts of POF and SCM, we proceed to formulate the three causal learning tasks.

\subsubsection{Formulation of Fundamental Causal Learning Tasks} \label{subsubsec:formulate_causal}
This part lays the groundwork for subsequent discussions of the connections between causal learning and CIGNNs. We first formulate causal reasoning under the POF and SCM. Then, we formulate causal discovery and causal representation learning under the SCM, given its flexibility in handling complex causal relations and unobserved variables.
\eat{\TODO{Again, you need a transition/overview, why introduce three tasks.}
\wz{I have added one. I think here a transition sentence should briefly introduce the importance of 'formulation' rather than why we introduce these three tasks. The importance of these three tasks are long discussed and perceived as fact in causal inference community, while the importance of them for building trustworthy GNNs are just the main theme of section 4.}}

\noindent
\textbf{Causal Reasoning.}
Causal reasoning aims to quantify cause-effect relations among variables.
\eat{\TODO{why ITE is essential in this survey?}
\wz{Explained as below.}}
Under POF, given a group of individuals that are assigned either treatment $t$ or $t',$ we quantify the Individual Treatment Effect (ITE) of individual $i,$ $\text{ITE}_i = Y_i(t') - Y_i(t),$ to boost more trustworthy decision-making in situations where individual interests matter.
Particularly, ITEs can reflect the causality between features and the label of each node or graph instance, which might improve the trustworthiness of GNNs~\cite{CGI2021sigir,gem2021icml}.
Unfortunately, observing both factual outcome $Y_i(t) = Y_i$ and counterfactual outcome $Y_i(t')$ is often impossible due to ethical or cost issues~\cite{causal_data_survey2020}. 
As a compromise, people approximately quantify the causal relations at a (sub)group level, \eg estimating Average Treatment Effect, $\text{ATE} = \mathbb{E}[Y(t')-Y(t)],$ or Conditional Average Treatment Effect, $\text{CATE}(\mathbf{x}) = \mathbb{E}[Y(t')-Y(t)|\mathbf{X}=\mathbf{x}].$ 
\eat{\TODO{I don't get what you want to say.}\wz{revised.}}
Under the SCM framework, individual-level causal reasoning focuses on estimating counterfactual distribution $P(Y(t')|Y_t, t)$~\cite{causality2009pearl}. Group-level tasks involve estimating $P(y|do(t))$ for varied treatment $t$, which enables calculating both ATE and CATE.
\eat{
Specifically, $\argmax_y P(y|do(t))$ can be a good estimation of potential outcome $Y(t)$ when there 
The above causal estimands can also be well formulated under SCM with the help of do-calculus. 
The potential outcome $Y(t)$ can be expressed as $Y|do(t).$
The expectation of the potential outcome can thus be described by interventional distribution as 
\begin{align*}
    \mathbb{E}[Y(t)] &= \mathbb{E}[Y|do(t)] = \int yP(y|do(t))dy.
\end{align*}
Therefore, under the SCM framework, the key to estimating ITE lies in updating the distribution of exogenous variables based on the observation, which enables the computation of potential outcome $Y|do(t')$ in the intervened SCM $\mathcal{M}^{do(T := t')}.$ 
As for ATE, our primary task is to estimate the interventional distribution under different treatments.}

\noindent
\textbf{Causal Discovery.}
\eat{
}
Formally, given an observed dataset $\mathcal{D}$ generated by an SCM $\mathcal{M}$, the causal discovery task recovers the causal graph induced by $\mathcal{M}$ from $\mathcal{D}$~\cite{causal_data_survey2020}.
For example, in transportation networks, causal discovery aims to answer questions like 'Is the road restriction at one road segment causally related to traffic congestion at another segment?' or 'Does the nearby sports event induce confounding effects between road restriction and congestion?'
\eat{
This involves the identification of:
Causal Markov Condition: A variable X is independent of its non-descendants, given its parents, in the true causal graph G. This condition implies that a variable is conditionally independent of all other variables in the graph, given its direct causes.
Causal Faithfulness Condition: All observed conditional independence relations in the dataset D are a result of the causal Markov condition, and there are no additional independencies in the true causal graph G.
Causal Sufficiency: The observed dataset D includes all common causes of the variables in V, and there are no hidden or latent confounders that influence multiple variables in V.
The goal of the causal discovery task is to uncover the true causal graph G that satisfies these conditions, using various statistical and algorithmic techniques such as constraint-based methods (e.g., PC algorithm), score-based methods (e.g., Bayesian Information Criterion), and hybrid approaches (e.g., Greedy Fast Causal Inference).
}

\eat{
\TODO{Causal discovery should be introduced: (1) only SCM can do; (2) the aim of introduction is for understanding the future direction, so need to highlight the necessity of this task, i.e., the prior knowledge is often insufficient}
\begin{definition}[Markov Blanket]
    A Markov blanket of Y under distribution P is any subset S of X for which
    $Y ⊥ (X\setminus S) | S$.
\end{definition}
}

\noindent
\textbf{Causal Representation Learning~(CRL).}
Suppose that the low-level observations $\mathcal{X} = \{X_1,..., X_n \}$ are generated by a few latent variables $\mathcal{S} = \{ S_1 , ..., S_d \}$, where $d \ll n$. 
\eat{\TODO{lost after in that xx.}
\wz{Replaced "in that" with "i.e.". The sentence after "i.e." is a more formal expression of the one before "i.e."}}
The latent variables $\mathcal{S}$ may be dependent and possess an underlying SCM $\mathcal{M}_S.$
CRL aims to recover $\mathcal{S}$ along with the causal relations. This can be crucial for reasoning about the underlying causal mechanism of the world~\cite{stat2causal2022ICM}.
For instance, within low-level visual images, a pendulum, light source and shadow may be causally related. By learning causal representations for these factors, one can estimate the counterfactuals of the shadow after manipulating the pendulum’s angle~\cite{icmvae2023nips}.

\eat{\TODO{Causal discovery and causal representation learning are embarrassing simple compared with causal reasoning. More importantly, I don't think they are clearly elaborated.}
\wz{
Introduction on causal discovery is short because there is indeed no paper incorporating causal discovery methods to improve GNNs. 
But I think it is still necessary to introduce causal discovery because 
(i) causal discovery is one of the most fundamental causal learning tasks. Introducing causal preliminaries without mentioning it seems weird;
(ii) I discussed a future direction related to causal discovery in the final section.

The definitions of CRL in orange are copied from~\cite{stat2causal2022ICM} (will revise later), which I believe is rigorous. So to elaborate more clearly, I added an example.}}

\subsubsection{Identification of Causal Quantities}
\label{subsubsec:identification}
\eat{
\TODO{What is the relationship between this part with ``transform causal learning tasks into tractable statistical estimation problems '' mentioned in the beginning of 2.2?} 
\wz{They are the same thing. I have revised the last sentence before section 2.2.1 to make this connection clearer.}
\TODO{You are the stakeholder of this survey and you should try your best to polish the survey after you write the very initial draft.}
\wz{Understood.}
}
A fundamental challenge of causal learning is how to estimate various causal quantities defined in the interventional or counterfactual world,
namely the \emph{causal identification} process~\cite{causality2009pearl}.
The Randomized Control Trial (RCT) is the golden standard approach to identify causal effects or relations~\cite{stat2causal2022ICM}. 
However, RCTs are often unfeasible in real-world applications due to ethical or cost issues. 
Therefore, researchers have developed practical identification methods that rely on data assumptions that are testable in principle or can be verified based on expert knowledge~\cite{causality2009pearl}.

\eat{
In this part, we only introduce two classical identification strategies for causal reasoning, backdoor and frontdoor adjustment, 
\rev{under the causal graph in \figref{fig:general_causalgraph}.}
\begin{figure}[h]
    \centering
    \includegraphics[width=0.3\linewidth]{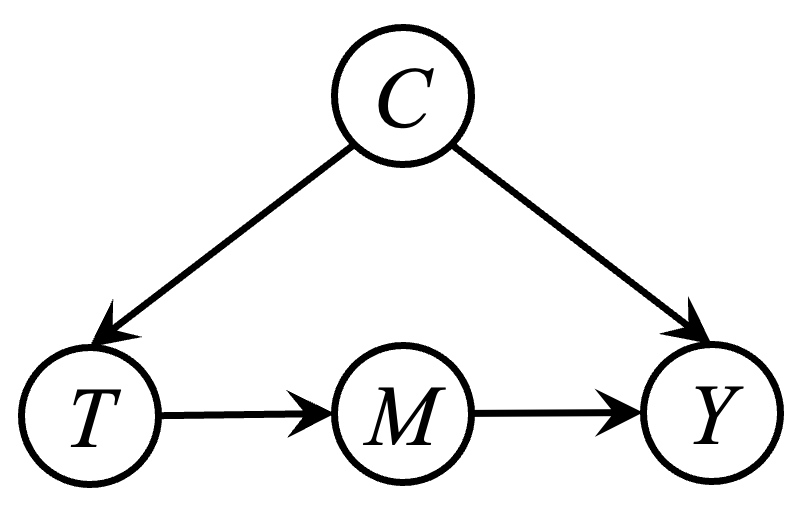}
    \caption{\rev{A illustrative causal graph for backdoor and frontdoor adjustment.}}
    \label{fig:general_causalgraph}
\end{figure}
}
In the following, we introduce the \emph{backdoor adjustment} for identification in causal reasoning tasks.
For further insights into the identification in causal discovery and CRL, kindly consult references~\cite{reviewcd2019,cdsurvey2023acm,identify_dgm2022nips}.

\noindent
\textbf{Backdoor Adjustment.} 
The existence of \textit{backdoor path}, which is an indirect path from treatment $T$ to outcome $Y$ with no collider along it, can lead to spurious correlations between $T$ and $Y$~\cite{causal_data_survey2020}. Notably, a confounder can create a backdoor path. 
Backdoor adjustment is a feasible way to identify $P(Y|do(T:=t))$ when backdoor paths are existing.
\begin{theorem}[Backdoor Adjustment]
Given observable variables $C$ that blocks all backdoor paths between $T$ and $Y$, under modularity and positivity assumptions~\cite{stat2causal2022ICM}, we have
\begin{align}\label{eqn:backdoor}
    P(Y|do(T:=t)) = \sum_{C}P(Y|T=t,C)P(C).
\end{align}
\end{theorem}
\noindent
Intuitively, backdoor adjustment stratifies the data based on $C$. Within each stratum, the spurious correlations induced backdoor paths are eliminated by fixing $C$, thus enabling the estimation of the causal effect of $T$ on $Y$ using statistical quantities that are more readily estimable from the data.
\section{Deciphering GNN Pitfalls: a Causal Lens}
\label{sec:trust_risk}

Causal learning empowers us to decipher the essential limitations of GNNs by inspecting the underlying data generation process.
In this section, we conduct an in-depth causal analysis of three typical trustworthiness risks of GNNs, emphasizing the necessity of understanding causal mechanisms inherent in graph data for constructing TGNNs.

\subsection{Out-Of-Distribution Generalizability in GNNs}
\label{subsec:ood}
We begin by formally introducing the graph OOD generalization problem below.
\begin{definition}[OOD Generalization on Graphs~\cite{oodgraphsurvey2022}]
Given a training set $\mathcal{D}^{tr} =\{(G_i,Y_i)\}^N_{i=1}$ drawn from training distribution
$P^{tr}(\mathcal{G},Y)$, where $G_i$ denotes a graph instance and $Y_i$ is the label,
we aim to learn an optimal graph predictor $f_{\theta^*}$ from $\mathcal{D}^{tr}$ that performs best on testing set drawn from testing distribution $P^{te}(\mathcal{G}, Y)$ where $P^{te}(\mathcal{G}, Y) \not= P^{tr}(\mathcal{G}, Y)$, \ie
\begin{align}
    f_{\theta^*} =\argmin_\theta \mathbb{E}_{\mathcal{G},Y \sim P^{te}}[\mathcal{L}(f_\theta(\mathcal{G}),Y)].
\end{align}
\end{definition}

\noindent
Distribution shifts can occur at both the feature-level and topology-level in training and testing graph datasets~\cite{oodgnn2022tkde}. In each level, the distribution shifts can be further categorized into covariate shift and concept shift, \ie{$P^{te}(\mathcal{G}) \not= P^{tr}(\mathcal{G})$ and $P^{te}(Y|\mathcal{G}) \not= P^{tr}(Y|\mathcal{G})$}~\cite{oodsurvey2021}.
Built upon i.i.d. assumption, mainstream GNNs tend to achieve high In-Distribution~(ID) accuracy but 
exhibit unstable performance in OOD scenarios~\cite{oodgraphsurvey2022}.
The essential reason is that GNNs tend to capture and rely on spurious correlations between non-causal graph components and the label, which can vary across data distributions shifted from training data~\cite{DIR2022iclr,stable2022nature}.

Considering a general graph generation process shown in \figref{fig:graph_DGP_ood+exp}, variant latent factor $V$ and label $Y$ might be spuriously correlated resulting from (i)~\emph{confounding bias} induced by confounder $C$~\cite{DisC2022nips}, \eg{the authors' affiliations of a paper in a citation network might causally affect its unimportant citation patterns ($V$) and its impact ($Y$), leading to spurious correlations between $V$ and $Y$}; (ii)~the correlation between $V$ and invariant latent factor $I,$ which exists due to the \emph{data selection bias}~\cite{stable2022nature} caused by conditioning on the graph $\mathcal{G}$, \eg{selecting a class of molecule graphs with the same type of scaffold ($V$) will induce spurious correlation between $V$ and the class-discriminative patterns ($I$)}; and (iii)~\emph{anti-causal effect} from $Y$ to $V$~\cite{EERM2022iclr,ciga2022nips}, \eg{the high impact ($Y$) of a paper may also result in some unimportant citation patterns ($V$), which conversely providing support for the paper's high impact.} 
Instead of capturing underlying causal mechanism, GNNs may learn spurious correlations $P^{tr}(Y|V)$, resulting in unstable performances when tested on datasets where $P^{te}(Y|V) \neq P^{tr}(Y|V)$.

Therefore, developing GNNs that can filter out such spurious correlations and capture the invariant causal relations $P(Y|I)$ becomes important to achieve stable and generalizable OOD prediction.
\eat{
\begin{figure}[t]
\centering
\includegraphics[width=100pt]{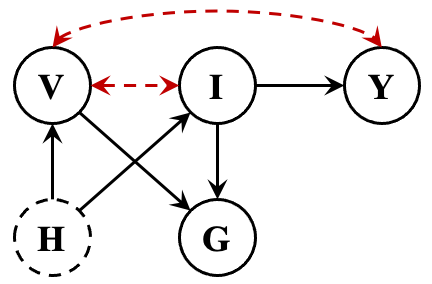}
\caption{The causal graph of the underlying data generation process in node- or graph-level prediction tasks. $\textbf{G}$ denotes the local subgraph around the node sample or the target graph sample, $\textbf{V}$ and $\textbf{I}$ denote the variant and invariant latent variables which generate the (sub)graph, respectively, $\textbf{Y}$ denotes the label of the target node or graph and $\mathbf{H}$ denotes the hidden confounders.}
\label{fig:ood_causalgraph}
\end{figure}
}
\begin{figure}[t] 
\centering
\subfigure[]{
    \label{fig:graph_DGP_ood+exp}
    \includegraphics[width=0.35\linewidth]{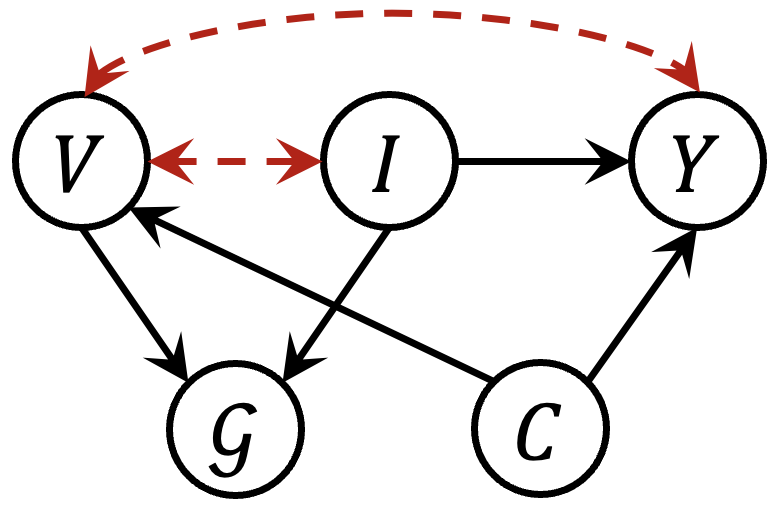}}
\quad \quad
\subfigure[]{
    \label{fig:graph_DGP_fair}
    \includegraphics[width=0.35\linewidth]{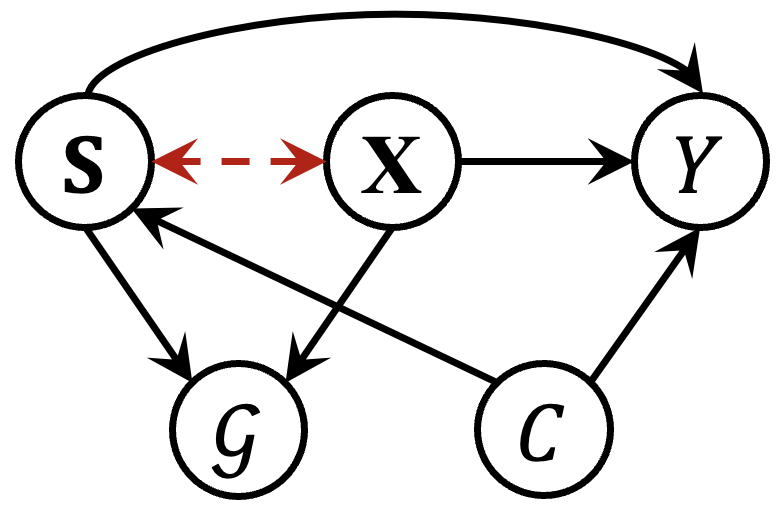}}
\caption{
Two causal graphs that characterize the graph generation process in graph- or node-level tasks.
$Y$ denotes the label or the model prediction of $\mathcal{G}.$ $C$ denotes (hidden) confounders. The black solid arrow indicates causal relation and the red dashed arrow indicates spurious correlation.
Fig. (a) helps reveal the reasons for GNNs' poor OOD generalizability and explainability, where $V$ and $I$ denote the variant and invariant graph generation factors that causally and non-causally affect $Y,$ respectively.
Fig. (b) aids in explaining graph unfairness, where $\mathbf{S}$ and $\mathbf{X}$ denote the sensitive and insensitive graph attributes, respectively.}
\end{figure}

\eat{
The above analysis implies that it is significant for GNNs to gain a deeper understanding of the underlying causal mechanisms operating within the graph so as to make informed predictions on OOD scenarios.
To achieve this goal, one promising way is to enable GNNs to conduct causal reasoning. In this way, GNNs can eliminate various data biases and learn the causal influences from observed graph features to the target label, which are stable and generalizable across different data distributions.
Besides, equipping GNNs with the ability to perform CRL holds promising potential. CRL allows GNNs to extract the essential causal variables that govern the graph structure while encoding graph data. By explicitly capturing the  
}

\subsection{Fairness in GNNs} 
\label{subsec:fairness_analysis}
Unfairness issue in GNNs leads to discriminatory predictions against certain populations with sensitive attributes.
Over the years, several correlation-based graph fairness notions have been proposed to reveal the existence of unfairness in model predictions~\cite{fairness4gnn2022survey}. 
However, they might increase discrimination without knowing the underlying causal mechanisms that lead to unfairness~\cite{cf2017nips}. 
The notion of Graph Counterfactual Fairness (GCF), which is defined based on causality, can address the limitation.
\eat{\TODO{Here is a gap, several notions $\rightarrow$ correlation-based notions $\rightarrow$ GCF.}
\wz{Revised.}}
\eat{
\TODO{You should justify this kind of fairness is general and important in all kinds all fairness. Is this mean causal can only help this kind of fairness?}
\wz{
- The third sentence is aimed to show the limitations of other fairness notions. I further added a sentence after definition 9 to show its specialty and importance. As for generality, I justified that GCF can cover a wide range of scenarios.
- This fairness notion is currently the only one defined based on causal quantities. To achieve this fairness notion, it is also necessary to use causality-inspired methods.}
}
\begin{definition}[Graph Counterfactual Fairness~\cite{gear2022wsdm}] \label{def:gcf}
    An encoder $\Phi(\cdot)$ satisfies GCF if for any node $u$,
    \begin{align}
        P(\mathbf{h}_u | do(\mathbf{S}:=\mathbf{s}'), \mathbf{X}, \mathbf{A}) = P(\mathbf{h}_u | do(\mathbf{S}:=\mathbf{s}''), \mathbf{X}, \mathbf{A}),
        \label{eqn:gcf}
    \end{align}
    holds for all $\mathbf{s}' \not= \mathbf{s}''$, where $\mathbf{s}', \mathbf{s}'' \in \{0, 1\}^{n}$ are arbitrary sensitive attribute values of \textbf{all nodes} and $\mathbf{h}_u$ denotes the representation of node $u$ output by $\Phi(\cdot).$
\end{definition}
\noindent
GNNs that meet correlation-based fairness notions may not meet GCF if they are unaware of the causal effects, leading to 
unfairness issue when there exists statistical anomalies~\cite{cf2017nips,causalfairsurvey2022}.
Pursuing GCF sets a higher standard for GNNs to understand the causal mechanism within graph features and ensure that sensitive attributes do not causally affect the output node embeddings.

From a causal standpoint, we elaborate on three possible reasons why mainstream GNNs fail to achieve GCF based on \figref{fig:graph_DGP_fair}:
(i) similar to the causal mechanism in \figref{fig:graph_DGP_ood+exp}, there might be spurious correlations between the sensitive attribute $\mathbf{S}$ and its label $Y$ due to various biases in graph data, \eg{the data selection bias caused by the imbalance of nodes belonging to different sensitive groups.}
The node representations produced by GNNs that rely on such spurious correlations should be encoded with sensitive information, leading to the violation of \equref{eqn:gcf};
(ii) differently, $\mathbf{S}$ might have causal effects on $Y$~\cite{gear2022wsdm}. Without any fairness-aware mechanisms, mainstream GNNs will generate node embeddings that are causally affected by $\mathbf{S}$, thereby violating GCF. Even if the node embeddings are statistically uncorrelated with $\mathbf{S}$, there is no guarantee that the same node embeddings will be produced on counterfactual graphs with varied $\mathbf{S}$~\cite{cf2017nips,nifty2021uai,gear2022wsdm};
(iii) the sensitive attributes of a node's neighbors might also have a causal effect on its label~\cite{gear2022wsdm}, \eg{a man's loan application might be disapproved due to the race of his friends.} Similar to (ii), this will also lead to the violation of GCF.

In summary, both spurious correlations and causal relations between $S$ and $Y$ can pose threats to the GCF of mainstream GNNs. It is thus a necessity for GNNs to be aware of the causal mechanisms leading to unfairness in order to achieve fair node embeddings in both factual and counterfactual graphs.

\subsection{Explainability in GNNs}
\label{subsec:xgnn}
\eat{\TODO{Do you need a formal definition like Def 7 \& 8?}
\wz{added.}}
Although GNNs are more interpretable than other types of deep neural networks because of the message-passing scheme, 
they still cannot avoid the opacity of feature mapping and information propagation within hidden GNN layers.
Recent efforts attempt to alleviate the black-box nature of GNNs through generating \emph{post-hoc explanations}~\cite{xgnnsurvey2022tpami} or improving their \emph{inherent interpretability}~\cite{disengcn2019icml}. 
\begin{definition}[Post-hoc explainability]
    It refers to the ability to identify a collection of human-understandable graph components, \eg{nodes, edges, or subgraphs}, that contribute to a given prediction of the target GNN.
\end{definition}
\begin{definition}[Inherent Interpretability]
    It refers to the alignment of the GNN inference mechanism with human-understandable principles.
\end{definition}
\noindent
\eat{
Causal effects of edges. It is crucial to specify the edges that are the plausible causation of the model outcome, rather than the edges irrelevant or spuriously correlated with the outcome [21], [22]. However, as the explainers using gradient- and attention-like scores typically ap- proach the input-outcome relationships from an associ- ational standpoint, they hardly distinguish causal from noncausal effects of edges. Take Figure 1 as a running example, where SA [13] and GNNExplainer [16] explain why the GIN model [23] predicts the molecule graph as mutagenic. As the nitrogen-carbon (N-C) bond often connects with the nitro group (NO2), it is spuriously correlated with the mutagenic property, thus ranked top
}
Nevertheless, these studies still lack reliability since they are confined to correlation modeling.

Mainstream post-hoc explanation methods typically learn to measure the importance score of different components of the input graph to the model prediction and attribute a model's prediction to those with the highest scores~\cite{ig2020nips,gradcam2019cvpr,gnnexplainer2019nips}. 
However, the generated importance scores might be unable to measure the causal effects of input graph components, but overrate unimportant components that are spuriously correlated with the model prediction~\cite{gem2021icml}.
\eat{
}
The spurious correlations can be induced by confounding bias or data selection bias, which can be concluded similarly if interpreting $Y$ as the model prediction and $V$ as the variant graph components that do not causally affect the model prediction in \figref{fig:graph_DGP_ood+exp}.
Besides, considering the potential OOD risk of the target GNN,
the distribution shift from the original graphs to the candidates serves as a special hidden confounder $C$ between $V$ and $Y$~\cite{DSE2022,align2023wsdm}.
$C \rightarrow V$ exists because the candidates are generated by inducing a distribution shift on original graphs.
$C \rightarrow Y$ exists because the distribution shift influences predictions produced by a trained GNN that suffers from the OOD problem.
\eat{\TODO{So the limitation of post-hoc explanation also belongs to OOD problem?}
\wz{OOD problem talks about the failure in predicting OOD samples. Here we are talking about that when interpreting traditional GNN with OOD problem, the process to measure the importance score of graph components might be confounded by the factor 'distributions shift'.}}

The reliability of inherent interpretable GNNs built upon attention or disentanglement mechanisms is also being questioned due to their limited understanding of the underlying causal mechanisms.
Attention mechanisms primarily model correlations between graph components rather than capturing deeper causal relations~\cite{gat}. 
Although existing graph disentanglement methods can help uncover latent factors in the graph formulation process and enhance the interpretability of information propagation within GNNs~\cite{disengcn2019icml}, it is more essential to identify causal latent factors in order to establish inherently interpretable GNNs that operate based on causal mechanisms~\cite{meta_CRL2020iclr,weaklyCRL2022jmlr}.

To sum up, spurious correlations are the key source of the limited explainability of current GNN systems. 
Causal learning is thus demanded to eliminate those spurious correlations captured by the post-hoc explainer and guide the latent representations of GNNs to move beyond preserving statistical dependence structure to causal structures, thus enhancing the explainability in both perspectives.

\section{Causality-Inspired GNNs} \label{sec:causalgnn}
\begin{figure*}[t]
\centering
\usetikzlibrary{shapes, arrows.meta, positioning}
\tikzstyle{heading} = [rectangle, rounded corners, minimum width=1cm, minimum height=1cm, text centered, draw=black, path picture={
    \shade[shading angle=90, top color=white, bottom color=purple!30] (path picture bounding box.south west) rectangle (path picture bounding box.north east);
},
text width=1.5cm, align=center]

\tikzstyle{subheading1} = [rectangle, rounded corners, minimum width=1cm, minimum height=0.5cm, text centered, draw=black, path picture={
    \shade[shading angle=90, top color=white, bottom color=blue!30] (path picture bounding box.south west) rectangle (path picture bounding box.north east);
},
text width=2.6cm, align=center]

\tikzstyle{subheading2} = [rectangle, rounded corners, minimum width=2cm, minimum height=1cm, text centered, draw=black, path picture={
    \shade[shading angle=90, top color=white, bottom color=teal!30] (path picture bounding box.south west) rectangle (path picture bounding box.north east);
},
text width=1.5cm, align=center]

\tikzstyle{content} = [rectangle, rounded corners, minimum width=1cm, minimum height=0.5cm, text centered, draw=black, path picture={
    \shade[shading angle=90, top color=white, bottom color=darkgray!30] (path picture bounding box.south west) rectangle (path picture bounding box.north east);
},
text width=2cm, align=center]

\tikzstyle{citation} = [rectangle, rounded corners, minimum width=1cm, minimum height=0.5cm, text centered, draw=black, path picture={
    \shade[shading angle=90, top color=white, bottom color=lightgray!30] (path picture bounding box.south west) rectangle (path picture bounding box.north east);
},
text width=6.37cm, align=left]


\tikzstyle{arrow} = [thick,-,>=stealth]

\begin{tikzpicture}[node distance=0.7cm, auto, every node/.style={font=\scriptsize}]
    \node (title) [heading] {Causal Learning on Graphs};
    \node (s1_top) [subheading1, right=of title, anchor=south west, yshift = 1.7cm] {Causal Reasoning on Graphs};
    \node (s1_below) [subheading1, right=of title, anchor=north west, yshift = -1.75cm] {Causal Representation Learning on Graphs};
    \node (s2_top) [subheading2, right=of s1_top, anchor = south west, yshift = 1.63cm]{Group-level Causal Effect Estimation};
    \node (s2_mid) [subheading2, right=of s1_top, anchor=west]{Individual-level Causal Effect Estimation};
    \node (s2_below) [subheading2, right=of s1_top, anchor=north west, yshift = -1.33cm]{Counterfactual Explanation Generation};
    \node (s2_4) [subheading2, right=of s1_below, anchor=north west, yshift = 1.25cm]{Supervised Learning};
    \node (s2_5) [subheading2, right=of s1_below, anchor=west, yshift = -0.88cm]{Self-supervised Learning};
    \node (text_1) [content, right=of s2_top, anchor=south west, yshift = 0.42cm]{Instrumental Variable};
    \node (text_2) [content, right=of s2_top, anchor= west]{Frontdoor Adjustment};
    \node (text_3) [content, right=of s2_top, anchor= north west, yshift = -0.41cm]{Stable Learning};
    \node (text_4) [content, right=of s2_mid, anchor= south west, yshift = 0.42cm]{Intervention};
    \node (text_5) [content, right=of s2_mid, anchor= west]{Matching};
    \node (text_6) [content, right=of s2_mid, anchor= north west, yshift = -0.3cm]{Deep Generative Modeling};
    \node (text_7) [content, right=of s2_below, anchor= west, yshift = 0.35cm]{Continuous Optimization};
    \node (text_8) [content, right=of s2_below, anchor= west, yshift = -0.35cm]{Heuristic Search};
    \node (text_9) [content, right=of s2_4, anchor= south west, yshift = 0.12cm]{Group Invariant Learning};
    \node (text_10) [content, right=of s2_4, anchor= north west, yshift = -0.03cm]{Joint Invariant and Variant Learning};
    
    \node (citation_1) [citation, right=of text_1]{RCGRL\cite{RCGRL2022}};
    \node (citation_2) [citation, right=of text_2]{DSE\cite{DSE2022}};
    \node (citation_3) [citation, right=of text_3]{DGNN\cite{debiasedgnn2022tnnls}, StableGNN\cite{stablegnn2021}, OOD-GNN\cite{oodgnn2022tkde}, L2R-GNN~\cite{L2R-GNN2024aaai}};
    \node (citation_4) [citation, right=of text_4]{CGI\cite{CGI2021sigir}, CSA\cite{csa2023ijcai}, DCE-RD\cite{DCERD2023kdd}, NIFTY\cite{nifty2021uai}, MCCNIFTY\cite{mccnifty2021icdm}, Gem\cite{gem2021icml}, RC-Explainer\cite{reinforced_explainer2023tpami}};
    \node (citation_5) [citation, right=of text_5]{CFLP\cite{cflp2022icml}, RFCGNN\cite{rfcgnn2023icdm}};
    \node (citation_6) [citation, right=of text_6]{GEAR\cite{gear2022wsdm}, GraphCFF\cite{graphcff2024}};
    \node (citation_7) [citation, right=of text_7]{CF-GNNExplainer\cite{cfgnnexplainer2022aistat}, MEG\cite{meg2021ijcnn}, CF$^2$\cite{cf2017nips}, CLEAR\cite{clear2022nips}};
    \node (citation_8) [citation, right=of text_8]{OBS and DBS\cite{obsdbs2021kdd}, RCExplainer\cite{rcexplainer2021nips}, 
    GNN-MOExp\cite{moexplanation2021icdm}, GCFExplainer\cite{GCFExplainer2023wsdm}, Banzhaf~\cite{banzhaf2024www}};
    \node (citation_9) [citation, right=of text_9]{EERM\cite{EERM2022iclr}, DIR\cite{DIR2022iclr}, inMvie\cite{inmvie2024fortune}, LiSA\cite{lisa2023cvpr}, GIL\cite{GIL2022nips}, BA-GNN\cite{BAGNN2022icde}, INL\cite{inl2023tois}};
    \node (citation_10) [citation, right=of text_10]{CAL\cite{CAL2022kdd}, CMRL\cite{cmrl2023kdd}, ICL\cite{icl2024aaai}, DisC\cite{DisC2022nips}, MoleOOD\cite{moleood2022nips}, OrphicX\cite{orphicx2022cvpr}, CANET\cite{canet2024www}, CIE\cite{cie2023cikm}, CI-GNN\cite{ci-gnn2023}, RC-GNN\cite{rcgnn2024}, CIGA\cite{ciga2022nips}, GALA\cite{gala2023nips}, CAF\cite{caf2023cikm}, LECI\cite{leci2023nips}, PNSIS\cite{pnsis2024}};
    \node (citation_12) [citation,  below=of citation_10, yshift = 0.665cm]{RGCL\cite{RGCL2022icml}, iMoLD\cite{imold2023nips}, CGC\cite{cgc2023www}, GCIL\cite{gcil2024aaai}, FLOOD\cite{flood2023kdd}};

    \draw [arrow] (title.east) -- ++(0.5,0) |- (s1_top.west);
    \draw [arrow] (title.east) -- ++(0.5,0) |- (s1_below.west);
    \draw [arrow] (s1_top.east) -- ++(0.5,0) |- (s2_top.west);
    \draw [arrow] (s1_top.east) -- ++(0.5,0) |- (s2_mid.west);
    \draw [arrow] (s1_top.east) -- ++(0.5,0) |- (s2_below.west);
    \draw [arrow] (s1_below.east) -- ++(0.5,0) |- (s2_4.west);
    \draw [arrow] (s1_below.east) -- ++(0.5,0) |- (s2_5.west);
    \draw [arrow] (s2_top.east) -- ++(0.5,0) |- (text_1.west);
    \draw [arrow] (s2_top.east) -- ++(0.5,0) |- (text_2.west);
    \draw [arrow] (s2_top.east) -- ++(0.5,0) |- (text_3.west);
    \draw [arrow] (s2_mid.east) -- ++(0.5,0) |- (text_4.west);
    \draw [arrow] (s2_mid.east) -- ++(0.5,0) |- (text_5.west);
    \draw [arrow] (s2_mid.east) -- ++(0.5,0) |- (text_6.west);
    \draw [arrow] (s2_below.east) -- ++(0.5,0) |- (text_7.west);
    \draw [arrow] (s2_below.east) -- ++(0.5,0) |- (text_8.west);
    \draw [arrow] (s2_4.east) -- ++(0.5,0) |- (text_9.west);
    \draw [arrow] (s2_4.east) -- ++(0.5,0) |- (text_10.west);
    \draw [arrow] (s2_5.east) --(citation_12.west);
    \draw [arrow] (text_1.east) -- ++(0.5,0) |- (citation_1.west);
    \draw [arrow] (text_2.east) -- ++(0.5,0) |- (citation_2.west);
    \draw [arrow] (text_3.east) -- ++(0.5,0) |- (citation_3.west);
    \draw [arrow] (text_4.east) -- ++(0.5,0) |- (citation_4.west);
    \draw [arrow] (text_5.east) -- ++(0.5,0) |- (citation_5.west);
    \draw [arrow] (text_6.east) -- ++(0.5,0) |- (citation_6.west);
    \draw [arrow] (text_7.east) -- ++(0.5,0) |- (citation_7.west);
    \draw [arrow] (text_8.east) -- ++(0.5,0) |- (citation_8.west);
    \draw [arrow] (text_9.east) -- ++(0.5,0) |- (citation_9.west);
    \draw [arrow] (text_10.east) -- ++(0.5,0) |- (citation_10.west);
\end{tikzpicture}
\caption{A detailed taxonomy of existing CIGNNs based on their empowered causal learning capability.}
\label{fig:taxonomy}
\end{figure*}
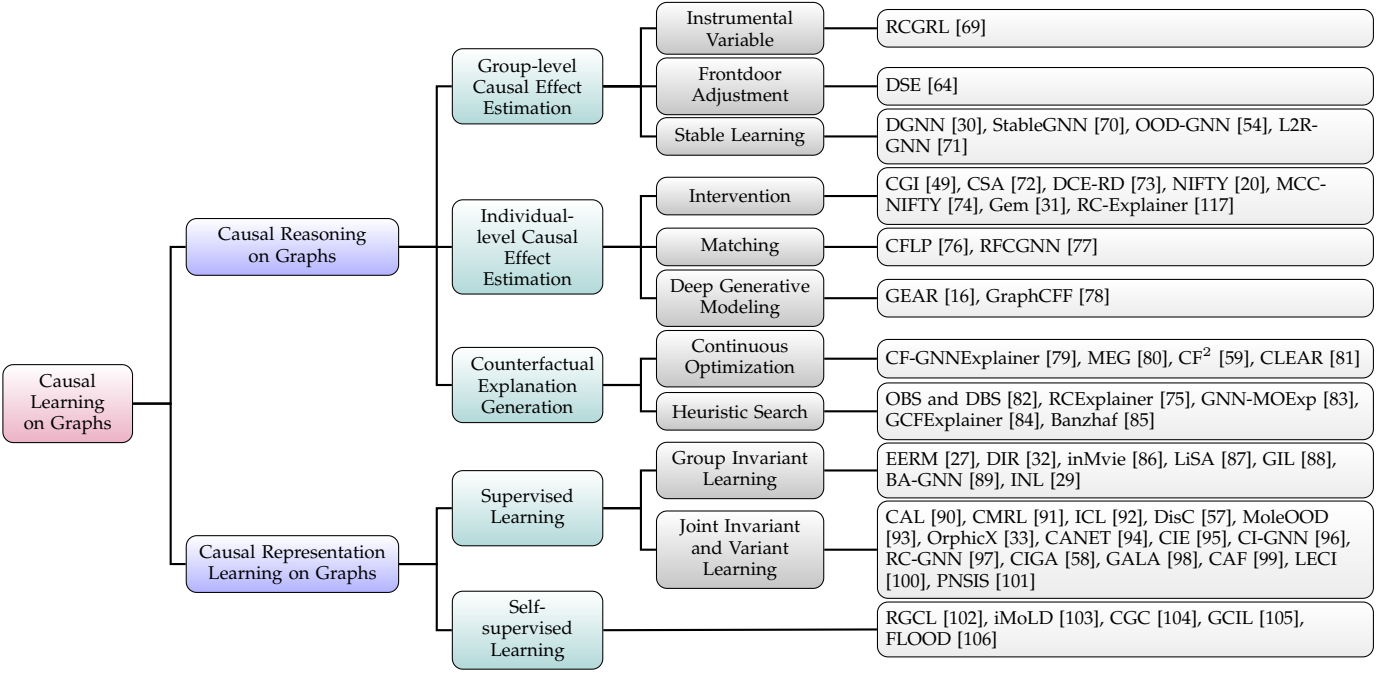

As discussed in \secref{sec:trust_risk}, the comprehension of underlying data causality is crucial for developing trustworthy GNNs.
In this section, we provide a systematic review of the existing CIGNNs within a taxonomy that highlights their diverse causal learning abilities.
The taxonomy is detailedly presented in \secref{subsec:taxonomy} and \figref{fig:taxonomy}.
We also summarize key characteristics of the reviewed works in \tabref{app:work_summary}.

\subsection{A Causal Learning Task Oriented Taxonomy} \label{subsec:taxonomy}
Motivated by a common observation that the enhanced trustworthiness of CIGNNs primarily stems from their superior causal learning capabilities, we categorize existing works based on the following two types of causal learning abilities: (i) causal reasoning, and (ii) causal representation learning.
\eat{Works falling into the first category quantify cause-effect relationships among different graph components as well as other outcomes of interest such as labels and model predictions, utilizing this type of causal knowledge to enhance the trustworthiness of GNNs.
On the other hand, works in the second category explore the potential of GNNs to learn causal representations directly from raw graph data and seamlessly integrate the CRL process into the standard problem-solving pipeline of GNN systems.}
As works on incorporating causal discovery for GNNs are limited, the discussion of this perspective is presented in \secref{sec:future} as part of future directions.
Within each category, we further introduce mainstream techniques that equip GNNs with the respective causal learning ability and further explain how these techniques contribute to mitigating different trustworthy risks.
It's worth noting that techniques proposed in works from different categories might be combined to enable GNNs to simultaneously handle multiple causal learning tasks, as exemplified by recent research~\cite{DSE2022,orphicx2022cvpr}.
We thereby classify these works based on the primary causal learning task they target to avoid confusion. 
This taxonomy is developed from a causality perspective, distinguishing it from viewpoints grounded in deep learning.
It is crucial as it not only clarifies the methodological diversity within the field but also reveals the subtle connections between different causal learning capabilities and GNN trustworthiness.

\subsection{Empowering Causal Reasoning on Graphs}
Empowering GNNs with causal reasoning capability allows for quantifying cause-effect relationships among graph components and other outcomes of interest, \eg{the label or the model prediction.}
Works falling into this category can be further subdivided based on the types of causal reasoning questions being addressed, including group-level causal effect estimation, individual-level causal effect estimation, and counterfactual explanation generation.

\subsubsection{Group-level Causal Effect Estimation on Graphs}
This type of question involves identifying the average causal effects of treatments on groups of nodes or graphs. The focused treatment varies according to the specific problem context, such as node attributes in node classification~\cite{debiasedgnn2022tnnls} or substructure alterations in graph classification~\cite{DSE2022,RCGRL2022}. 
Proper question formulation and resolution allow models to distinguish causation from spurious correlations among graph components and the target labels or GNN predictions, which is crucial for improving GNNs' trustworthiness as elaborated in \secref{sec:trust_risk}.

A major challenge in addressing this causal reasoning question lies in controlling confounders between the treatment and the outcome.
\eat{\TODO{Actually, here is a gap between ATE and confounder.}
\wz{I filled this gap by mentioning the harmfulness of confounder when introducing Fig. 1.}}
The commonly used backdoor adjustment is often impractical in graph learning scenarios for two main reasons: (i) the available domain knowledge is often insufficient to indicate the complete set of confounders~\cite{DSE2022}; (ii) even with all confounders observed, their high dimensionality due to the complexity of graph data renders confounder stratification unfeasible
~\cite{debiasedgnn2022tnnls,stablegnn2021}.
Along this line, three types of methods have been leveraged to circumvent the above challenge, namely \emph{instrumental variable}, \emph{frontdoor adjustment} and \emph{stable learning}.

\noindent
\textbf{Instrumental Variable (IV).}
IV is a powerful technique used to identify causal effects when there are observed or unobserved confounders~\cite{IVsurvey2022}. It involves identifying a variable that is correlated with the treatment variable but not with the confounders, thereby providing a way to isolate the causal effect of the treatment variable on the outcome variable from the confounding biases.
Assume we have i.i.d. observations $\{(X_i, Y_i )\}^n_{i=1}$ generated from an additive noise model $Y = f(X) + U$~\cite{ANM2008nips}, where the error term $U$ represents the unobserved confounding effects. Estimating the causal effect of $X$ on $Y$ requires learning a model $f_\theta(\cdot)$ to approximate $f(\cdot)$ without being affected by $U.$ 
One classical approach addresses this issue by selecting a set of IVs $Z$ satisfying conditional moment restrictions $\mathbb{E}[U|Z]=0$~\cite{IV2003ai}.
With these IVs, we have $\mathbb{E}[f(X)-Y|Z]=0$, hence we can learn $f_\theta(\cdot)$ by minimizing the conditional expectation loss of model output, \ie $\min_{\theta} \mathbb{E}[f_\theta(X)-Y|Z].$

Grounded in the idea of IV, Gao~\etal~\cite{RCGRL2022} developed the \emph{RCGRL} framework to maximize the causal effects of GNNs' output representations on the graph label, aiming to improve the OOD generalizability of GNNs on graph-level tasks.
In detail, the authors adopted a GNN $q_\phi(\cdot)$ to generate edge masking weights $\mathbf{Z} = q_\phi(\mathcal{G})$ as IVs, which satisfy
(i) $\mathbb{E}[C|q_{\phi^*}(\mathcal{G})]=0$ and (ii) $\mathbb{E}[X|q_{\phi^*}(\mathcal{G})]=X,$ where $C$ and $X$ denote the confounding and causal components mixed in the input graph $\mathcal{G}.$
The ideal $q_{\phi}(\cdot)$ can be theoretically obtained by optimizing the following objective,
\begin{align}
    \phi^* = \argmax_\phi \text{MI}(Y, f_{\theta^*} (r(\mathcal{G},q_\phi(\mathcal{G})))),  
    \label{formula:RCGRL_q}
\end{align}
where $r(\cdot)$ removes edges of $\mathcal{G}$ based on the IVs $\mathbf{Z},$ and $\text{MI}(\cdot,\cdot)$ denotes mutual information. 
Conditioning on the IVs $\mathbf{Z},$ the causal effect of the graph representation on the target label can be identified with unobservable confounders removed. This induces an IV-based training objective for the target model $f_\theta(\cdot)$,
\begin{align}
    \theta^* = \argmin_\theta \mathcal{L}(Y, f_{\theta} (r(\mathcal{G},q_\phi^*(\mathcal{G}))).
    \label{formula:RCGRL_f}
\end{align}
In practice, objective (\ref{formula:RCGRL_q}) and (\ref{formula:RCGRL_f}) can be optimized in an alternative manner~\cite{RCGRL2022}.

\noindent
\textbf{Frontdoor Adjustment.}
Frontdoor adjustment identifies interventional distribution $P(Y|do(T:=t))$ when unobserved confounders exist by harnessing a set of mediator variables $M$ that satisfies the frontdoor criterion. 
\begin{definition}[Frontdoor Criterion]
    A set of variables $M$ satisfies the frontdoor criterion if (i) $M$ blocks all directed paths from $T$ to $Y$, (ii) there are no unblocked backdoor paths from $T$ to $M$, and (iii) $T$ blocks all the backdoor paths from $M$ to $Y$.
\end{definition}
\begin{theorem}[Frontdoor Adjustment]
    If $(T, M, Y)$ satisfy the frontdoor criterion and positivity assumption~\cite{stat2causal2022ICM}, then
    \begin{align}
    \begin{split}
        P(Y| do(T:=t)) &=  \sum_m P(M=m | T=t) \times \\  
        &\sum_{t'} P(Y|M=m,T=t')P(T=t') \\ 
        &= \mathbb{E}_{P(M|t)} \mathbb{E}_{P(T')} \left[ P(Y|M, T') \right].
    \end{split}
    \label{eqn:frontdoor}
    \end{align}
\end{theorem}
\noindent
Intuitively, frontdoor adjustment isolates the causal effect of $T$ on $Y$ by summarizing the causal effect of $M$ on $Y.$

Enlightened by this causal identification approach, Wu~\etal~\cite{DSE2022} proposed the \emph{DSE} method for unbiased evaluation of post-hoc subgraph explanations on GNNs.
They estimated the causal effect of a given candidate subgraph $G_s$ on the model prediction via frontdoor adjustment, aiming to mitigate the confounding bias induced by the distribution shift from original graphs to candidate subgraphs, as illustrated in \secref{subsec:xgnn}.
Specifically, a Variational Graph Auto-Encoder (VGAE)~\cite{vgae2016} was adopted to generate mediate graph $\mathcal{G}_s^*$ from $\mathcal{G}_s$ to identify the causal effect as
\begin{align}
    P(\hat{Y}| do(\mathcal{G}_s:=G_s)) = \mathbb{E}_{P(\mathcal{G}_s^*|G_s)} \mathbb{E}_{P(\mathcal{G}_s')} \left[ P(\hat{Y}|\mathcal{G}_s^*, \mathcal{G}_s') \right].
\end{align}
Such $\mathcal{G}_s^*$ will naturally satisfy conditions (ii) and (iii) of the frontdoor criterion.
As for condition (i), 
the authors proposed a contrastive learning module~\cite{infonce2018} to promote $\mathcal{G}_s^*$ to capture more class-discriminative graph information so that the causal effect of candidate graph $\mathcal{G}_s$ is completely mediated by $\mathcal{G}_s^*$. 
Furthermore, the statistical estimands in \equref{eqn:frontdoor} can be well estimated, 
(i) Monte Carlo sampling based on the VGAE for estimating the expectation \wrt $P(\mathcal{G}_s^*|\mathcal{G}_s)$, (ii) traversing all candidate subgraphs for estimating the expectation \wrt $P(\mathcal{G}_s')$, and (iii) feeding the mediator graphs generated from candidate graph $\mathcal{G}_s'$ into the target GNN for estimating $P(\hat{Y}|\mathcal{G}_s^*, \mathcal{G}_s').$
With the above efforts, $P(\hat{Y}|do(\mathcal{G}_s = G_s))$ can be identified to facilitate unbiased subgraph evaluation.

\noindent
\textbf{Stable Learning.}
Stable learning methods estimate direct causal effects of high dimensional features on the label by learning to reweight the samples to achieve mutual independence among feature variables~\cite{stable_2018,stable_misspecification_2020,dvd2020kdd,deepstable2021cvpr}. This independence ensures that no feature creates a backdoor path from any target feature to the label, implying that any observed correlation between a feature and the label is a result of direct causation. Consequently, one can train a correlation-based model to identify direct causal features for generalizable and interpretable prediction~\cite{stable2022nature}.

\begin{figure}
    \centering
    \includegraphics[width=0.98\linewidth]{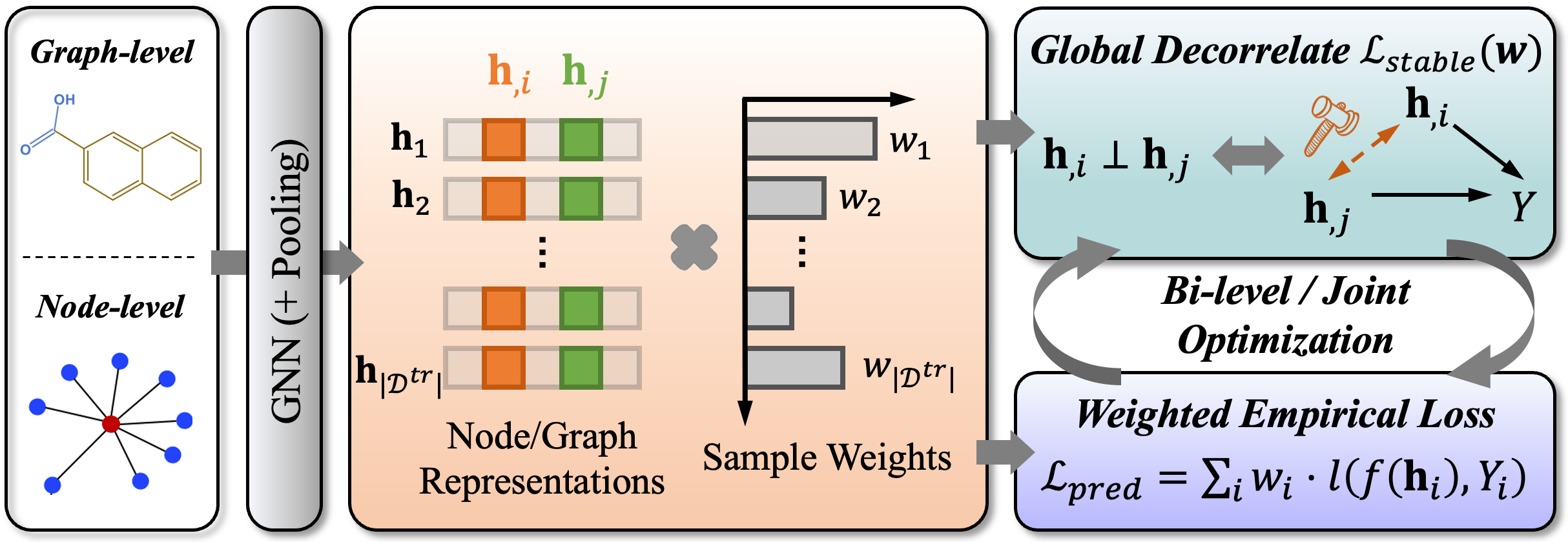}
    \caption{General pipeline of integrating stable learning into GNNs. 
    }
    \label{fig:stable_learning_copied}
\end{figure}

The idea of stable learning can be adapted to automatically identify causal graph components for learning TGNNs. 
Though potentially promising, its direct application in input graph space poses challenges due to the high dimensionality of raw graph features and the unmeasurable nature of high-level, causally significant semantics.
To this end, existing works instead estimate stable causal effects between the latent node/graph representations and the label. This is based on the assumption that spurious correlations in input graph space can be inherited into the latent space of GNNs~\cite{debiasedgnn2022tnnls,fairness4gnn2022survey}.
The general framework of incorporating stable learning for GNNs is illustrated in \figref{fig:stable_learning_copied}. It consists of three primary steps: (i)~Encoding input nodes/graphs into latent representations, (ii)~learning sample weights to decorrelate latent features based on stable learning regularizers, and (iii)~training the GNN on reweighted samples.

In practice, great efforts have been devoted to designing appropriate regularizers for step (ii).
\emph{DGNN}~\cite{debiasedgnn2022tnnls} proposed a Differentiated Variable Decorrelation (DVD) objective for node classification tasks 
to decorrelate all dimensions of latent node representations,
\begin{align} \label{eqn:DVD_loss}
    \mathcal{L}_{\text{DVD}}(\mathbf{w}) =  \sum_{i\not=j} \alpha_i \alpha_j \lVert \mathbf{h}_{,i}^T \Sigma_\mathbf{w} \mathbf{h}_{,j} / n - \mathbf{h}_{,i}^T \mathbf{w}/n \cdot \mathbf{h}_{,j}^T \mathbf{w}/n \rVert _2^2,
\end{align}
where 
$\mathbf{w}\in \mathbb{R}^n$ denotes the learnable sample weights, $\Sigma_\mathbf{w} = \text{diag}(w_1,...,w_n),$ $\alpha_j$ denotes the regression coefficients for $j$-th feature variable.
In order to better decorrelate nonlinearly correlated feature variables, \emph{StableGNN}~\cite{stablegnn2021} and \emph{OOD-GNN}~\cite{oodgnn2022tkde} incorporated a Hilbert-Schmidt Independence Criterion (HSIC) based objective~\cite{hsic2005,deepstable2021cvpr} to globally decorrelate the graph representations output by graph pooling layers,
\begin{align} \label{eqn:hsic_loss}
    \mathcal{L}_{\text{HSIC}}(\mathbf{w}) = \sum_{1\le i < j \le d} \lVert \hat{\mathbf{C}}^{\mathbf{w}}_{\mathbf{h}_{,i}^\mathcal{G},\mathbf{h}_{,j}^\mathcal{G}} \rVert _F^2,
\end{align}
where 
$\hat{\mathbf{C}}^{\mathbf{w}}_{\mathbf{h}_{,i}^\mathcal{G},\mathbf{h}_{,j}^\mathcal{G}}$ denotes the weighted partial cross-covariance matrix between $\mathbf{h}_{,i}^\mathcal{G}$ and $\mathbf{h}_{,j}^\mathcal{G}$ in the Repreducing Kernal Hilbert Spaces (RKHS)~\cite{hsic2005}. 
Moreover, \emph{L2R-GNN}~\cite{L2R-GNN2024aaai} extends the work~\cite{dvd2020kdd} to decorrelate only the inter-dependencies among latent feature clusters, reducing training variance and improving computational efficiency.

\begin{tcolorbox}[breakable, colback=gray!5!white, colframe=gray]
\textit{
\textbf{Discussion.}
\eat{
First, the IV and frontdoor adjustment approaches treat the whole input (sub)graph as a treatment variable and can estimate the treatment effect even when the confounders are unobservable. 
In contrast, while stable learning cannot guarantee its performance when unobservable confounders exist, it can measure the causal effect of each feature (cluster) from high-dimensional features simultaneously, which provides more fine-grained causal knowledge to the GNN.
}
Both IV and stable learning have been leveraged to improve the generalizability of GNNs, while frontdoor adjustment has been applied primarily to address biases in post-hoc GNN explanations. 
These approaches also have potential to enhance other GNN trustworthiness aspects. For instance,
stable learning could boost GNN's inherent interpretability, allowing for decisions to be interpreted by the causal effects of varied graph components.
Frontdoor adjustment offers a promising way to mitigate the confounding bias harming the GNN generalizability.
However, a common limitation of these approaches is their inability to improve the GCF of GNNs, as the group-level causal effect does not precisely reflect the counterfactual outcome for each individual~\cite{seventools2019pearl}. 
More technical discussions are provided in \appref{app:gte_discuss}.
}
\end{tcolorbox}


\subsubsection{Individual-level Causal Effect Estimation on Graphs}
This type of question delves into the causal effects of factual or hypothesized interventions exerted on the individual node or (sub)graph. Such interventions often involve modifications to node features or localized structural changes, which influence the generation of the label, model prediction or representation of the individual of interest. 
By estimating the interventional or counterfactual outcome for each individual, the GNN can distinguish genuine causal effects from mere correlations, thereby increasing the trustworthiness of model predictions at the individual level.
In the following, we introduce three established methods for estimating this causal reasoning question: \emph{intervention}, \emph{matching}, and \emph{generative modeling}.

\noindent
\textbf{Causal Intervention.}
The causal intervention has been utilized to assess the individual-level causal effects of graph components on \emph{GNN outputs $\hat{Y}$}. 
Works falling into this category first intervene on different elements of the graph, \eg{adding or deleting edges}, and then feed each intervened graph into the target GNN to generate an interventional output. 
This output is essentially the counterfactual outcome of the intervention, since both the graph and the GNN's inference mechanism remain unchanged under control or intervention conditions.
Therefore, by contrasting the factual and counterfactual outputs, the causal effect on GNN outputs can be estimated. 
In practice, several ways have been developed to incorporate such causal knowledge to boost the GNN trustworthiness.

First, the causal effect can be directly adopted as the criterion to generate more reliable instance-level graph explanations on the behaviors of the target GNN. 
For each candidate subgraph $\mathcal{G}_s,$ it corresponds to an intervention on the original input graph $\mathcal{G},$ \ie{deleting the complement graph $\mathcal{G} \setminus \mathcal{G}_s.$}
The ITE of this intervention on certain outcome variables can thus be used to evaluate the contribution of $\mathcal{G}_s$ to the model prediction.
The specific selection of the outcome variable may differ according to the forms of GNN outputs.
\emph{Gem}~\cite{gem2021icml} defined the ITE as the change in model error before and after intervention, while \emph{RC-Explainer}~\cite{reinforced_explainer2023tpami} measured the information gain caused by the intervention.
In practice, both works proposed to select the edge step-by-step to reduce the expensive computation overhead of searching over the whole subgraph space. 
Differently, Gem adopted greedy search expert-curated graph rules, followed by a VGAE to abstract the explaining process, while RC-Explainer took advantage of Reinforcement Learning~(RL)~\cite{RL4graphgen20218nips} to learn optimal edge selection strategies.

Second, the estimated ITEs can trigger active calibrations of the GNN inference process to improve its generalizability. 
Feng~\etal~\cite{CGI2021sigir} validated the efficacy of this idea through the development of a \emph{CGI} model tailored for node-level tasks. 
Specifically, after each message-passing iteration within the GNN, the GNN is forced to produce the prediction result $f(X_u, do(\mathcal{N}_u = \emptyset))$ relying solely on the intrinsic attributes of the node.
This counterfactual prediction is shielded from the potential spurious correlations between the target node and its neighbors, serving as a robust alternative for the final prediction result. 
Furthermore, the authors trained a binary classifier to determine the final prediction from the original and counterfactual predictions. This choice model uses the ITE of the neighbor nodes as well as other factors such as the prediction confidence, which reflect the impacts of varying local structures on model predictions. Integrating this choice model with the GNN thus facilitates more adaptive and generalizable inferences.

Third, the obtained ITEs can be utilized to regularize the inference mechanism of GNNs towards trustworthiness.
\emph{CSA}~\cite{csa2023ijcai} guide the message-passing process in attention-based GNNs~\cite{gat,general_gat} by minimizing the disparity between each node's label and the ITE of the attention scores. This promoted the alignment between attention and causal relations, facilitating generalizable node classification.
\emph{DCE-RD}~\cite{DCERD2023kdd} trained the GNN on diversified subgraphs of the input graph, intervening on which causes the maximal changes in the GNN prediction, to gain a multi-view causal understanding of the graph generation process, effectively boosting GNN generalizability and interpretability.
\emph{NIFTY}~\cite{nifty2021uai} and \emph{MCCNIFTY}~\cite{mccnifty2021icdm} explicitly conducted causal interventions on the sensitive attributes of each node and generated counterfactual node representations through the GNN accordingly. Then, a regularizer is introduced to maximize the similarity among node representations from different views, which promotes the GCF of the GNN.

\noindent
\textbf{Matching.}
Due to limited access to the graph generation process, obtaining counterfactual graphs or labels for specific interventions is challenging, complicating the estimation of individual-level causal effects. 
Matching offers a viable solution by pairing each treated individual with a controlled individual that has similar covariates.
The outcome of this controlled counterpart can then be used as an approximated counterfactual~\cite{nnm1973biometrics,propensity1983biometrika,balancing_counterfactual_regression2016icml}, as the potential confounding or mediation effects from the covariates are effectively mitigated.

Building on this concept, \emph{CFLP}~\cite{cflp2022icml} augmented the generalizability of GNNs in link prediction tasks by exploring the causal effect of graph structure on link existence.
Specifically, for each node pair $(u, v)$, whether the nodes belong to the same cluster (\eg{community}) was chosen as the treatment $T_{uv} \in \{0,1\},$ which also became part of the sample features, and the link existence $A_{uv} \in \{0,1\}$ is the outcome of interest.
The counterfactual counterpart of $(u,v)$ was approximated by the nearest node pair $(a,b)$ with treatment $T_{ab}=1-T_{uv}.$ 
With both factual and counterfactual samples, the authors followed the training strategy from work~\cite{balancing_counterfactual_regression2016icml} to encourage the GNN to produce node representations that absorb the ITE of $T_{uv}$ on $A_{uv},$ thereby generalizing to varied structural contexts.
Similarly, \emph{RFCGNN}~\cite{rfcgnn2023icdm} curated counterfactual samples for each node as the most similar nodes with the same label but flipped sensitive attributes in their neighborhoods. By aligning both node representations of both samples, the GNN avoided capturing the correlation between sensitive attributes and the node label, thereby achieving improved GCF.

\noindent
\textbf{Deep Generative Modeling.}
Deep generative modeling, known for its ability to replicate complex distributions and produce realistic data samples, has proven effective in learning the SCM that formalizes the data generation processes~\cite{cgmsurvey2023}. 
By encoding causal structures into graph generative models, \eg{VGAE as illustrated in \figref{fig:vae+scm}, researchers can simulate interventions on individual nodes/graphs, generate corresponding counterfactual graphs while adhering to observed data distributions and further identify the causal effects of these interventions.
\begin{figure}[ht]
    \centering
    \includegraphics[width=1.0\linewidth]{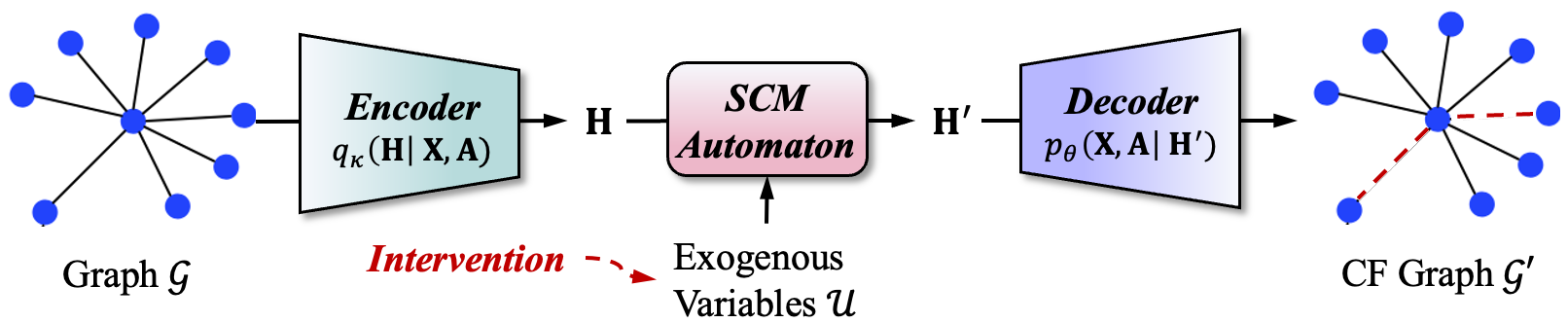}
    \caption{An illustrative framework integrating SCM into VGAE.
    }
    \label{fig:vae+scm}
\end{figure}
}

\emph{GEAR}~\cite{gear2022wsdm} utilized this approach to enhance GCF in scenarios where sensitive attributes of a node causally affect both the node and its neighbors. Mirroring the strategy used by NIFTY~\cite{nifty2021uai}, the GNNs were trained to maintain similarity between node representations under perturbations to sensitive attributes. 
A primary challenge was the generation of counterfactual graphs for different sensitive attribute interventions. 
To overcome this, a fairness-constrained VGAE~\cite{vgae2016} was employed to embed the SCM into latent node representations $\mathbf{H}$, in the meantime ensuring they were statistically independent of the sensitive attributes $\mathbf{S}$. This independence was enforced through an adversarial learning regularizer $\mathcal{L}_d = \mathbb{E}_{\mathbf{S}}[- \log(P(\mathbf{S}|\mathbf{H}))],$
which aimed to make $\mathbf{S}$ unpredictable from $\mathbf{H}$. The trained VGAE enables the SCM to be rerun to reconstruct multiple counterfactual graphs based on varied intervened sensitive attributes $\mathbf{S}'$ as well as the node representations $\mathbf{H}$.
\emph{GraphCFF}~\cite{graphcff2024} extended GEAR to accommodate hidden confounders between each node and its neighbors. An identifiable Gaussian mixture based generative model~\cite{gaussiandgm2022nips} was adopted to approximate the prior distribution of hidden confounders and simultaneously encode the SCM for counterfactual graph generation.

\begin{tcolorbox}[breakable, colback=gray!5!white, colframe=gray]
\textit{\textbf{Discussion.}
\eat{
Causal intervention approaches estimate the individual-level causal effects of graph components on model predictions, facilitated by the capability to repeatedly execute the GNN inference mechanism. In contrast, matching and deep generative methods measure causal effects among graph components or labels by approximating counterfactual outcomes of hypothetical graph interventions.
}
Causal intervention has been used to improve all three aspects of trustworthiness owing to its straightforward implementation and the ability to execute the GNN inference mechanism repeatedly. 
In contrast, matching improves only the generalizability and fairness of GNNs, as it cannot flexibly evaluate the causal effects of varied graph components on predictions, which is crucial for enhancing explainability. 
Additionally, deep generative modeling has advanced GNN fairness by automating the encoded SCM for causal effect estimation. This method also holds the potential to enhance generalizability and explainability, using causal effects in ways similar to those in the other two categories of approaches.
More technical discussions can be found in \appref{app:ite_discuss}.}
\end{tcolorbox}

\subsubsection{Graph Counterfactual Explanation Generation}
\begin{figure}[ht]
    \centering
    \includegraphics[width=1.0\linewidth]{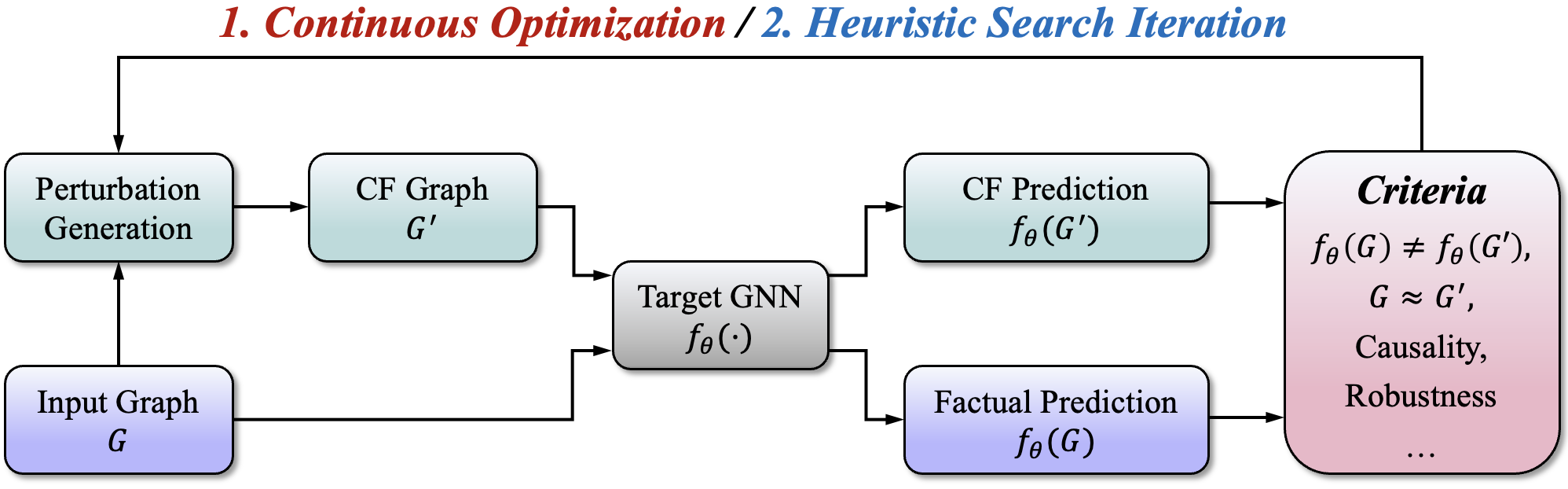}
    \caption{A general pipeline of generating GCEs for a target GNN. 
    }
    \label{fig:gce_pipeline}
\end{figure}

The Graph Counterfactual Explanation (GCE) answers the counterfactual question of the type ``what is the minimum perturbation required on the input graph sample to change the model prediction?''.
\begin{definition}[Graph Counterfactual Explanation, GCE~\cite{gce_survey2022}]
    Let $f$ be a prediction model that classifies $\mathcal{G}$ into a class $c$ from a set of classes $\mathcal{C}$. Let $\mathbb{G}'$ be the set of all possible counterfactual examples and $S(\cdot, \cdot)$ be a graph similarity measure. Then, we define the set of counterfactual explanations $\mathcal{E}_f(\mathcal{G})$ w.r.t. $f$ as
    \begin{align}
        & s(c', \mathcal{G}) := \max_{\mathcal{G}'\in \mathbb{G}', \mathcal{G} \not= \mathcal{G}'} \{ S(\mathcal{G}, \mathcal{G}') | f(\mathcal{G}') = c' \}, \\
        & \mathcal{E}_f(\mathcal{G}) := \mathop{\bigcup}_{c'\in C \setminus \{c\}} \{\mathcal{G}'\in \mathbb{G}' |\mathcal{G} \not= \mathcal{G}', S(\mathcal{G},\mathcal{G}')=s(c',\mathcal{G}) \}.
    \end{align}
    \label{def:gce}
\end{definition} 
\noindent
Compared with factual graph explanation generation where we measure the causal effects of candidate graphs on model predictions and find the \emph{sufficient} explanation, GCE generation focuses on finding the \emph{necessary} interventions on graphs that have a fixed causal effect (changing model prediction). 
As shown in \figref{fig:gce_pipeline}, the generation process can be viewed as a loop where candidate GCEs are iteratively updated using specific criteria.
Based on the updating type, we review two categories of representative methods below. 

\noindent
\textbf{Continuous Optimization.}
This type of method generates GCEs by learning a parameterized perturbation \wrt tailored objectives consistent with \defref{def:gce}. 
\emph{CF-GNNExplainer}~\cite{cfgnnexplainer2022aistat} generated node-level GCEs within each node's computational graph~\cite{gnnexplainer2019nips} by simultaneously maximizing the prediction difference and the similarity. 
\emph{MEG}~\cite{meg2021ijcnn} generates GCEs for GNNs designed for compound prediction within a multi-objective RL framework. However, its dependency on domain-specific expertise constrains its adaptability to more general GNNs.
\emph{CF$^2$}~\cite{cff2022www} combines factual and counterfactual reasoning objectives to learn graph explanations that are concurrently necessary and sufficient for GNNs' predictions in both node and graph classification tasks. Given the inevitable trade-off between sufficiency and necessity, the learned graph explanations often do not represent the optimal GCEs~\cite{gce_survey2022}.
\emph{CLEAR}~\cite{clear2022nips} and \emph{RSGG-CE}~\cite{rsggce2024aaai} leverages VGAE to capture the latent distribution of GCEs. 
Differently, CLEAR inferred an observed auxiliary variable within the VGAE framework to ensure the GCEs preserve inherent causality within the graph data, while RSGG-CE introduced a discriminator network to strengthen the similarity between the GCEs and the input graph.

\noindent
\textbf{Heuristic Search.} 
This category of methods searches for GCEs directly in the discrete graph space in an exploratory manner. Instead of updating candidate GCEs via gradient backpropagation, heuristic strategies are adopted to select, modify or search for candidates based on predefined criteria.
\eat{

}
\emph{OBS and DBS}~\cite{obsdbs2021kdd} first perturbed the input graph $\mathcal{G}$ randomly or based on the edge existence frequencies until a counterfactual graph $\mathcal{G}'$ is obtained. Then, some perturbations were rolled back such that the similarity $S(\mathcal{G}, \mathcal{G}')$ increases while the counterfactual property is maintained.
\emph{GNN-MOExp}~\cite{moexplanation2021icdm} combined DFS and BFS to look for optimal factual graph explanations that not only have the maximum faithfulness but also contain a GCE. However, this method can hardly ensure the minimality of $S(\mathcal{G}, \mathcal{G}')$.
Differently, Huang and Kosan~\etal~\cite{GCFExplainer2023wsdm} defined a new problem of model-level~\cite{xgnn2020kdd} GCE generation, \ie finding a small \emph{set} of representative GCEs that explains model predictions on all input graphs. 
They proposed \emph{GCFExplainer} with three key steps, (i) constructing a meta-graph $\mathcal{G}_m$ with each node representing an edited graph and each edge representing a graph edit operation, (ii) leveraging vertex-reinforced random walks~\cite{vrrw1992} on $\mathcal{G}_m$ to identify diverse GCE candidates, and (iii) iteratively select the candidates that bring maximal gain of coverage metric~\cite{GCFExplainer2023wsdm} over the already selected GCE set, aiming to maximize $S(\mathcal{G}, \mathcal{G}').$ 

Some works particularly emphasize the robustness of the search algorithms.
\emph{RCExplainer}~\cite{rcexplainer2021nips} designed a heuristic method to identify a GNN's decision regions that are shaped by a set of linear transformations in the downstream predictor of the GNN. 
These decision regions not only help meet the basic GCE constraints, but also allow for enhancing robustness by pushing the representations of the input graph and its GCE to be far away from the decision boundaries.
An assumption of RCExplainer is the accessibility of the latent representations output by the GNN encoder. 
Chhablani and Jain~\etal~\cite{banzhaf2024www} proposed a game-theoretic search algorithm based on thresholded \emph{Banzhaf} values, which exhibits great efficiency and robustness when measuring the impact of individual edges and generating GCEs.

\eat{
\noindent
\textbf{Discussion.}
Continuous optimization based methods leverage gradient information to iteratively refine counterfactual explanations, which is highly efficient and scalable but can be sensitive to the choice of optimization parameters and might converge to local minima \TODO{addref}. 
On the other hand, heuristic search based methods use a trial-and-error approach to explore the graph space for possible GCEs, which typically are more flexible and have mechanisms to prevent from being stuck in local minima \TODO{addref}. However, they often require more computational resources and have less predictable performance, as their success heavily depends on the design of the heuristic and its parameters. 
Both methods have their merits and limitations, and the choice between them may depend on the specific requirements of the task, such as the need for scalability or precision.

The robustness of ... is under-explored.
Continuous GCE can also benefit other trustworthiness aspects~\cite{cgc2023www,pnsis2024}

}

\begin{tcolorbox}[breakable, colback=gray!5!white, colframe=gray]
\textit{
\textbf{Discussion.}
Both types of approaches have shown efficacy in providing necessary post-hoc explanations for GNNs across a range of applications.
Beyond explainability, GCE generation ability could also improve other trustworthiness aspects of GNNs.
For example, training GNNs to predict both an input graph and its similar GCE counterpart could strengthen the identification of causal components for better generalizability~\cite{cgc2023www,ci-gnn2023}. 
If the GCE is generated particularly from perturbing sensitive attributes, the GNN can gain a more intricate understanding of the causal impact of sensitive attributes, thereby boosting GCF.
Specifically, generating GCEs by altering sensitive attributes could enhance the GCF of GNNs.
The continuous GCE generation can be more seamlessly integrated with primary GNN training objectives to realize these goals, whereas heuristic search might necessitate efficient two-stage combination designs.
More technical discussions can be found in \appref{app:gce_discuss}.
}
\end{tcolorbox}

\eat{
\subsubsection{Discussion on Causal Reasoning Capability}
\TODO{Compared with individual-level causal effect estimation, GCE ...}

Similar to group-level causal effect estimation, individual-level causal effect estimation can help eliminate spurious correlations and improve GNNs' generalizability and explainability.
\wz{One speciality of this subcategory:} Moreover, the primary focus in individual-level causal effect estimation involves inferring the counterfactual outcomes for each node or graph instance. This particular focus enables the training of GNNs that are counterfactually fair by incorporating both factual observations and counterfactual outcomes.

Third, these approaches fall short in improving the GCF of GNNs, as the group-level causal effect is unable to precisely reflect the counterfactual outcome of each individual~\cite{seventools2019pearl}.

}

\subsection{Empowering Causal Representation Learning on Graphs}
Empowering GNNs with CRL capability aims to unravel the latent causal structures within complex graph data to improve the trustworthiness of the GNN reasoning process.
GNNs are uniquely capable of capturing non-linear relationships, which allows them to distill high-level causal variables into sophisticated latent representations. Advanced learning strategies are further devised to maintain the causal structures among these causal representations.
In this following, we begin with an overview of invariant learning, which has sparked numerous graph CRL studies in current literature. Then we classify existing works into supervised or self-supervised methods based on their learning strategies, providing a detailed elaboration on their innovations.


\subsubsection{Basics of Invariant Learning}
\label{subsubsec:inv_learning}
\eat{\TODO{risk or loss? unify!} \wz{changed to loss except for specific method names like Invariant Risk Minimization, Heterogeneous Risk Minimization.}}
We begin with an important assumption indicating the invariance of causal mechanisms.
\begin{assumption}[Invariance of Causality~\cite{buhlmann2020invariance}] \label{ass:inv_ass}
    The SCM $\mathcal{M} = (\{Y^e, \mathcal{PA}_Y^e\} , U_Y^e, {f_Y}, P_{U_Y^e} )$ such that
    \begin{align}
        Y^e := f_Y (\mathcal{PA}_Y^e , U_Y^e ), \ U_Y^e \ \bot \  \mathcal{PA}_Y^e, \ \mathcal{PA}_Y^e \subset \mathcal{X}^e
    \end{align}
    remains the same across any data environment $e \in supp(\mathcal{E}_{all})$, that is, $P_{U_Y^e}$ remains same across all environments. Here each $e$ consists of a feature set $\mathcal{X}^e$ and a corresponding label set $\mathcal{Y}^e.$
\end{assumption}


\noindent
This assumption indicates that causal relations between the target variable $Y$ and its direct causal parents are invariant. 
In contrast, Peters~\etal~\cite{peters2016invariant_causal} proposed Invariant Causal Prediction (ICP) to investigate under what circumstances could 'invariance' infer the 'causality' for the first time.
However, ICP is limited in the causal relations between raw features and the target label.

Invariant Risk Minimization (IRM)~\cite{IRM2019} first extended ICP to handle latent causal mechanisms with the help of representation learning, and inspired a series of invariant learning methods afterwards.
The invariance assumption of IRM is presented below.
\begin{assumption}[IRM’s Invariance Assumption~\cite{IRM2019}] \label{ass:irm_inv_ass}
    There exists a data representation $\Phi(X)$ such that $\mathbb{E}_{(X,Y)\in e}[Y|\Phi(X)] = \mathbb{E}_{(X,Y)\in e'}[Y|\Phi(X)]$, $\forall e, e' \in \mathcal{E}_{tr}$, where $\mathcal{E}_{tr}$ denotes the available training environments.
\end{assumption}
\noindent
The representation $\Phi(X)$ can characterize a high-level causal variable that has a direct causal effect on the label.
To find such $\Phi(X),$ IRM formulates a bi-level optimization problem based on the fact that ground-truth $\Phi(X)$ corresponds to an invariant predictor $w(\cdot)$ optimal across $\mathcal{E}_{tr}$,
\begin{align}
    \begin{split}
        & \min _{\Phi, w} \sum_{e \in \mathcal{E}_{tr}} \mathcal{L}^e(w \circ \Phi(X), Y) \\
        & \text{s.t.  } w \in \bigcap_{e \in \mathcal{E}_{tr}} \argmin_{\bar{w}} \mathcal{L}^e(\bar{w} \circ \Phi(X), Y). 
    \end{split}
    \label{eqn:bilevel_irm}
\end{align}

\eat{
Invariant learning~\cite{IRM2019} offers us a roundabout way to capture invariant causal relations between features and labels by training the prediction model $f=w \circ \Phi$ to achieve good performance on diverse environments, and this idea can be directly adopted in GRL given multiple training environments. 
It is theoretically guaranteed that such a global optimal model $f$ must satisfy that $\Phi(G)$ captures all causal features, hence it perfectly matches the need of ensuring the causality of the invariant representation $\mathbf{h}^I$ in Eqn.(\ref{eqn:inv_repre}).
}
\noindent
where $\mathcal{L}^e(\cdot, \cdot)$ is the empirical loss in environment $e.$
Many works then transformed the constraint in problem (\ref{eqn:bilevel_irm}) to an invariant regularizer $\Psi_{\text{inv}}(\{\mathcal{L}^e:e \in \mathcal{E}_{tr}\})$~\cite{oodsurvey2021}, which measures the variations of the model across $\mathcal{E}_{tr}$.
We list the ones that have been adopted in CIGNN literature in \tabref{tab:inv_regs}.
\begin{table}[!ht]
    \centering
    \caption{Three commonly adopted invariant regularizers.}
    \begin{tabular}{c|c}
    \toprule
        \textbf{Work} & \textbf{Invariant Regularizer} \\ 
    \midrule
        IRM~\cite{IRM2019} & $\Psi_{\text{IRM}} = \sum_{e \in \mathcal{E}_{tr}} \| \nabla_{w|w=1.0} \mathcal{L}^e (\cdot,\cdot) \|^2$ \\ 
        V-REx~\cite{rex2021icml} & $\Psi_{\text{V-REx}} = \text{Var}_{e \in \mathcal{E}_{tr}} \left[ \mathcal{L}^e(\cdot,\cdot) \right]$ \\ 
        IGA~\cite{IGA2020} & $\Psi_{\text{IGA}} = \text{trace}(\text{Var}_{e \in \mathcal{E}_{tr}}\left[ \nabla_\theta \mathcal{L}^e(\cdot,\cdot) \right])$\\ 
    \bottomrule
    \end{tabular}
    \label{tab:inv_regs}
\end{table}

\subsubsection{Supervised Causal Representation Learning}
Supervision signals from downstream tasks provide explicit feedback to guide the learning process of GNNs and shape the representation space.
However, CRL requires a deeper understanding of data distribution heterogeneity~\cite{towards_causal2021}, necessitating more targeted supervision utilization. IRM harnesses the supervision signals from multiple data environments to provide guidance from both accuracy and stability perspectives, which has sparked a series of works to learn invariant graph representations via tailored model architectures or optimization strategies.

\begin{figure}
    \centering
    \includegraphics[width=1.0\linewidth]{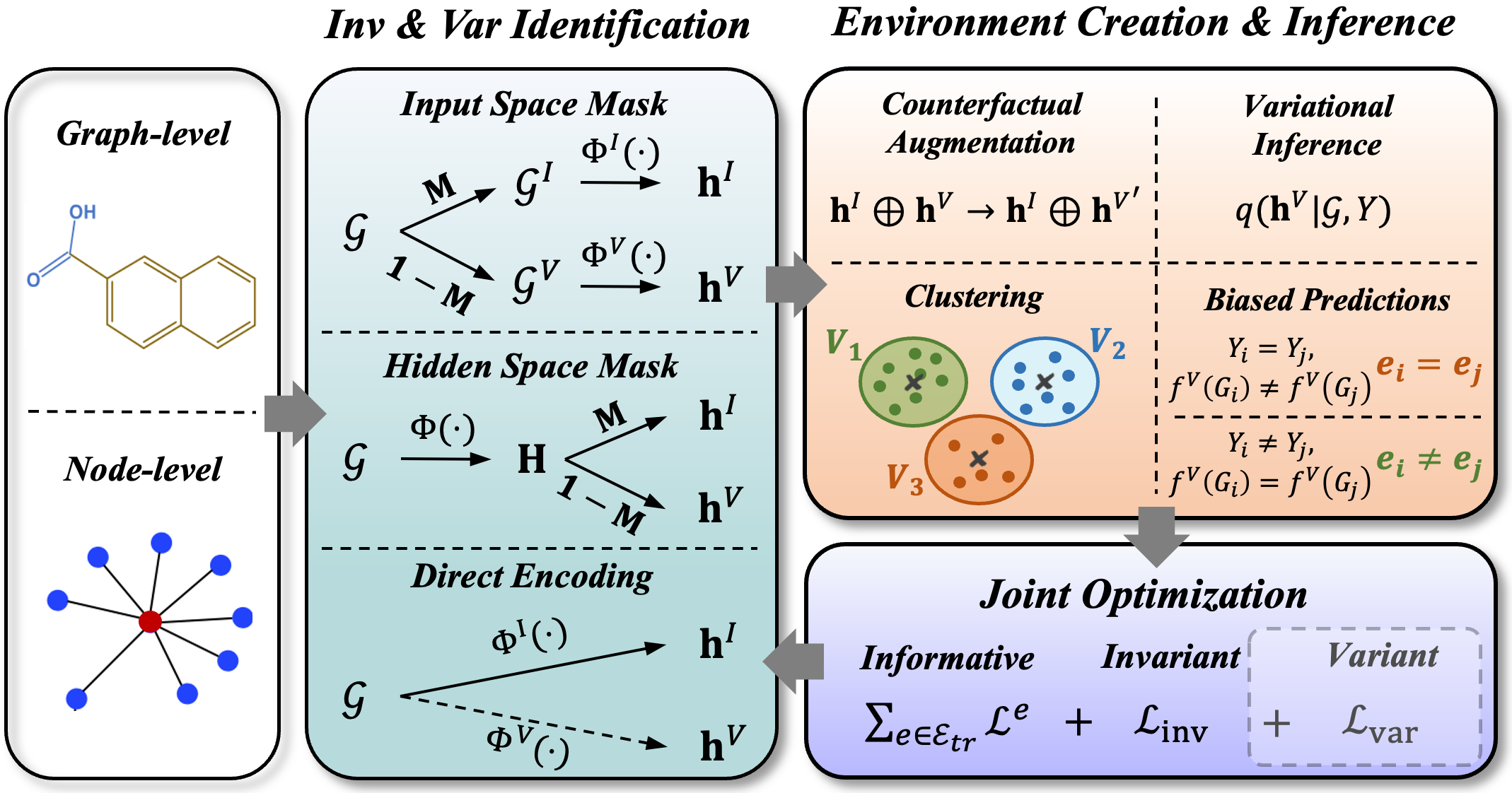}
    \caption{The general pipeline of supervised CRL on graphs.}
    \label{fig:gcrl}
\end{figure}

\noindent
\textbf{A Unified Framework.}
A fundamental challenge of conducting graph CRL is the lack of explicit indicators of the data heterogeneity in practical applications~\cite{DIR2022iclr}. To this end, mainstream methods endeavor to create diverse data environments solely based on the training dataset~\cite{DIR2022iclr,EERM2022iclr,GIL2022nips}.
These works follow a unified pipeline as illustrated in~\figref{fig:gcrl}, which consists of three key steps:
(i)~\emph{Invariant and variant representation identification:} 
Classical GNNs are adopted to encode input graphs into invariant and variant representations $\mathbf{h}^I$ and $\mathbf{h}^V.$ These two types of representations complement each other, representing the latent factors $I$ and $V$ that generate the observed graph as illustrated in \figref{fig:graph_DGP_ood+exp}. Structure masks $\mathbf{M}^a \in [0,1]^{|\mathcal{V}|\times |\mathcal{V}|}$ and feature masks $\mathbf{M}^x \in [0,1]^{|\mathcal{V}| \times d}$ are often generated in input or latent space to enhance the identification of both factors;
(ii)~\emph{Environment creation or inference:} 
Latent data environments $\mathcal{E}_{tr}$ are typically created or inferred from those variant components due to the unknown environment labels in most real-world scenarios.
However, there are still works assuming the availability of ground-truth environment labels for exploring more general representation identification results~\cite{leci2023nips,pnsis2024};
(iii)~\emph{Joint optimization:} 
Both $\mathbf{h}^I$ and $\mathbf{h}^V$ are optimized to preserve predictivity or causal semantics with downstream labels or other specific invariance or variance enforcing regularizers across different environments. The overall objective can be generally summarized below, 
\begin{align}
    \min_{\theta,\psi} \mathbb{E}_{e\in\mathcal{\hat{E}}_{tr}}\mathcal{L}^{e}(f(\mathcal{G}), Y) + \lambda_1 \mathcal{L}_{\text{inv}} + \lambda_2 \mathcal{L}_{\text{var}},
    \label{eqn:general_joint_obj}
\end{align}
where $\lambda_1$ and $\lambda_2$ are hyperparameters that control the importance of learning $\mathbf{h}^I$ and $\mathbf{h}^V$. 
We introduce two branches of methods categorized by the variant representation learning strategy as follows. 

\noindent
\textbf{Group Invariant Learning.}
These methods extend invariant learning for GNNs by approximating the latent data environments with inferred ones. 
They do not introduce explicit objectives to regularize variant representation, \ie{$\lambda_2 = 0$}.


\emph{EERM}~\cite{EERM2022iclr} introduced a RL-based graph editor $\pi(\cdot)$ to generate discrete structure masks $\mathbf{M}^a_{k} \in \{0, 1\}^{|\mathcal{V}| \times |\mathcal{V}|}, \ k = 1,...,K$. These masks guide the addition and deletion of edges in the original graph structure via $\mathbf{A}^{k} = \mathbf{A} + \mathbf{M}^a_{k} \odot (\overline{\mathbf{A}} - \mathbf{A}).$
The modified graphs create $K$ distinct training environments, facilitating the adaptation of invariant learning to train the GNN encoder to extract $\mathbf{h}^I$. Moreover, the authors used the policy gradient method to iteratively update $\pi(\cdot)$. By maximizing the invariant regularizer, $\pi(\cdot)$ will perturb more variant components, thereby creating more diverse environments.
Overall, the invariant learning objective in EERM is
\begin{align}
    \mathcal{L}_{\text{inv}} = \Psi_{\text{V-REx}}(\{\mathcal{L}^{k}: 1\le k \le K\}) - \lambda \max_{\pi}\Psi_{\text{V-REx}}.
\end{align}

\emph{DIR}~\cite{DIR2022iclr}, instead of directly abstracting $\mathbf{h}^I$ from the input graph $\mathcal{G}$ via the GNN, explicitly disentangled invariant and variant components $\mathcal{G}^{I}$ and $\mathcal{G}^{V}$ from $\mathcal{G}$ via soft structure masks and top-$r$ truncation strategy, \ie
\begin{align}
    \begin{split}
        \mathbf{M}^a &= \sigma(\mathbf{H}^T \mathbf{H}), \ \mathbf{H} = \Phi_1(\mathbf{X}), \\ 
        \mathbf{A}^{I} &= \text{Top}_r(\mathbf{M}^a \odot \mathbf{A}), \
        \mathbf{A}^{V} = \mathbf{A} - \mathbf{A}^{I}, 
    \end{split}
    \label{eqn:dir_mask}
\end{align}
where $\Phi_1(\cdot)$ is a learnable GNN encoder that enables the mask generation in inductive settings~\cite{graphsage}, $\sigma(\cdot)$ is Sigmoid function.
Then, $\mathcal{G}^{I}$s and $\mathcal{G}^{V}$s from different graph samples were randomly paired to create augmented graphs. Those sharing the same $\mathcal{G}^{V}$ formed a data environment
\begin{align}
    e(\mathcal{G}^{V}) & := \{(\mathcal{G}^{I} \cup \mathcal{G}^{V}, Y): \mathcal{G}^{I} \subset \mathcal{G}, \forall (\mathcal{G}, Y)\in \mathcal{D}_{tr} \}.
    \label{eqn:dir_env}
\end{align}
The $\Psi_{\text{V-REx}}(\cdot)$ is calculated based on these created environments for optimizing the disentanglement and representation learning module.
Furthermore, the prediction of each graph was made not only dependent on $\mathcal{G}^{I}$ but also adjusted by the predictions based on the $\mathcal{G}^{V}$, \ie 
\begin{align}
    f(\mathcal{G}) = \left( w_I \circ \Phi_2(\mathcal{G}^{I}) \right) \odot \sigma \left( w_V \circ \Phi_2(\mathcal{G}^{V}) \right),
\end{align}
where $\Phi_2(\cdot)$ is a shared GNN encoder to produce $\mathbf{h}^I$ and $\mathbf{h}^V$, $w_I(\cdot)$ and $w_V(\cdot)$ are two downstream predictors and the gradients of $w_V(\cdot)$ are not backpropagated to $\Phi_2(\cdot)$.
\emph{InMvie}~\cite{inmvie2024fortune}
incorporated a similar idea for node-level tasks with both structure and feature masks. 

\emph{LiSA}~\cite{lisa2023cvpr} recognized the limitations of EERM and DIR that the perturbed graphs might be wrongly assigned with the original label. This phenomenon existed because those works failed to constraint that $\mathcal{G} \setminus \mathcal{G}^V$ should still be labelled as $Y.$ To this end, the invariant loss of LiSA is designed as 
\begin{align}
    \mathcal{L}_{\text{inv}} = \mathcal{L}(w_I \odot \Phi_I(\mathcal{G}^I), Y) + \Psi_{\text{V-REx}}.
\end{align}
In this way, perturbing the identified $\mathcal{G}^V$ will preserve better label invariance, facilitating high-quality environment creation for better invariant representation learning.

\emph{GIL}~\cite{GIL2022nips}
adopted a similar disentanglement strategy as in \equref{eqn:dir_mask} on each graph sample. Then they employed an invariant and a variant GNN encoders $\Phi_I(\cdot)$ and $\Phi_V(\cdot)$ to generate $\mathbf{h}^{I}$ and $\mathbf{h}^{V}$, respectively.
The specialities of these methods lie in two perspectives:
(i) they inferred the latent data environments in the training set by clustering $\mathbf{h}^{V}$ rather than creating them explicitly, \ie
\begin{align}
    \mathcal{\hat{E}}_{tr} = \text{Clustering}(\{\mathbf{h}^V_1, ..., \mathbf{h}^V_n \}), \ n = |\mathcal{D}_{tr}|;
\end{align}
(ii) they produced the prediction of a sample solely based on the invariant representation $\mathbf{h}^{I}$;
(iii) they optimized the invariant regularizer $\Psi_{\text{IGA}}(\cdot)$ to obtain the ideal invariant representation generator $\Phi(\cdot)$. Compared with $\Psi_{\text{IRM}}(\cdot)$ and $\Psi_{\text{V-REx}}(\cdot),$ the optimal $\Phi(\cdot)$ that minimizes $\Psi_{\text{IGA}}(\cdot)$ satisfies a stronger invariance than that in \assref{ass:irm_inv_ass}, \ie 
\begin{align}
    P^e(Y|\Phi(X)) = P^{e'}[Y|\Phi(X)], \ \forall e, e' \in \mathcal{E}_{tr}.
\end{align}
\emph{BA-GNN}~\cite{BAGNN2022icde} and \emph{INL}~\cite{inl2023tois} leveraged a similar idea for invariant node representation learning. 

\noindent
\textbf{Joint Invariant and Variant Learning.}
In essence, invariant and variant graph representation learning mutually enhance each other. On the one hand, exploring $\mathbf{h}^V$ facilitates a more comprehensive data environment inference, which in turn improves the learning of $\mathbf{h}^I$. Conversely, as more invariance is identified, more variant information will be captured. This reciprocal relationship has prompted recent research to simultaneously optimize both $\mathcal{\text{inv}}$ and $\mathcal{\text{var}}$, aiming to achieve more precise graph CRL for boosting GNN trustworthiness.

\emph{CAL}~\cite{CAL2022kdd} first designed the attention-based soft structure and feature masks without top-$r$ truncation for disentangling $\mathcal{G}^I$ and $\mathcal{G}^V.$
Then, the authors created environments similarly as in \equref{eqn:dir_env}.
Next, the $\mathcal{G}^I$ and $\mathcal{G}^V$ of each sample were passed separately through $\Phi_I(\cdot)$ and $\Phi_V(\cdot)$ to produce $\mathbf{h}^I$ and $\mathbf{h}^V.$
Finally, three predictors were activated to make predictions based on these representations, forming the joint learning objective in the form of \equref{eqn:general_joint_obj}:
(i) $f(\mathcal{G})$ in the primary supervision loss was produced by taking into both $\mathbf{h}^I$ and $\mathbf{h}^V$ as inputs;
(ii) $\mathcal{L}_{\text{inv}}$ was designed as the supervision losses over predictions produced only from $\mathbf{h}^I$, aiming to maintain its causal effect on labels;
(iii) $\mathcal{L}_{\text{var}}$ was defined as the KL-Divergence between the predicted distribution from $\mathbf{h}^V$ and uniform distribution, which suppresses the predictivenss of $\mathbf{h}^V.$
\emph{CMRL}~\cite{cmrl2023kdd} adapted the idea to enhance GNN generalizability in molecular relational learning tasks by identifying invariant relational substructures.
\emph{ICL}~\cite{icl2024aaai} derived similar objectives from a information-theoretic perspective.


\emph{DisC}~\cite{DisC2022nips} generated only the soft structure mask for graph-level disentanglement, and conducted a similar environment inference and encoding strategy as in CAL.
The distinctions arise from two aspects:
(i) The predictors $w_I(\cdot)$ ($w_V(\cdot)$) took the concatenated representation $\mathbf{h} := \mathbf{h}^I \oplus \mathbf{h}^V$ as input, yet its gradient was not backpropagated to $\Phi_V(\cdot)$ ($\Phi_I(\cdot)$), thereby preventing the interference in the representation learning process;
(ii) $\mathcal{L}_{\text{var}}$ adopted a tailored generalized Cross-Entropy~(CE) loss~\cite{gceloss2018nips} to expedite the learning of severe variant information,
\begin{align}
    l_{\text{var}}(w_V(\mathbf{h}), y) = \left[ 1 - \left( w_V(\mathbf{h}) \right)_y^q\right] / q,
    \label{eqn:gce_loss}
\end{align}
where $q \in (0,1]$ is a hyperparameter that controls the extent of acceleration;
(iii) The rapidly learned variant representation was then leveraged to reweight the supervision loss on invariant predictions in order to prevent the biases in the learning process of $\mathbf{h}^I$, \ie
\begin{align}
    l_{\text{inv}}(w_I(\mathbf{h}), y)) = \frac{\text{CE}(w_V(\mathbf{h}), y)\cdot \text{CE}(w_I(\mathbf{h}), y)}{\text{CE}(w_V(\mathbf{h}), y) + \text{CE}(w_I(\mathbf{h}), y)}.
\end{align}

Some works resorted to variational inference~\cite{variational2017jasa} to recover $\mathbf{h}^V,$ which are mainly motivated by two primary motivations, (i) the discovery of latent distribution of $\mathbf{h}^V$ enables more consistent data environment inference or creation, and (ii) they can generate graphs that pertain to real-world graph distribution from $\mathbf{h}^V,$ which facilitate the interpretation of GNNs.
Specifically, 
\emph{MoleOOD}~\cite{moleood2022nips} inferred the variant representation from sample $(\mathcal{G}, Y)$ via a variational distribution $q_\kappa(\mathbf{h}^V|\mathcal{G}, Y),$ which were learned by minimizing the Evidence Lower BOund (ELBO) loss. Intuitively, the ELBO loss consists of the first term of \equref{eqn:general_joint_obj} and another term that constrains the variational distribution with prior knowledge.
With the inferred $\mathbf{h}^V,$ the authors further designed objectives to ensure that the $\mathbf{h}^I$ is sufficiently informative of $Y,$ \ie{$\mathbf{h}^V \perp Y \mid \mathbf{h}^I$}, resulting in the objectives below,
\begin{align}
    \mathcal{L}_\text{inv} = \mathcal{L}(w_I(\mathbf{h}^I)), Y),
        \mathcal{L}_\text{var} = \mathbb{E}
        \left| l(w_I(\mathbf{h}^I), Y) - \mathbb{E}_{q_\kappa} l^{e(\mathbf{h}^V)} \right|.
\end{align}
\emph{OrphicX}~\cite{orphicx2022cvpr} and \emph{CANET}~\cite{canet2024www} applied a similar idea to node classification and graph explanation generation tasks, respectively. Differently, they relaxed the conditional independence restriction of $\mathbf{h}^V$ and considered it as a confounder between the $\mathbf{h}^I$ and the label $Y$ or model prediction $\hat{Y}.$ With the variational distribution of, they could stratify the confounder $\mathbf{h}^V$ for estimating the causal effect of $\mathbf{h}^I$ on the target outcome. By maximizing the causal effect, the $\mathbf{h}^I$ approximated the actual invariant factor $I$ better.

Several works assumed that the variant factor should be inherently independent of the invariant factor, \ie $I \perp V$, thereby introducing variant regularizers to strengthen the independence of $\mathbf{h}^V$ from $\mathbf{h}^I$ instead of reducing the informativeness of $\mathbf{h}^V.$ 
\emph{CIE}~\cite{cie2023cikm} promoted the independence by minimizing the normalized HSIC~score~\cite{normhsic2019icml} between the two representations.
\emph{CI-GNN}~\cite{ci-gnn2023} and \emph{RC-GNN}~\cite{rcgnn2024} achieved this goal by minimizing the mutual information between $\mathbf{h}^V$ from $\mathbf{h}^I,$ which are practically estimated via the matrix-based Renyi’s $\delta$-order entropy functional or contrastive learning, respectively.

Another way of utilizing supervision signals for invariant and variant learning is to contrast the learned representations from intra-class and inter-class views.
\emph{CIGA}~\cite{ciga2022nips} theoretically proved that under an invariant graph size assumption, this type of method is an approximation of the objective $- \text{MI}(\mathbf{h}^I, Y) + \text{MI}(\mathbf{h}^I, \mathbf{h}^V)$ when environment labels are unavailable. 
This leads to a practical contrastive learning objective $\mathcal{L}_{\text{inv}}$ where positive samples are selected as invariant representations from graphs that share the same label, and negative samples are those from graphs with different labels. 
The authors further provided a solution to relax the size assumption by maximizing the predictive power of $\mathbf{h}^V$ while limiting its predictive power to be lower than that of $\mathbf{h}^I,$ \ie
\begin{align}
    \mathcal{L}_{\text{var}} = \text{MI}(\mathbf{h}^V, Y) \cdot \mathbb{I}[\text{MI}(\mathbf{h}^V, Y) \le \text{MI}(\mathbf{h}^I, Y)].
\end{align}
This objective effectively prevented variant information from leaking into $\mathbf{h}^I.$
\emph{GALA}~\cite{gala2023nips} further relax the predictive power constraint by assuming that for any variant graph components $V,$ there exists two data environments $e_1, e_2 \in \mathcal{E}_{tr}$ such that $P^{e_1}(Y|V) \not= P^{e_2}(Y|V)$ but $P^{e_1}(Y|I) = P^{e_2}(Y|I).$ A biased model trained via empirical loss minimization is then adopted to find such data environments based on the correctness of model predictions. The contrastive learning strategy is improved accordingly by further requiring positive graphs to have different biased model predictions and negative graphs to share the same model predictions.
\emph{CAF}~\cite{caf2023cikm} incorporated a similar idea for enhancing the GCF of GNNs.
Specifically, 
two ego-graph counterfactual matching strategies are conducted for node $u$ based on the node label and sensitive attributes,
\begin{align} 
\mathcal{G}_u^I=\underset{\mathcal{G}^V \in \mathbb{G}}{\arg \min }\left\{D\left(\mathcal{G}_u, \mathcal{G}^V\right) \mid Y_u \neq Y_v, S_u=S_v\right\}, \\ 
\mathcal{G}_u^V=\underset{\mathcal{G}^V \in \mathbb{G}}{\arg \min }\left\{D\left(\mathcal{G}_u, \mathcal{G}^V\right) \mid Y_u=Y_v, S_u \neq S_v\right\},
\end{align}
where $\mathcal{G}_u$ denotes the ego-graph of node $u.$ 
Considering that counterfactual pairs with unchanged $Y$ should preserve similar $I$, and those with unchanged $S$ should have similar $V$, the following objective is proposed:
\begin{align}
    \mathcal{L}_{\text{inv}} &= \mathbb{E}_{u\in\mathcal{V}}\left[\mathbb{E}_{\mathcal{G}_u^I}\left[D(\mathbf{h}^I_u, \Phi_I(\mathcal{G}_u^I))\right] \right], \\
    \mathcal{L}_{\text{var}} &=
    \mathbb{E}_{u\in\mathcal{V}}\left[ \mathbb{E}_{\mathcal{G}_u^V}\left[D(\mathbf{h}^V_u, \Phi_V(\mathcal{G}_u^V))\right]\right].
\end{align}

Some other works pushed the generality of the identification results under assuming the availability of the environment labels.
\emph{LECI}~\cite{leci2023nips} adopted both structure and feature soft masks for disentanglement and adopted the adversarial learning objectives to enhance the precision of invariant and variant representations,
\begin{align}
    \mathcal{L}_{\text{var}} &= \max_{w_{E}} - \mathcal{L} \left( w_{E}(\mathbf{h}^I), E \right) + \max_{w_{V}} - \mathcal{L} \left( w_{V}(\mathbf{h}^V), Y \right),
\end{align}
where $w_E(\cdot)$ is another classifier that predicts the environment label. Minimizing $\mathcal{L}_{\text{var}}$ encourages both $\mathbf{h}^I \perp \mathbf{h}^V$ and $\mathbf{h}^V \perp Y.$ 
\emph{PNSIS}~\cite{pnsis2024} designed a tailored loss to promote $\mathbf{h}^I$ to be both sufficient and necessary for predicting the label, inspired by the work~\cite{pns_inv2023nips}. The authors further ensembled a variant predictor to enhance the generalization~\cite{inv_spu_ens2023nips}, assuming that $\mathbf{h}^V$ is also informative of the label.



\begin{tcolorbox}[breakable, colback=gray!5!white, colframe=gray]
\textit{\textbf{Discussion.}
Existing research has shown efficacy in enhancing the generalizability of GNNs, as invariant representations stably influence labels across environments. Approaches like graph disentanglement and generative graph reconstruction further improved interpretability by identifying critical invariant graph components. However, a trade-off exists between generalizability and interpretability, where precise GNN interpretation might require limiting the size of invariant graphs~\cite{DIR2022iclr} or extensively exploring graph distributions~\cite{ci-gnn2023}, which could detract from focusing on generalizability. Moreover, increasing attention on improving fairness and post-hoc explainability through joint invariant and variant learning~\cite{caf2023cikm,orphicx2022cvpr,ci-gnn2023} highlights these areas as promising avenues for further exploration.
Please refer to \appref{app:super_crl_discuss} for more technical comparison.}
\end{tcolorbox}

\eat{
Comparatively, group invariant learning offers theoretical causality assurances under accurate environment inference, with a straightforward optimization scheme. Yet, its effectiveness hinges on the diversity of training environments, which, if inadequate, can lead to spurious correlation leakage~\cite{towards_causal2021,zin2022nips,gala2023nips}. 
Conversely, joint invariant and variant learning approaches utilize the detailed interplay between factors $I$ and $V,$ holding the flexibility to tailor CRL to specific graph generation processes. This could circumvent the limitations seen in group invariant learning approaches. Nonetheless, the variability in assumptions about dependencies among $I$, $V$ and label $Y$ across studies can diminish these models' general applicability. Moreover, the increasing number of variant regularizers such as independence-enforcing ones complicates optimization, with a lack of comprehensive evaluations of their different practical implementations.

Overall, the prevailing focus in supervised approaches is on learning invariant and variant representations. Despite advancements, the field has yet to extensively explore more fine-grained causal representations, which could deepen understanding of the data generation process and support various downstream tasks~\cite{stcrl2023kdd}.
}


\subsubsection{Self-supervised Causal Representation Learning}
Self-supervised learning based methods learn causal representations with supervision signals from the augmented views of observed graph samples~\cite{graphssl2023tkde}.
Compared with supervised learning methods, this type of method possesses the potential to recover more fine-grained causal representations, as the augmented views can be flexibly generated by intervening different parts of the whole graph generation process~\cite{stat2causal2022ICM}.
Typically in existing works, the augmented views of one sample are crafted to share the same instance-discriminative characteristics as the original view. Based on~\assref{ass:inv_ass}, the invariant representations of graphs from related views should be highly similar, while those from unrelated views should be less similar.
This fact has inspired researchers to develop tailored self-supervised objectives to promote GNNs and generate causal representations for improving their trustworthiness.
The key challenges lie in how to generate diverse augmented views from graph data and how to choose the similarity measurement to identify causal representations. 

\eat{\TODO{
1. you may add some equations like 4.2.2.

2. it is wired that this category only has one subcategory Invariant Contrastive Learning. In such a case, you don't need a next-level subtitle.
}
\wz{revised}}


Graph Contrastive Learning (GCL) methods adopt similarity measurements built upon mutual information, such as InfoNCE~\cite{infonce2018} and SimCLR~\cite{SimCLR2020icml}, to promote the GNN to learn discriminative representations.
\emph{RGCL}~\cite{RGCL2022icml} designed an invariance-aware graph augmentation to unleash the potential of GCL in learning invariant causal representations on graphs.
\eat{
However, existing GCL methods are limited in the random or knowledge-guided manner of graph augmentation.
On the one hand, Randomly augmenting graphs might damage the discriminative semantics, \eg{the functionality of molecule graphs}. On the other hand, Knowledge-guided augmentation requires expensive human efforts and might not be good at generalizing to unseen graphs.
Incorporating the invariance assumption into GCL has shown its effectiveness in addressing the above limitations~\cite{RGCL2022icml,igcl2023tnnls}.  
Additionally, these invariance-aware GCL methods present a promising avenue to identify invariant causal representations on graphs~\cite{RGCL2022icml}.
RGCL~\cite{RGCL2022icml} organically combines the invariant and variant disentanglement similar to that of DIR~\cite{DIR2022iclr} with the graph contrastive learning, achieving noteworthy improvements on both of them. 
\hao{revise the above sentence.}
}
In detail, for each graph instance $\mathcal{G}$, RGCL first generated probability $P_I(\cdot|\mathcal{G})$ and $P_V(\cdot|\mathcal{G})$ for disentangling invariant and variant graph components, respectively,
\begin{align}
    P_I(\mathcal{G}^I|\mathcal{G}) &= \prod_{v \in \mathcal{V}^I} P(v|\mathcal{G}) \prod_{v \in \mathcal{V}^V} (1 - P(v|\mathcal{G})), \\
    P_V(\mathcal{G}^V|\mathcal{G}) &= \prod_{v \in \mathcal{V}^I} (1-P(v|\mathcal{G})) \prod_{v \in \mathcal{V}^V} P(v|\mathcal{G}),
\end{align}
where $\mathcal{V}^I$ and $\mathcal{V}^V$ denote the sets of nodes within $\mathcal{G}^I$ and $\mathcal{G}^V$, respectively, and $P(v|\mathcal{G})$ denotes the parameterized probability of $v \in \mathcal{V}^I.$
Then, graph augmentation was conducted by sampling from $P_I(\cdot|\mathcal{G})$ and $P_V(\cdot|\mathcal{G})$, with two graphs sampled from $P_I$ treated as positive pairs and those sampled from $P_V$ are treated as negative pairs.
By optimizing the GCL loss over positive and negative pairs,
RGCL optimized the disentanglement module and learned invariant graph representations to improve the OOD generalizability and inherent interpretability of GNNs.
\emph{IMoLD}~\cite{imold2023nips} counterfactually replaced the variant graph component $\mathcal{G}^V$ disentangled from the input graph with the variant factor of other graphs, creating a set of views sharing the same invariant factors $I.$ 
\emph{CGC}~\cite{cgc2023www} generated a set of graph counterfactual explanations as the hard negative samples for learning more discriminative invariant representations.
\emph{GCIL}~\cite{gcil2024aaai} further emphasized distinguishing invariant and variant information to enhance the quality of data augmentation in node-level tasks. The authors transformed graph signals into the spectral domain and treated those low (high) frequency signals as invariant (variant) components. In addition, they integrated an HSIC-based regularizer to promote the independence between $\mathbf{h}^I$ and $\mathbf{h}^V$. 

\eat{
CAF~\cite{caf2023cikm} leverages sensitive-attribute-aware counterfactual augmentation for each node, which enhances the GCF of GNNs.
Specifically, the authors assume that sensitive attributes $S$ causally affect $V$, while have no causal effect on node labels $Y$. Two ego-graph counterfactual matching strategies are conducted to for node $u$,
\begin{align} 
\mathcal{G}_u^I=\underset{\mathcal{G}^V \in \mathbb{G}}{\arg \min }\left\{D\left(\mathcal{G}_u, \mathcal{G}^V\right) \mid Y_u \neq Y_v, S_u=S_v\right\}, \\ 
\mathcal{G}_u^V=\underset{\mathcal{G}^V \in \mathbb{G}}{\arg \min }\left\{D\left(\mathcal{G}_u, \mathcal{G}^V\right) \mid Y_u=Y_v, S_u \neq S_v\right\},
\end{align}
where $\mathcal{G}_u$ denotes the ego-graph of node $u,$ $\mathbb{G}=\{\mathcal{G}_u:u\in\mathcal{V}\},$ $D(\cdot,\cdot)$ is a graph distance measurement.
Considering that counterfactual pairs with unchanged $Y$ should preserve similar $I$, and those with unchanged $S$ should have similar $V$, the following objective is proposed:
\begin{align}
    \mathcal{L}= \mathbb{E}_{u\in\mathcal{V}}\left[\mathbb{E}_{\mathcal{G}_u^I}\left[D(h^I_u, \Phi_I(\mathcal{G}_u^I))\right] + \mathbb{E}_{\mathcal{G}_u^V}\left[D(h^V_u, \Phi_V(\mathcal{G}_u^V))\right]\right].
\end{align}}


Differently, \emph{FLOOD}~\cite{flood2023kdd} designed a self-supervised task to adapt the invariant learning capability of the GNN to the new test distribution. It adopted a bootstrapped representation learning approach~\cite{floodbootstrap2020nips} to avoid the high computational and memory cost associated with negative sampling in GCL.
Specifically, a GNN-based online encoder $\Phi_\theta(\cdot)$ and a target encoder $\Phi_\xi(\cdot)$ were initiated to map the two invariant representations of two graph views into a common high-dimensional space. Then, the online representation $\Phi_\theta(\Phi(\mathcal{G}_i))$ was fed into a predictor $w_\theta(\cdot)$ to predict the target representation $\Phi_\xi(\mathcal{G}_j),$ aiming to ensure the invariance across views. During the training phase, these encoders are jointly trained under a group invariant learning framework. While in the testing phase, for a test graph, the target and online view were first generated, followed by fine-tuning the main invariant encoder $\Phi(\cdot)$ via minimizing the distance between the target and online representations.

\begin{tcolorbox}[breakable, colback=gray!5!white, colframe=gray]
\textit{
\textbf{Discussion.}
The graph augmentation in self-supervised learning aligns with the creation of data environments aimed at finding invariant graph representations, leading to great progress in GNN generalizability in existing works. These works could further enhance GNNs' interpretability by incorporating invariant and variant disentanglement~\cite{RGCL2022icml}. 
Another promising yet underexplored avenue is self-supervised CRL for enhancing the GCF of GNNs, which could be achieved by perturbing sensitive attributes to generate multiple augmented views. A challenge may arise when combining sensitive attribute-based augmentation with other augmentation types to ensure both the utility and fairness of the learned representations in downstream tasks. 
Furthermore, with no reliance on graph labels, self-supervised CRL could facilitate a more intricate identification of latent causal factors and relationships, further bolstering GNN trustworthiness.
}
\end{tcolorbox}

\section{Datasets}

In this section, we review existing graph datasets, serving as a groundwork for conducting CIGNN research.
In practice, the preference for graph datasets varies in different aspects of trustworthiness,
This section reviews existing graph datasets for conducting CIGNN research. Dataset selection differs in various trustworthiness aspects,
(i) datasets with significant training-testing distribution shifts are helpful when testing OOD generalizability;
(ii) datasets with sensitive features are preferable for evaluating fairness;
(iii) expert-curated datasets are suitable for post-hoc graph explainability studies.
Furthermore, the graph generation process within different applications can result in varied types of trustworthiness issues.
Therefore, it is crucial to first understand the dataset details and then select appropriate ones that show diversity in 
application scenarios, 
the rationale behind the trustworthiness risk, 
and the hardness of overcoming the risk.
In the following, we focus on representative data synthesis methods in the literature, and leave open-source benchmarks for evaluating the three focused trustworthiness aspects in \appref{app:benchmark}.

\eat{
\subsection{Existing Graph Datasets} \label{subsec:exist_data}
Dozens of graph benchmark datasets have been constructed over years to facilitate the advancement in the graph research community.
Real-world datasets \rev{constitute a significant portion of these benchmarks} due to the ubiquitous nature of graph structures in multiple application fields such as social networks \TODO{addref} and drug development \TODO{addref}.
However, real-world datasets often contain noise due to varied collection biases, and they lack the flexibility needed to systematically explore the properties and limitations of graph algorithms. To overcome these limitations, synthetic datasets are generated to mimic the key properties of real-world datasets while providing greater control and flexibility.


}

\subsection{Data Synthesis for Evaluating OOD Generalizability}

\label{subsubsec:ood_data_syn}
Despite the diversity in existing benchmark graph OOD datasets,
\eat{
they are limited in the inability to adjust the degrees of distribution shifts and the ambiguity of underlying causal mechanisms. It is crucial, yet nontrivial, to controllably synthesize diverse distribution shifts spanning various severities and causal mechanisms to more comprehensively assess the generalizability of a GNN}
it is crucial, yet nontrivial, to controllably synthesize distribution shifts spanning various degrees of severity under different data causal mechanisms to more comprehensively assess the generalizability of a GNN.
As illustrated in \figref{fig:graph_DGP_ood+exp}, the spurious correlation $P(Y|V)$ is the key source of the poor OOD performance of GNNs. 
Therefore, it is reasonable to create multiple levels of distribution shifts by manually altering $P(Y|V).$
Inspired by~\cite{oodsurvey2021}, we classify existing graph distribution shift synthesis strategies into two categories based on the way of varying $P(Y|V)$: introducing data selection bias or anti-causal effect into the data generation process. 

\noindent
\textbf{Varying data selection bias.}
Prior works~\cite{DIR2022iclr,CAL2022kdd,GIL2022nips} generate training sets by repeatedly sampling an invariant subgraph $\mathcal{G}^I$ from a uniform distribution and combining it with different variant subgraphs $\mathcal{G}^V$ based on distribution $P(\mathcal{G}^V|\mathcal{G}^I) = b \cdot \mathbb{I}(\mathcal{G}^V = \mathcal{G}^I) + (1-b)\cdot \mathbb{I}(\mathcal{G}^V \neq \mathcal{G}^I).$ Larger hyperparameter $b$ intensify the correlation between $\mathcal{G}^I$ and $\mathcal{G}^V$, thereby exacerbating spurious correlation $P(Y|\mathcal{G}^I)$. 

Other approaches implicitly varying $P(V|I)$ by generating data selection biases \wrt{different types of graph properties}.
Li~\etal~\cite{oodgnn2022tkde} introduced feature selection bias by adding independent Gaussian noise with controllable distribution.
To introduce topology-level bias, 
Li~\etal~\cite{oodgnn2022tkde} and Li~\etal~\cite{GIL2022nips} created testing sets that consist of graphs with unseen sizes or unseen structures, respectively.
Fan~\etal~\cite{debiasedgnn2022tnnls} selected nodes with neighbor distribution ratio larger than a controllable threshold for GNN training.
Chen~\etal~\cite{BAGNN2022icde} grouped graph nodes according to their labels or degrees to create testing environments with varied levels of distribution shifts.

\noindent
\textbf{Generating spurious features that are anti-causally affected by the label.}
Take the strategy adopted in work~\cite{EERM2022iclr} as an example. 
A randomly initialized GNN was first adopted to generate node labels $\mathbf{Y}$ for a given graph with node features $\mathbf{X}_1$ and adjacency matrix as input.
Then to generate $m$ data environments, the authors utilized another randomly initialized GNN to generate spurious node features $\mathbf{X}_2^i$ for environment $i$ with input of label $\mathbf{Y}$ and environment index $i, 1 \le i \le m$. After that, they concatenated $\mathbf{X}_1$ and $\mathbf{X}_2^i$ as the input node features in environment $i$. In this way, multiple data environments were constructed with varied spurious correlations between $\mathbf{X}_2$ and $\mathbf{Y}$. 
\subsection{Data Synthesis for Evaluating Graph Fairness}

\eat{
\subsubsection{Benchmark Datasets}
The datasets used for graph fairness research are generated to include examples of potential bias and unfairness, such as under-represented groups or imbalanced classes, and require additional considerations beyond traditional graph learning benchmarks. 
Dong~\etal~\cite{fairness4gnn2022survey} summarized the benchmark graph fairness datasets and categorized them into social networks, recommendation-based networks, academic networks, and other types of networks. 
Qian and Guo~\etal~\cite{fairgraphbench2024} curated a collection of synthetic, semi-synthetic, and real-world datasets. They are tailored to consider graph structure utility and bias information, offering flexibility to create data with controllable biases for comprehensive evaluation.

\eat{
Different datasets might support studies on varied fairness notions.
Here we exclusively present datasets related to the evaluation of graph counterfactual fairness in \tabref{tab:fairness_data}.
\begin{table}[h]
    \centering
    \begin{tabular}{c|c}
         &  \\
         & 
    \end{tabular}
    \caption{Widely-used benchmark graph datasets for graph counterfactual fairness research.}
    \label{tab:fairness_data}
\end{table}
}

\subsubsection{Specialized Data Synthesis}
}
Evaluating the fairness of GNNs on real-world datasets possesses two limitations: (1) defining sensitive attributes in real-world graph datasets requires strong domain knowledge, which is not always available; (2) intransparency of the graph generation mechanism poses additional difficulties in obtaining the ground-truth counterfactual graphs required to evaluate GCF. 
To this end, controllable synthetic graph datasets are demanded.

Ma \etal~\cite{gear2022wsdm} synthesized data based on a predefined causal model for evaluating the GCF of GNNs, where the influence of sensitive attributes can be varied manually,
\begin{align}
    & S_i \sim Bern(p), \mathbf{h}_i, \mathbf{v} \sim \mathcal{N}(0,\mathbf{I}), \mathbf{X}_i = \mathcal{S}(\mathbf{h}_i) + S_i \mathbf{v}, \\
    & P(a_{i,j}=1) = \sigma(\cos(\mathbf{h}_i, \mathbf{h}_j) + \alpha \mathbb{I}(S_i=S_j)), \label{eqn:edge_gen_prob}\\
    & \mathbf{w} \sim \mathcal{N}(0,\mathbf{I}), Y_i = \mathcal{B}(\mathbf{w}^T\mathbf{h}_i + w_s \frac{\sum_{j\in\mathcal{N}_i} S_j}{|\mathcal{N}_i|}),
\end{align}
Here sensitive attribute is sampled from a Bernoulli distribution of probability $p$, node features and graph structures are both generated from latent factors $\{\mathbf{h}_i\}$ and sensitive attributes $\{S_i\},$ $\mathbf{v} \in \mathbb{R}^d$ controls the influence of the sensitive attribute on other features. $\alpha$ is a hyperparameter and $\sigma(\cdot)$ is Sigmoid function. Each node’s and their 1-hop neighbors’ sensitive attributes are aggregated to generate a binary label $Y_i$.
Under this definition, one can compute the ground-truth counterfactual graph after perturbing certain sensitive attributes via rerunning this causal model.
Guo~\etal~\cite{caf2023cikm} also leveraged this data synthesis idea to generate node features consisting of an invariant part only affected by $Y_i$ and a variant part only affected by $S_i.$

\section{Evaluation Metrics}
Evaluation metric selection is a crucial step to ensure a comprehensive and accurate evaluation of the proposed models. 
Using a single metric may not be sufficient, and can lead to potential biases or errors. 
Therefore, appropriate and diverse metrics should be carefully adopted to avoid incorrect conclusions and provide a more comprehensive evaluation.
Researchers often prioritize accuracy-related metrics to quantify the model's utility in graph-related applications, including Accuracy~\cite{GCN}, ROC-AUC~\cite{ogb2020nips}, F1-score~\cite{graphsmote2021wsdm} and Precision~\cite{DIR2022iclr}.
However, they might fail to reveal the trustworthiness of the model. To this end, several TGNN metrics have been proposed,
including average accuracy, standard deviation accuracy and worst-case accuracy for evaluating OOD generalizability, 
correlation-based and GCF-induced metrics for evaluating graph fairness,
and factual and GCE-induced metrics for evaluating graph explainability.
Please refer to \appref{app:metric} for more details.

\section{Codes and Packages} 
\label{sec:causal_package}
We summarize available codes for the reviewed literature in \tabref{tab:work_summary} to facilitate comparative study with existing CIGNNs.
\eat{
}
In addition, there are some high-quality Python toolboxes and libraries that facilitate researchers and practitioners in implementing, developing, and systematically evaluating TGNN methods. 
\eat{\hao{What is a benchmark? I feel you misunderstood it.} 
\wz{
1. A benchmark is a standard or point of reference people can use to measure something else.
2. Changed 'benchmark' to 'toolbox'.
}}
GOOD\footnote{https://github.com/divelab/GOOD/} provides convenient APIs for reproducing state-of-the-art graph OOD methods. Moreover, the extensible pipeline of GOOD assists in producing new methods and datasets as well as conducting comprehensive evaluations.
\eat{
GOOD, a graph OOD benchmark:
Easy-to-use APIs: GOOD provides simple APIs for loading OOD algorithms, graph neural networks, and datasets so that you can take only several lines of code to start.
Flexibility: Full OOD split generalization code is provided for extensions and any new graph OOD dataset contributions. OOD algorithm base class can be easily overwritten to create new OOD methods.
Easy-to-extend architecture: In addition to playing as a package, GOOD is also an integrated and well-organized project ready to be further developed. All algorithms, models, and datasets can be easily registered by register and automatically embedded into the designed pipeline like a breeze! The only thing the user needs to do is write your own OOD algorithm class, your own model class, or your new dataset class. Then you can compare your results with the leaderboard.
}
Prado-Romero \etal~\cite{gretel2022cikm} proposed GRETEL\footnote{https://github.com/MarioTheOne/GRETEL}, a unified toolbox that 
contains both real and synthetic datasets, GNN models, state-of-the-art GCE methods, and evaluation metrics. Moreover, it provides a systematic evaluation pipeline along with a user-friendly interface for developing GCE methods and testing them across various application domains and evaluation metrics.
\eat{
GRETEL, a unified framework to develop and test GCE methods in several settings. 
providing a set of well-defined mechanisms to integrate and manage easily: both real and synthetic datasets, ML models, state-of-the-art explanation techniques, and evaluation measures.
It can be adopted effortlessly by future researchers who want to create and test their new explanation methods by comparing them to existing techniques across several application domains, data and evaluation measures.
}

We also present some causal learning related Python toolboxes that might facilitate researchers conducting causal learning on graph datasets and discovering essential knowledge for developing more advanced CIGNNs.
Causebox\footnote{https://github.com/paras2612/CauseBox}~\cite{causebox2021cikm} reproduces seven state-of-the-art causal reasoning methods and conducts a comparative analysis using two benchmark datasets. It also provides a complete interface for executing its evaluation pipeline on specified methods.
Causal-learn\footnote{https://github.com/py-why/causal-learn}~\cite{causallearn2023} is an open-source python package for causal discovery, 
which includes classic causal discovery algorithms and APIs, and provides modularized code to facilitate researchers in implementing their own algorithms.
\section{Conclusion and Future Directions}\label{sec:future}
In this article, we presented a comprehensive survey of existing CIGNNs works, focusing on how they empower GNNs with different causal learning abilities to improve the trustworthiness of GNNs.
We first analyzed the trustworthiness risks of GNNs in the lens of causality. Then, we categorized existing CIGNNs based on the causal learning capability they are equipped with, and introduced representative causal techniques in each category. 
Finally, we listed useful resources to facilitate further exploration on CIGNNs.

In the following, we discuss several future directions to illuminate further research on incorporating causal learning to enhance the trustworthiness of GNNs.


\noindent
\textbf{(1) Scale CIGNNs to large graphs.}
Large-scale graphs are prevalent in real-world applications, including biochemistry~\cite{alphago2021nature} 
and recommendation systems~\cite{pinsage2018kdd}. Mainstream GNNs struggle to scale up to large graphs due to the costly neighborhood expansion within GNNs' message passing scheme~\cite{pasca2022www}.
While numerous scalable GNNs have been proposed~\cite{scalablegnn_benchmark2022nips}, 
there exists a notable research gap concerning the scalability of CIGNNs.
On the one hand, techniques adopted in CIGNNs might 
not be scalable to larger graphs. For instance, graph perturbation adopted to create multiple counterfactual graphs~\cite{EERM2022iclr, gear2022wsdm} will lead to higher computation cost as the size of the graph grows.
On the other hand, techniques devised for scalable GNNs may not be seamlessly integrated into CIGNNs. For example, the sampling strategy employed in the message passing of GNNs~\cite{graphsage} inevitably perturbs both invariant and variant components within a node's neighborhood, raising concerns regarding its compatibility with node-level causal representation learning methods.
Further exploration is needed to improve the scalability of CIGNNs.


\noindent
\textbf{(2) Causality-inspired graph foundation models.}
The success of Large Language Models (LLMs) has sparked extensive exploration into the development of graph foundation models that are pre-trained on diverse graph data and can subsequently be adapted for a wide array of downstream graph tasks~\cite{gfmsurvey2023}.
\eat{
A graph foundation model is a large graph model that can handle different graph tasks across various domains. It is equipped with “commonsense knowledge” of graphs, which requires the model to gain understanding of the inherent structural information and properties of graphs. The emergence and homogenization capabilities of foundation models have piqued the interest of graph machine learning researchers, sparking discussions about developing the next graph learning paradigm that is pre-trained on broad graph data and can be adapted to a wide range of downstream graph tasks.
}
Integrating causality into the development of trustworthy large graph models is a promising direction~\cite{causal&llm2023,trustllm2023}.
Nevertheless, the increase in model size raises scepticism regarding the efficacy of existing causality-inspired approaches that are mainly evaluated on GNNs with smaller sizes.
For instance, the graph invariant learning method proposed to enhance the generalizability of GNNs might fail to reduce spurious correlations due to the overfitting problem inherent in overparameterized large models~\cite{birm2022cvpr,sparseirm2022icml}.
Given the substantial potential of large graph models in revolutionizing the graph learning paradigm, it is imperative to critically assess existing works and explore novel causality-based approaches to enhance the trustworthiness of large graph models.

\noindent
\textbf{(3) Causal discovery on graphs.}
Empowering GNNs with causal reasoning or CRL abilities comes with inherent limitations.
Domain knowledge regarding causal relations among graph components is often required to abstract meaningful causal reasoning tasks for enhancing trustworthiness~\cite{cflp2022icml}, which might be lacking in certain applications.
Although graph CRL methods can recover latent causal structures from raw input space, their identifiability heavily relies on assumptions on the underlying graph generation process~\cite{towards_causal2021,inv_identifiability2022iclr}.
Causal discovery aims to identify causal relations among variables in a data-driven manner~\cite{causal_data_survey2020,cdsurvey2023acm}.
This not only complements the lack of domain knowledge in causal reasoning but also enables examining the satisfaction of data causality assumptions.
Moreover, the discovered causal knowledge can be further instilled into GNN models to facilitate learning semantically meaningful and identifiable graph representations~\cite{castle2020nips}.
Hence, equipping GNNs with causal discovery ability is promising for developing TGNNs that excel in diverse application contexts, despite limited studies in this direction~\cite{crcg2024aaai,dcsgl2024kbs}.

\noindent

\eat{
\noindent
\textbf{Reducing unmeasured confounding.}
It is a great step to attributing prominent subgraph \wrt GNN predictions on the basis of estimated causal effects, however, most works neglect the importance of controlling the confounders between subgraph and predictions.
Although DSE~\cite{oodexplanation2022} has discussed the unmeasured confounding induced by the distribution shift before and after feature removal, it still lacks theoretical support for the proposed generative front-door adjustment.
Other unknown confounding effects might also need to be discussed to further improve the consistency and faithfulness of causal explanations.
Methods like instrumental variables~\cite{IVsurvey2022} or generative unknown confounders approximation~\cite{robustevent2022kdd} might be helpful for future research.
}

\noindent
\textbf{(4) Beyond graph counterfactual fairness.}
GCF notion measures the fairness of GNNs based on the total causal effect~\cite{causality2009pearl} of sensitive attributes on the output node representations. 
However, GCF is not suitable in scenarios where unfairness exists because sensitive attributes causally affect the outcome along certain causal paths~\cite{pathcf2019aaai}.
Intervention-based fairness notion~\cite{interventionfair2012tkde,interventionfair2018aaai} has been proposed to distinctly capture the most prominent causal mechanisms that result in discrimination in real-world applications, \eg{direct or indirect causal paths}.
Path-specific counterfactual fairness notion~\cite{pathcf2019aaai} accounts for the counterfactual fairness of model decisions along the unfair paths.
Nevertheless, neither of them has been adapted to improve graph fairness, which necessitates the development of causality-based graph fairness notions beyond GCF.


\noindent
\textbf{(5) Causality-inspired privacy preservation.}
Privacy preservation constitutes another critical facet of trustworthiness, imposing additional constraints on GNNs~\cite{fedgnn2022nc, fedgnn4recsys2021arxiv}. 
Vo \etal~\cite{ce_privacy2023kdd} recently explored the intersection of privacy and causality-inspired AI by studying the generation of privacy-preserving counterfactual explanations.
However, there is a gap in the literature, with no works examining CIGNN systems from the privacy-preserving perspective. It is worthwhile to investigate the compatibility of existing CIGNNs with privacy-preserving techniques to establish trustworthy GNN systems applicable in privacy-critical scenarios.

\ifCLASSOPTIONcaptionsoff
  \newpage
\fi


\bibliographystyle{IEEEtran}
\bibliography{ref}

\begin{thebibliography}{100}
\providecommand{\url}[1]{#1}
\csname url@samestyle\endcsname
\providecommand{\newblock}{\relax}
\providecommand{\bibinfo}[2]{#2}
\providecommand{\BIBentrySTDinterwordspacing}{\spaceskip=0pt\relax}
\providecommand{\BIBentryALTinterwordstretchfactor}{4}
\providecommand{\BIBentryALTinterwordspacing}{\spaceskip=\fontdimen2\font plus
\BIBentryALTinterwordstretchfactor\fontdimen3\font minus
  \fontdimen4\font\relax}
\providecommand{\BIBforeignlanguage}[2]{{%
\expandafter\ifx\csname l@#1\endcsname\relax
\typeout{** WARNING: IEEEtran.bst: No hyphenation pattern has been}%
\typeout{** loaded for the language `#1'. Using the pattern for}%
\typeout{** the default language instead.}%
\else
\language=\csname l@#1\endcsname
\fi
#2}}
\providecommand{\BIBdecl}{\relax}
\BIBdecl

\bibitem{gnnsurvey2021tnnls}
Z.~Wu, S.~Pan, F.~Chen, G.~Long, C.~Zhang, and P.~S. Yu, ``A comprehensive
  survey on graph neural networks,'' \emph{IEEE Trans. Neural Netw. Learn.
  Syst.}, vol.~32, no.~1, pp. 4--24, 2021.

\bibitem{gnn4bio2022}
H.~Yi, Z.~You, D.~Huang, and C.~K. Kwoh, ``Graph representation learning in
  bioinformatics: trends, methods and applications,'' \emph{Brief. Bioinform.},
  vol.~23, no.~1, p. bbab340, 2022.

\bibitem{gnn4rs2023acm}
S.~Wu, F.~Sun, W.~Zhang, X.~Xie, and B.~Cui, ``Graph neural networks in
  recommender systems: {A} survey,'' \emph{ACM Comput. Surv.}, vol.~55, no.~5,
  pp. 97:1--97:37, 2023.

\bibitem{gnn4kg2017tkde}
Q.~Wang, Z.~Mao, B.~Wang, and L.~Guo, ``Knowledge graph embedding: {A} survey
  of approaches and applications,'' \emph{IEEE Trans. Knowl. Data Eng.},
  vol.~29, no.~12, pp. 2724--2743, 2017.

\bibitem{uukg2023nips}
Y.~Ning, H.~Liu, H.~Wang, Z.~Zeng, and H.~Xiong, ``{UUKG:} unified urban
  knowledge graph dataset for urban spatiotemporal prediction,'' \emph{CoRR},
  vol. abs/2306.11443, 2023.

\bibitem{talent2022kdd}
Z.~Guo, H.~Liu, L.~Zhang, Q.~Zhang, H.~Zhu, and H.~Xiong, ``Talent
  demand-supply joint prediction with dynamic heterogeneous graph enhanced
  meta-learning,'' in \emph{Proc. 28th ACM SIGKDD Conf. Knowl. Discov. Data
  Mining}, 2022, pp. 2957--2967.

\bibitem{mugrep2021kdd}
W.~Zhang, H.~Liu, L.~Zha, H.~Zhu, J.~Liu, D.~Dou, and H.~Xiong, ``Mugrep: {A}
  multi-task hierarchical graph representation learning framework for real
  estate appraisal,'' in \emph{Proc. 27th ACM SIGKDD Conf. Knowl. Discov. Data
  Mining}, 2021, pp. 3937--3947.

\bibitem{semiparking2022tkde}
W.~Zhang, H.~Liu, Y.~Liu, J.~Zhou, T.~Xu, and H.~Xiong, ``Semi-supervised
  city-wide parking availability prediction via hierarchical recurrent graph
  neural network,'' \emph{IEEE Trans. Knowl. Data Eng.}, vol.~34, no.~8, pp.
  3984--3996, 2022.

\bibitem{airpred2023tkde}
J.~Han, H.~Liu, H.~Zhu, and H.~Xiong, ``Kill two birds with one stone: {A}
  multi-view multi-adversarial learning approach for joint air quality and
  weather prediction,'' \emph{{IEEE} Trans. Knowl. Data Eng.}, vol.~35, no.~11,
  pp. 11\,515--11\,528, 2023.

\bibitem{TrustGNNsurvey2022}
E.~Dai, T.~Zhao, H.~Zhu, J.~Xu, Z.~Guo, H.~Liu, J.~Tang, and S.~Wang, ``A
  comprehensive survey on trustworthy graph neural networks: Privacy,
  robustness, fairness, and explainability,'' \emph{CoRR}, vol. abs/2204.08570,
  2022.

\bibitem{TwGL2022}
B.~Wu, J.~Li, J.~Yu, Y.~Bian, H.~Zhang, C.~Chen, C.~Hou, G.~Fu, L.~Chen, T.~Xu,
  Y.~Rong, X.~Zheng, J.~Huang, R.~He, B.~Wu, G.~Sun, P.~Cui, Z.~Zheng, Z.~Liu,
  and P.~Zhao, ``A survey of trustworthy graph learning: Reliability,
  explainability, and privacy protection,'' \emph{CoRR}, vol. abs/2205.10014,
  2022.

\bibitem{tgnn2022peijian}
H.~Zhang, B.~Wu, X.~Yuan, S.~Pan, H.~Tong, and J.~Pei, ``Trustworthy graph
  neural networks: Aspects, methods and trends,'' \emph{CoRR}, vol.
  abs/2205.07424, 2022.

\bibitem{oodgraphsurvey2022}
H.~Li, X.~Wang, Z.~Zhang, and W.~Zhu, ``Out-of-distribution generalization on
  graphs: {A} survey,'' \emph{CoRR}, vol. abs/2202.07987, 2022.

\bibitem{ASTFA2022nips}
F.~Liu, H.~Liu, and W.~Jiang, ``Practical adversarial attacks on spatiotemporal
  traffic forecasting models,'' in \emph{Adv. Neural Inf. Process. Syst. 35},
  2022, pp. 19\,035--19\,047.

\bibitem{fairness4gnn2022survey}
Y.~Dong, J.~Ma, C.~Chen, and J.~Li, ``Fairness in graph mining: {A} survey,''
  \emph{CoRR}, vol. abs/2204.09888, 2022.

\bibitem{gear2022wsdm}
J.~Ma, R.~Guo, M.~Wan, L.~Yang, A.~Zhang, and J.~Li, ``Learning fair node
  representations with graph counterfactual fairness,'' in \emph{Proc. 15th
  {ACM} Int. Conf. Web Search and Data Mining}, 2022, pp. 695--703.

\bibitem{xgnnsurvey2022tpami}
H.~Yuan, H.~Yu, S.~Gui, and S.~Ji, ``Explainability in graph neural networks: A
  taxonomic survey,'' \emph{IEEE Trans. Pattern Anal. Mach. Intell.}, vol.~45,
  no.~05, pp. 5782--5799, 2023.

\bibitem{gnnexplainer2019nips}
Z.~Ying, D.~Bourgeois, J.~You, M.~Zitnik, and J.~Leskovec, ``{GNNExplainer}:
  Generating explanations for graph neural networks,'' in \emph{Adv. Neural
  Inf. Process. Syst. 32}, 2019, pp. 9240--9251.

\bibitem{gnn4anomaly2021}
X.~Ma, J.~Wu, S.~Xue, J.~Yang, Q.~Z. Sheng, and H.~Xiong, ``A comprehensive
  survey on graph anomaly detection with deep learning,'' \emph{CoRR}, vol.
  abs/2106.07178, 2021.

\bibitem{nifty2021uai}
C.~Agarwal, H.~Lakkaraju, and M.~Zitnik, ``Towards a unified framework for fair
  and stable graph representation learning,'' in \emph{Proc. 37th Conf.
  Uncertainty in Artif. Intell.}, vol. 161, 2021, pp. 2114--2124.

\bibitem{causality2009pearl}
J.~Pearl, \emph{Causality}.\hskip 1em plus 0.5em minus 0.4em\relax Cambridge
  university press, 2009.

\bibitem{causal_data_survey2020}
R.~Guo, L.~Cheng, J.~Li, P.~R. Hahn, and H.~Liu, ``A survey of learning
  causality with data: Problems and methods,'' \emph{ACM Comput. Surv.},
  vol.~53, no.~4, pp. 75:1--75:37, 2020.

\bibitem{stat2causal2022ICM}
B.~Sch{\"{o}}lkopf and J.~von K{\"{u}}gelgen, ``From statistical to causal
  learning,'' \emph{CoRR}, vol. abs/2204.00607, 2022.

\bibitem{SRAIsurvey2021ijcai}
L.~Cheng, A.~Mosallanezhad, P.~Sheth, and H.~Liu, ``Causal learning for
  socially responsible {AI},'' in \emph{Proc. 30th Int. Joint Conf. Artif.
  Intell.}, 2021, pp. 4374--4381.

\bibitem{causal_tai_survey2023}
H.~Liu, M.~Chaudhary, and H.~Wang, ``Towards trustworthy and aligned machine
  learning: {A} data-centric survey with causality perspectives,'' \emph{CoRR},
  vol. abs/2307.16851, 2023.

\bibitem{seventools2019pearl}
J.~Pearl, ``The seven tools of causal inference, with reflections on machine
  learning,'' \emph{Commun. {ACM}}, vol.~62, no.~3, pp. 54--60, 2019.

\bibitem{EERM2022iclr}
Q.~Wu, H.~Zhang, J.~Yan, and D.~Wipf, ``Handling distribution shifts on graphs:
  An invariance perspective,'' in \emph{10th Int. Conf. Learn.
  Representations}, 2022.

\bibitem{causalrecsurvey2022}
C.~Gao, Y.~Zheng, W.~Wang, F.~Feng, X.~He, and Y.~Li, ``Causal inference in
  recommender systems: {A} survey and future directions,'' \emph{CoRR}, vol.
  abs/2208.12397, 2022.

\bibitem{inl2023tois}
H.~Li, Z.~Zhang, X.~Wang, and W.~Zhu, ``Invariant node representation learning
  under distribution shifts with multiple latent environments,'' \emph{{ACM}
  Trans. Information Systems}, vol.~42, no.~1, pp. 26:1--26:30, 2024.

\bibitem{debiasedgnn2022tnnls}
S.~Fan, X.~Wang, C.~Shi, K.~Kuang, N.~Liu, and B.~Wang, ``Debiased graph neural
  networks with agnostic label selection bias,'' \emph{CoRR}, vol.
  abs/2201.07708, 2022.

\bibitem{gem2021icml}
W.~Lin, H.~Lan, and B.~Li, ``Generative causal explanations for graph neural
  networks,'' in \emph{Proc. 38th Int. Conf. Mach. Learn.}, vol. 139, 2021, pp.
  6666--6679.

\bibitem{DIR2022iclr}
Y.~Wu, X.~Wang, A.~Zhang, X.~He, and T.~Chua, ``Discovering invariant
  rationales for graph neural networks,'' in \emph{10th Int. Conf. Learn.
  Representations}, 2022.

\bibitem{orphicx2022cvpr}
W.~Lin, H.~Lan, H.~Wang, and B.~Li, ``Orphicx: {A} causality-inspired latent
  variable model for interpreting graph neural networks,'' in \emph{2022
  IEEE/CVF Conf. Comput. Vision Pattern Recognit.}, 2022, pp. 13\,719--13\,728.

\bibitem{good2022nips}
S.~Gui, X.~Li, L.~Wang, and S.~Ji, ``{GOOD:} {A} graph out-of-distribution
  benchmark,'' in \emph{Adv. Neural Inf. Process. Syst. 35}, 2022, pp.
  2059--2073.

\bibitem{gce_survey2022}
M.~A. Prado-Romero, B.~Prenkaj, G.~Stilo, and F.~Giannotti, ``A survey on graph
  counterfactual explanations: Definitions, methods, evaluation, and research
  challenges,'' \emph{ACM Comput. Surv.}, 2023, to be published.

\bibitem{graph4causal2022aimagazine}
J.~Ma and J.~Li, ``Learning causality with graphs,'' \emph{{AI} Mag.}, vol.~43,
  no.~4, pp. 365--375, 2022.

\bibitem{explorecausalgnn2023}
S.~Job, X.~Tao, T.~Cai, H.~Xie, L.~Li, J.~Yong, and Q.~Li, ``Exploring causal
  learning through graph neural networks: An in-depth review,'' \emph{CoRR},
  vol. abs/2311.14994, 2023.

\bibitem{cf4graph2023survey}
Z.~Guo, T.~Xiao, C.~Aggarwal, H.~Liu, and S.~Wang, ``Counterfactual learning on
  graphs: {A} survey,'' \emph{CoRR}, vol. abs/2304.01391, 2023.

\bibitem{mpnn2017icml}
J.~Gilmer, S.~S. Schoenholz, P.~F. Riley, O.~Vinyals, and G.~E. Dahl, ``Neural
  message passing for quantum chemistry,'' in \emph{Proc. 34th Int. Conf. Mach.
  Learn.}, vol.~70, 2017, pp. 1263--1272.

\bibitem{spectral_before_1}
D.~I. Shuman, S.~K. Narang, P.~Frossard, A.~Ortega, and P.~Vandergheynst, ``The
  emerging field of signal processing on graphs: Extending high-dimensional
  data analysis to networks and other irregular domains,'' \emph{IEEE Signal
  Process. Mag.}, vol.~30, no.~3, pp. 83--98, 2013.

\bibitem{spectral_before_2}
J.~Bruna, W.~Zaremba, A.~Szlam, and Y.~LeCun, ``Spectral networks and locally
  connected networks on graphs,'' in \emph{2nd Int. Conf. Learn.
  Representations}, 2014.

\bibitem{powerfulspectral2022icml}
X.~Wang and M.~Zhang, ``How powerful are spectral graph neural networks,'' in
  \emph{Proc. 39th Int. Conf. Mach. Learn.}, vol. 162, 2022, pp.
  23\,341--23\,362.

\bibitem{spectral}
M.~Defferrard, X.~Bresson, and P.~Vandergheynst, ``Convolutional neural
  networks on graphs with fast localized spectral filtering,'' in \emph{Adv.
  Neural Inf. Process. Syst. 29}, 2016.

\bibitem{diffpool2018nips}
Z.~Ying, J.~You, C.~Morris, X.~Ren, W.~L. Hamilton, and J.~Leskovec,
  ``Hierarchical graph representation learning with differentiable pooling,''
  in \emph{Adv. Neural Inf. Process. Syst. 31}, 2018, pp. 4805--4815.

\bibitem{sapooling2019icml}
J.~Lee, I.~Lee, and J.~Kang, ``Self-attention graph pooling,'' in \emph{Proc.
  36th Int. Conf. Mach. Learn.}, vol.~97, 2019, pp. 3734--3743.

\bibitem{pretraingnn2022}
J.~Xia, Y.~Zhu, Y.~Du, and S.~Z. Li, ``A survey of pretraining on graphs:
  Taxonomy, methods, and applications,'' \emph{CoRR}, vol. abs/2202.07893,
  2022.

\bibitem{netembed2019tkde}
P.~Cui, X.~Wang, J.~Pei, and W.~Zhu, ``A survey on network embedding,''
  \emph{IEEE Trans. Knowl. Data Eng.}, vol.~31, no.~5, pp. 833--852, 2019.

\bibitem{potential_outcome}
D.~B. Rubin, ``Estimating causal effects of treatments in randomized and
  nonrandomized studies.'' \emph{J. Educational Psychol.}, vol.~66, no.~5, p.
  688, 1974.

\bibitem{CGI2021sigir}
F.~Feng, W.~Huang, X.~He, X.~Xin, Q.~Wang, and T.~Chua, ``Should graph
  convolution trust neighbors? {A} simple causal inference method,'' in
  \emph{Proc. 44th Int. {ACM} {SIGIR} Conf. Res. Develops. Inf. Retrieval},
  2021, p. 1208–1218.

\bibitem{icmvae2023nips}
A.~Komanduri, Y.~Wu, F.~Chen, and X.~Wu, ``Learning causally disentangled
  representations via the principle of independent causal mechanisms,''
  \emph{CoRR}, vol. abs/2306.01213, 2023.

\bibitem{reviewcd2019}
C.~Glymour, K.~Zhang, and P.~Spirtes, ``Review of causal discovery methods
  based on graphical models,'' \emph{Frontiers in Genetics}, vol.~10, p. 524,
  2019.

\bibitem{cdsurvey2023acm}
M.~J. Vowels, N.~C. Camg{\"{o}}z, and R.~Bowden, ``D'ya like dags? {A} survey
  on structure learning and causal discovery,'' \emph{ACM Comput. Surv.},
  vol.~55, no.~4, pp. 82:1--82:36, 2023.

\bibitem{identify_dgm2022nips}
B.~Kivva, G.~Rajendran, P.~Ravikumar, and B.~Aragam, ``Identifiability of deep
  generative models without auxiliary information,'' in \emph{Adv. Neural Inf.
  Process. Syst. 35}, 2022, pp. 5687--15\,701.

\bibitem{oodgnn2022tkde}
H.~Li, X.~Wang, Z.~Zhang, and W.~Zhu, ``{OOD-GNN:} out-of-distribution
  generalized graph neural network,'' \emph{{IEEE} Trans. Knowl. Data Eng.},
  vol.~35, no.~7, pp. 7328--7340, 2023.

\bibitem{oodsurvey2021}
Z.~Shen, J.~Liu, Y.~He, X.~Zhang, R.~Xu, H.~Yu, and P.~Cui, ``Towards
  out-of-distribution generalization: A survey,'' \emph{CoRR}, vol.
  abs/2108.13624, 2021.

\bibitem{stable2022nature}
P.~Cui and S.~Athey, ``Stable learning establishes some common ground between
  causal inference and machine learning,'' \emph{Nature Machine Intelligence},
  vol.~4, no.~2, pp. 110--115, 2022.

\bibitem{DisC2022nips}
S.~Fan, X.~Wang, Y.~Mo, C.~Shi, and J.~Tang, ``Debiasing graph neural networks
  via learning disentangled causal substructure,'' \emph{CoRR}, vol.
  abs/2209.14107, 2022.

\bibitem{ciga2022nips}
Y.~Chen, Y.~Zhang, Y.~Bian, H.~Yang, M.~KAILI, B.~Xie, T.~Liu, B.~Han, and
  J.~Cheng, ``Learning causally invariant representations for
  out-of-distribution generalization on graphs,'' in \emph{Adv. Neural Inf.
  Process. Syst. 35}, 2022, pp. 22\,131--22\,148.

\bibitem{cf2017nips}
M.~J. Kusner, J.~R. Loftus, C.~Russell, and R.~Silva, ``Counterfactual
  fairness,'' in \emph{Adv. Neural Inf. Process. Syst. 30}, 2017, pp.
  4066--4076.

\bibitem{causalfairsurvey2022}
K.~Makhlouf, S.~Zhioua, and C.~Palamidessi, ``Survey on causal-based machine
  learning fairness notions,'' \emph{CoRR}, vol. abs/2010.09553, 2020.

\bibitem{disengcn2019icml}
J.~Ma, P.~Cui, K.~Kuang, X.~Wang, and W.~Zhu, ``Disentangled graph
  convolutional networks,'' in \emph{Proc. 36th Int. Conf. Mach. Learn.},
  vol.~97, 2019, pp. 4212--4221.

\bibitem{ig2020nips}
B.~S{\'{a}}nchez{-}Lengeling, J.~N. Wei, B.~K. Lee, E.~Reif, P.~Wang, W.~W.
  Qian, K.~McCloskey, L.~J. Colwell, and A.~B. Wiltschko, ``Evaluating
  attribution for graph neural networks,'' in \emph{Adv. Neural Inf. Process.
  Syst. 33}, 2020, pp. 5898--5910.

\bibitem{gradcam2019cvpr}
P.~E. Pope, S.~Kolouri, M.~Rostami, C.~E. Martin, and H.~Hoffmann,
  ``Explainability methods for graph convolutional neural networks,'' in
  \emph{2019 IEEE/CVF Conf. Comput. Vision Pattern Recognit.}, 2019, pp.
  10\,772--10\,781.

\bibitem{DSE2022}
Y.~Wu, X.~Wang, A.~Zhang, X.~Hu, F.~Feng, X.~He, and T.~Chua, ``Deconfounding
  to explanation evaluation in graph neural networks,'' \emph{CoRR}, vol.
  abs/2201.08802, 2022.

\bibitem{align2023wsdm}
T.~Zhao, D.~Luo, X.~Zhang, and S.~Wang, ``Towards faithful and consistent
  explanations for graph neural networks,'' in \emph{Proc. 16th {ACM} Int.
  Conf. Web Search and Data Mining}, 2023, p. 634–642.

\bibitem{gat}
P.~Velickovic, G.~Cucurull, A.~Casanova, A.~Romero, P.~Li{\`{o}}, and
  Y.~Bengio, ``Graph attention networks,'' in \emph{6th Int. Conf. Learn.
  Representations}, 2018.

\bibitem{meta_CRL2020iclr}
Y.~Bengio, T.~Deleu, N.~Rahaman, N.~R. Ke, S.~Lachapelle, O.~Bilaniuk,
  A.~Goyal, and C.~J. Pal, ``A meta-transfer objective for learning to
  disentangle causal mechanisms,'' in \emph{8th Int. Conf. Learn.
  Representations}, 2020.

\bibitem{weaklyCRL2022jmlr}
X.~Shen, F.~Liu, H.~Dong, Q.~Lian, Z.~Chen, and T.~Zhang, ``Weakly supervised
  disentangled generative causal representation learning,'' \emph{J. Mach.
  Learn. Res.}, vol.~23, pp. 1--55, 2022.

\bibitem{RCGRL2022}
H.~Gao, J.~Li, W.~Qiang, L.~Si, B.~Xu, C.~Zheng, and F.~Sun, ``Robust causal
  graph representation learning against confounding effects,'' \emph{CoRR},
  vol. abs/2208.08584, 2022.

\bibitem{stablegnn2021}
S.~Fan, X.~Wang, C.~Shi, P.~Cui, and B.~Wang, ``Generalizing graph neural
  networks on out-of-distribution graphs,'' \emph{CoRR}, vol. abs/2111.10657,
  2021.

\bibitem{L2R-GNN2024aaai}
Z.~Chen, T.~Xiao, K.~Kuang, Z.~Lv, M.~Zhang, J.~Yang, C.~Lu, H.~Yang, and
  F.~Wu, ``Learning to reweight for generalizable graph neural network,'' in
  \emph{Proc. 38th {AAAI} Conf. Artif. Intell.}, 2024, pp. 8320--8328.

\bibitem{csa2023ijcai}
H.~Wang, J.~Chen, L.~Du, Q.~Fu, S.~Han, and X.~Song, ``Causal-based supervision
  of attention in graph neural network: {A} better and simpler choice towards
  powerful attention,'' in \emph{Proc. 32nd Int. Joint Conf. Artif. Intell.},
  2023, pp. 2315--2323.

\bibitem{DCERD2023kdd}
K.~Zhang, J.~Yu, H.~Shi, J.~Liang, and X.~Zhang, ``Rumor detection with diverse
  counterfactual evidence,'' in \emph{Proc. 29th {ACM} {SIGKDD} Conf. Knowl.
  Discov. and Data Mining}, 2023, pp. 3321--3331.

\bibitem{mccnifty2021icdm}
X.~Zhang, L.~Zhang, B.~Jin, and X.~Lu, ``A multi-view confidence-calibrated
  framework for fair and stable graph representation learning,'' in \emph{2021
  IEEE Int. Conf. Data Mining}, 2021, pp. 1493--1498.

\bibitem{rcexplainer2021nips}
M.~Bajaj, L.~Chu, Z.~Y. Xue, J.~Pei, L.~Wang, P.~C. Lam, and Y.~Zhang, ``Robust
  counterfactual explanations on graph neural networks,'' in \emph{Adv. Neural
  Inf. Process. Syst. 34}, 2021, pp. 5644--5655.

\bibitem{cflp2022icml}
T.~Zhao, G.~Liu, D.~Wang, W.~Yu, and M.~Jiang, ``Learning from counterfactual
  links for link prediction,'' in \emph{Proc. 39th Int. Conf. Mach. Learn.},
  vol. 162.\hskip 1em plus 0.5em minus 0.4em\relax PMLR, 2022, pp.
  26\,911--26\,926.

\bibitem{rfcgnn2023icdm}
Z.~Wang, G.~Narasimhan, X.~Yao, and W.~Zhang, ``Mitigating multisource biases
  in graph neural networks via real counterfactual samples,'' in \emph{2023
  {IEEE} Int. Conf. Data Mining}, 2023, pp. 638--647.

\bibitem{graphcff2024}
\BIBentryALTinterwordspacing
H.~Ling, Z.~Jiang, N.~Zou, and S.~Ji, ``Counterfactual fairness on graphs:
  Augmentations, hidden confounders, and identifiability,'' 2024. [Online].
  Available: \url{https://openreview.net/forum?id=lr0byX2aNO}
\BIBentrySTDinterwordspacing

\bibitem{cfgnnexplainer2022aistat}
A.~Lucic, M.~A. ter Hoeve, G.~Tolomei, M.~de~Rijke, and F.~Silvestri,
  ``Cf-gnnexplainer: Counterfactual explanations for graph neural networks,''
  in \emph{Proc. 25th Int. Conf. Artif. Intell. and Statist.}, vol. 151, 2022,
  pp. 4499--4511.

\bibitem{meg2021ijcnn}
D.~Numeroso and D.~Bacciu, ``Meg: Generating molecular counterfactual
  explanations for deep graph networks,'' in \emph{2021 Int. Joint Conf. Neural
  Netw.}, 2021, pp. 1--8.

\bibitem{clear2022nips}
J.~Ma, R.~Guo, S.~Mishra, A.~Zhang, and J.~Li, ``{CLEAR:} generative
  counterfactual explanations on graphs,'' in \emph{Adv. Neural Inf. Process.
  Syst. 35}, 2022, pp. 25\,895--25\,907.

\bibitem{obsdbs2021kdd}
C.~Abrate and F.~Bonchi, ``Counterfactual graphs for explainable classification
  of brain networks,'' in \emph{Proc. 27th ACM SIGKDD Conf. Knowl. Discov. Data
  Mining}, 2021, pp. 2495--2504.

\bibitem{moexplanation2021icdm}
Y.~Liu, C.~Chen, Y.~Liu, X.~Zhang, and S.~Xie, ``Multi-objective explanations
  of {GNN} predictions,'' in \emph{2021 {IEEE} Int. Conf. Data Mining}, 2021,
  pp. 409--418.

\bibitem{GCFExplainer2023wsdm}
Z.~Huang, M.~Kosan, S.~Medya, S.~Ranu, and A.~K. Singh, ``Global counterfactual
  explainer for graph neural networks,'' in \emph{Proc. 16th {ACM} Int. Conf.
  Web Search and Data Mining}, 2023, pp. 141--149.

\bibitem{banzhaf2024www}
C.~Chhablani, S.~Jain, A.~Channesh, I.~A. Kash, and S.~Medya, ``Game-theoretic
  counterfactual explanation for graph neural networks,'' \emph{CoRR}, vol.
  abs/2402.06030, 2024.

\bibitem{inmvie2024fortune}
G.~Zhang, Y.~Chen, S.~Wang, K.~Wang, and J.~Fang, ``Fortune favors the
  invariant: Enhancing gnns’ generalizability with invariant graph
  learning,'' \emph{Knowledge-Based Systems}, p. 111620, 2024.

\bibitem{lisa2023cvpr}
J.~Yu, J.~Liang, and R.~He, ``Mind the label shift of augmentation-based graph
  {OOD} generalization,'' in \emph{2023 IEEE/CVF Conf. Comput. Vision Pattern
  Recognit.}, 2023, pp. 11\,620--11\,630.

\bibitem{GIL2022nips}
H.~Li, Z.~Zhang, X.~Wang, and W.~Zhu, ``Learning invariant graph
  representations for out-of-distribution generalization,'' in \emph{Adv.
  Neural Inf. Process. Syst. 35}, 2022, pp. 11\,828--11\,841.

\bibitem{BAGNN2022icde}
Z.~Chen, T.~Xiao, and K.~Kuang, ``{BA-GNN:} on learning bias-aware graph neural
  network,'' in \emph{2022 IEEE 38th Int. Conf. Data Eng.}, 2022, pp.
  3012--3024.

\bibitem{CAL2022kdd}
Y.~Sui, X.~Wang, J.~Wu, M.~Lin, X.~He, and T.~Chua, ``Causal attention for
  interpretable and generalizable graph classification,'' in \emph{Proc. 28th
  ACM SIGKDD Conf. Knowl. Discov. Data Mining}, 2022, pp. 1696--1705.

\bibitem{cmrl2023kdd}
N.~Lee, K.~Yoon, G.~S. Na, S.~Kim, and C.~Park, ``Shift-robust molecular
  relational learning with causal substructure,'' in \emph{Proc. 29th {ACM}
  {SIGKDD} Conf. Knowl. Discov. Data Mining}, 2023, pp. 1200--1212.

\bibitem{icl2024aaai}
Z.~Zhao, P.~Wang, H.~Wen, Y.~Zhang, Z.~Zhou, and Y.~Wang, ``A twist for graph
  classification: Optimizing causal information flow in graph neural
  networks,'' in \emph{Proc. 38th {AAAI} Conf. Artif. Intell.}, 2024, pp.
  17\,042--17\,050.

\bibitem{moleood2022nips}
N.~Yang, K.~Zeng, Q.~Wu, X.~Jia, and J.~Yan, ``Learning substructure invariance
  for out-of-distribution molecular representations,'' in \emph{Adv. Neural
  Inf. Process. Syst. 35}, 2022, pp. 12\,964--12\,978.

\bibitem{canet2024www}
Q.~Wu, F.~Nie, C.~Yang, T.~Bao, and J.~Yan, ``Graph out-of-distribution
  generalization via causal intervention,'' \emph{CoRR}, vol. abs/2402.11494,
  2024.

\bibitem{cie2023cikm}
G.~Chen, Y.~Wang, F.~Guo, Q.~Guo, J.~Shao, H.~Shen, and X.~Cheng, ``Causality
  and independence enhancement for biased node classification,'' in \emph{Proc.
  32nd {ACM} Int. Conf. Inf. Knowl. Manage.}, 2023, pp. 203--212.

\bibitem{ci-gnn2023}
K.~Zheng, S.~Yu, and B.~Chen, ``{CI-GNN:} {A} granger causality-inspired graph
  neural network for interpretable brain network-based psychiatric diagnosis,''
  \emph{CoRR}, vol. abs/2301.01642, 2023.

\bibitem{rcgnn2024}
J.~Rao, J.~Xie, H.~Lin, S.~Zheng, Z.~Wang, and Y.~Yang, ``Incorporating
  retrieval-based causal learning with information bottlenecks for
  interpretable graph neural networks,'' \emph{CoRR}, vol. abs/2402.04710,
  2024.

\bibitem{gala2023nips}
Y.~Chen, Y.~Bian, K.~Zhou, B.~Xie, B.~Han, and J.~Cheng, ``Does invariant graph
  learning via environment augmentation learn invariance?'' \emph{CoRR}, vol.
  abs/2310.19035, 2023.

\bibitem{caf2023cikm}
Z.~Guo, J.~Li, T.~Xiao, Y.~Ma, and S.~Wang, ``Towards fair graph neural
  networks via graph counterfactual,'' in \emph{Proc. 32nd {ACM} Int. Conf.
  Inf. Knowl. Manage.}, 2023, pp. 669--678.

\bibitem{leci2023nips}
S.~Gui, M.~Liu, X.~Li, Y.~Luo, and S.~Ji, ``Joint learning of label and
  environment causal independence for graph out-of-distribution
  generalization,'' in \emph{Advances in Neural Information Processing Systems
  36}, 2023.

\bibitem{pnsis2024}
X.~Chen, R.~Cai, K.~Zheng, Z.~Jiang, Z.~Huang, Z.~Hao, and Z.~Li, ``Unifying
  invariance and spuriousity for graph out-of-distribution via probability of
  necessity and sufficiency,'' \emph{CoRR}, vol. abs/2402.09165, 2024.

\bibitem{RGCL2022icml}
S.~Li, X.~Wang, A.~Zhang, Y.~Wu, X.~He, and T.~Chua, ``Let invariant rationale
  discovery inspire graph contrastive learning,'' in \emph{Proc. 39th Int.
  Conf. Mach. Learn.}, vol. 162, 2022, pp. 13\,052--13\,065.

\bibitem{imold2023nips}
X.~Zhuang, Q.~Zhang, K.~Ding, Y.~Bian, X.~Wang, J.~Lv, H.~Chen, and H.~Chen,
  ``Learning invariant molecular representation in latent discrete space,''
  \emph{CoRR}, vol. abs/2310.14170, 2023.

\bibitem{cgc2023www}
H.~Yang, H.~Chen, S.~Zhang, X.~Sun, Q.~Li, X.~Zhao, and G.~Xu, ``Generating
  counterfactual hard negative samples for graph contrastive learning,'' in
  \emph{Proc. ACM Web Conf. 2023}, 2023, pp. 621--629.

\bibitem{gcil2024aaai}
Y.~Mo, X.~Wang, S.~Fan, and C.~Shi, ``Graph contrastive invariant learning from
  the causal perspective,'' in \emph{Proc. 38th {AAAI} Conf. Artif. Intell.},
  2024, pp. 8904--8912.

\bibitem{flood2023kdd}
Y.~Liu, X.~Ao, F.~Feng, Y.~Ma, K.~Li, T.~Chua, and Q.~He, ``{FLOOD:} {A}
  flexible invariant learning framework for out-of-distribution generalization
  on graphs,'' in \emph{Proc. 29th {ACM} {SIGKDD} Conf. Knowl. Discov. Data
  Mining}, 2023, pp. 1548--1558.

\bibitem{IVsurvey2022}
A.~Wu, K.~Kuang, R.~Xiong, and F.~Wu, ``Instrumental variables in causal
  inference and machine learning: {A} survey,'' \emph{CoRR}, vol.
  abs/2212.05778, 2022.

\bibitem{ANM2008nips}
P.~O. Hoyer, D.~Janzing, J.~M. Mooij, J.~Peters, and B.~Sch{\"{o}}lkopf,
  ``Nonlinear causal discovery with additive noise models,'' in \emph{Adv.
  Neural Inf. Process. Syst. 21}, 2008, pp. 689--696.

\bibitem{IV2003ai}
C.~Ai and X.~Chen, ``Efficient estimation of models with conditional moment
  restrictions containing unknown functions,'' \emph{Econometrica}, vol.~71,
  no.~6, pp. 1795--1843, 2003.

\bibitem{vgae2016}
T.~N. Kipf and M.~Welling, ``Variational graph auto-encoders,'' \emph{CoRR},
  vol. abs/1611.07308, 2016.

\bibitem{infonce2018}
A.~van~den Oord, Y.~Li, and O.~Vinyals, ``Representation learning with
  contrastive predictive coding,'' \emph{CoRR}, vol. abs/1807.03748, 2018.

\bibitem{stable_2018}
K.~Kuang, P.~Cui, S.~Athey, R.~Xiong, and B.~Li, ``Stable prediction across
  unknown environments,'' in \emph{Proc. 24th ACM SIGKDD Int. Conf. Knowl.
  Discov. Data Mining}, 2018, pp. 1617--1626.

\bibitem{stable_misspecification_2020}
K.~Kuang, R.~Xiong, P.~Cui, S.~Athey, and B.~Li, ``Stable prediction with model
  misspecification and agnostic distribution shift,'' in \emph{Proc. 34th
  {AAAI} Conf. Artif. Intell.}, vol.~34, no.~04, 2020, pp. 4485--4492.

\bibitem{dvd2020kdd}
Z.~Shen, P.~Cui, J.~Liu, T.~Zhang, B.~Li, and Z.~Chen, ``Stable learning via
  differentiated variable decorrelation,'' in \emph{Proc. 26th ACM SIGKDD Int.
  Conf. Knowl. Discov. Data Mining}, 2020, pp. 2185--2193.

\bibitem{deepstable2021cvpr}
X.~Zhang, P.~Cui, R.~Xu, L.~Zhou, Y.~He, and Z.~Shen, ``Deep stable learning
  for out-of-distribution generalization,'' in \emph{2021 IEEE/CVF Conf.
  Comput. Vision Pattern Recognit.}, 2021, pp. 5372--5382.

\bibitem{hsic2005}
A.~Gretton, O.~Bousquet, A.~J. Smola, and B.~Sch{\"{o}}lkopf, ``Measuring
  statistical dependence with hilbert-schmidt norms,'' in \emph{Proc. 16th Int.
  Conf. Algorithmic Learning Theory}, vol. 3734.\hskip 1em plus 0.5em minus
  0.4em\relax Springer, 2005, pp. 63--77.

\bibitem{reinforced_explainer2023tpami}
X.~Wang, Y.~Wu, A.~Zhang, F.~Feng, X.~He, and T.~Chua, ``Reinforced causal
  explainer for graph neural networks,'' \emph{IEEE Trans. Pattern Anal. Mach.
  Intell.}, vol.~45, no.~2, pp. 2297--2309, 2023.

\bibitem{RL4graphgen20218nips}
J.~You, B.~Liu, Z.~Ying, V.~S. Pande, and J.~Leskovec, ``Graph convolutional
  policy network for goal-directed molecular graph generation,'' in \emph{Adv.
  Neural Inf. Process. Syst. 31}, 2018, pp. 6412--6422.

\bibitem{general_gat}
G.~Wang, R.~Ying, J.~Huang, and J.~Leskovec, ``Improving graph attention
  networks with large margin-based constraints,'' \emph{CoRR}, vol.
  abs/1910.11945, 2019.

\bibitem{nnm1973biometrics}
D.~B. Rubin, ``Matching to remove bias in observational studies,''
  \emph{Biometrics}, vol.~29, no.~1, pp. 159--183, 1973.

\bibitem{propensity1983biometrika}
P.~R. Rosenbaum and D.~B. Rubin, ``The central role of the propensity score in
  observational studies for causal effects,'' \emph{Biometrika}, vol.~70,
  no.~1, pp. 41--55, 1983.

\bibitem{balancing_counterfactual_regression2016icml}
F.~D. Johansson, U.~Shalit, and D.~A. Sontag, ``Learning representations for
  counterfactual inference,'' in \emph{Proc. 33rd Int. Conf. Mach. Learn.},
  vol.~48, 2016, pp. 3020--3029.

\bibitem{cgmsurvey2023}
G.~Zhou, L.~Yao, X.~Xu, C.~Wang, L.~Zhu, and K.~Zhang, ``On the opportunity of
  causal deep generative models: {A} survey and future directions,''
  \emph{CoRR}, vol. abs/2301.12351, 2023.

\bibitem{gaussiandgm2022nips}
B.~Kivva, G.~Rajendran, P.~Ravikumar, and B.~Aragam, ``Identifiability of deep
  generative models without auxiliary information,'' in \emph{Advances in
  Neural Information Processing Systems 35: Annual Conference on Neural
  Information Processing Systems 2022, NeurIPS 2022, New Orleans, LA, USA,
  November 28 - December 9, 2022}, 2022.

\bibitem{cff2022www}
J.~Tan, S.~Geng, Z.~Fu, Y.~Ge, S.~Xu, Y.~Li, and Y.~Zhang, ``Learning and
  evaluating graph neural network explanations based on counterfactual and
  factual reasoning,'' in \emph{Proc. ACM Web Conf. 2022}, 2022, pp.
  1018--1027.

\bibitem{rsggce2024aaai}
M.~A. Prado{-}Romero, B.~Prenkaj, and G.~Stilo, ``Robust stochastic graph
  generator for counterfactual explanations,'' in \emph{Proc. 38th {AAAI} Conf.
  Artif. Intell.}, 2024, pp. 21\,518--21\,526.

\bibitem{xgnn2020kdd}
H.~Yuan, J.~Tang, X.~Hu, and S.~Ji, ``{XGNN:} towards model-level explanations
  of graph neural networks,'' in \emph{Proc. 26th ACM SIGKDD Int. Conf. Knowl.
  Discov. Data Mining}, 2020, pp. 430--438.

\bibitem{vrrw1992}
R.~Pemantle, ``Vertex-reinforced random walk,'' \emph{Probability Theory and
  Related Fields}, vol.~92, no.~1, pp. 117--136, 1992.

\bibitem{buhlmann2020invariance}
P.~B{\"u}hlmann, ``{Invariance, Causality and Robustness},'' \emph{Statistical
  Sci.}, vol.~35, no.~3, pp. 404 -- 426, 2020.

\bibitem{peters2016invariant_causal}
J.~Peters, P.~B{\"u}hlmann, and N.~Meinshausen, ``Causal inference by using
  invariant prediction: identification and confidence intervals,'' \emph{J.
  Roy. Statistical Soc. Ser. B: Statistical Methodology}, vol.~78, no.~5, pp.
  947--1012, 2016.

\bibitem{IRM2019}
M.~Arjovsky, L.~Bottou, I.~Gulrajani, and D.~Lopez{-}Paz, ``Invariant risk
  minimization,'' \emph{CoRR}, vol. abs/1907.02893, 2019.

\bibitem{rex2021icml}
D.~Krueger, E.~Caballero, J.~Jacobsen, A.~Zhang, J.~Binas, D.~Zhang, R.~L.
  Priol, and A.~C. Courville, ``Out-of-distribution generalization via risk
  extrapolation (rex),'' in \emph{Proc. 38th Int. Conf. Mach. Learn.}, vol.
  139, 2021, pp. 5815--5826.

\bibitem{IGA2020}
M.~Koyama and S.~Yamaguchi, ``Out-of-distribution generalization with maximal
  invariant predictor,'' \emph{CoRR}, vol. abs/2008.01883, 2020.

\bibitem{towards_causal2021}
B.~Sch{\"{o}}lkopf, F.~Locatello, S.~Bauer, N.~R. Ke, N.~Kalchbrenner,
  A.~Goyal, and Y.~Bengio, ``Towards causal representation learning,''
  \emph{CoRR}, vol. abs/2102.11107, 2021.

\bibitem{graphsage}
W.~L. Hamilton, Z.~Ying, and J.~Leskovec, ``Inductive representation learning
  on large graphs,'' in \emph{Adv. Neural Inf. Process. Syst. 30}, 2017, pp.
  1024--1034.

\bibitem{gceloss2018nips}
Z.~Zhang and M.~R. Sabuncu, ``Generalized cross entropy loss for training deep
  neural networks with noisy labels,'' in \emph{Adv. Neural Inf. Process. Syst.
  31}, 2018, pp. 8792--8802.

\bibitem{variational2017jasa}
D.~M. Blei, A.~Kucukelbir, and J.~D. McAuliffe, ``Variational inference: A
  review for statisticians,'' \emph{J. Am. Stat. Assoc.}, vol. 112, no. 518,
  pp. 859--877, 2017.

\bibitem{normhsic2019icml}
S.~Kornblith, M.~Norouzi, H.~Lee, and G.~E. Hinton, ``Similarity of neural
  network representations revisited,'' in \emph{Proc. 36th Int. Conf. Mach.
  Learn.}, vol.~97, 2019, pp. 3519--3529.

\bibitem{pns_inv2023nips}
M.~Yang, Y.~Zhang, Z.~Fang, Y.~Du, F.~Liu, J.~Ton, J.~Wang, and J.~Wang,
  ``Invariant learning via probability of sufficient and necessary causes,'' in
  \emph{Adv. Neural Inf. Process. Syst. 36}, 2023.

\bibitem{inv_spu_ens2023nips}
C.~Eastwood, S.~Singh, A.~L. Nicolicioiu, M.~V. Pogancic, J.~von
  K{\"{u}}gelgen, and B.~Sch{\"{o}}lkopf, ``Spuriosity didn't kill the
  classifier: Using invariant predictions to harness spurious features,'' in
  \emph{Adv. Neural Inf. Process. Syst. 36}, 2023.

\bibitem{graphssl2023tkde}
Y.~Liu, M.~Jin, S.~Pan, C.~Zhou, Y.~Zheng, F.~Xia, and P.~S. Yu, ``Graph
  self-supervised learning: {A} survey,'' \emph{{IEEE} Trans. Knowl. Data
  Eng.}, vol.~35, no.~6, pp. 5879--5900, 2023.

\bibitem{SimCLR2020icml}
T.~Chen, S.~Kornblith, M.~Norouzi, and G.~E. Hinton, ``A simple framework for
  contrastive learning of visual representations,'' in \emph{Proc. 37th Int.
  Conf. Mach. Learn.}, vol. 119, 2020, pp. 1597--1607.

\bibitem{floodbootstrap2020nips}
J.~Grill, F.~Strub, F.~Altch{\'{e}}, C.~Tallec, P.~H. Richemond,
  E.~Buchatskaya, C.~Doersch, B.~{\'{A}}. Pires, Z.~Guo, M.~G. Azar, B.~Piot,
  K.~Kavukcuoglu, R.~Munos, and M.~Valko, ``Bootstrap your own latent - {A} new
  approach to self-supervised learning,'' in \emph{Adv. Neural Inf. Process.
  Syst. 33}, 2020, pp. 21\,271--21\,284.

\bibitem{GCN}
T.~N. Kipf and M.~Welling, ``Semi-supervised classification with graph
  convolutional networks,'' in \emph{5th Int. Conf. Learn. Representations},
  2017.

\bibitem{ogb2020nips}
W.~Hu, M.~Fey, M.~Zitnik, Y.~Dong, H.~Ren, B.~Liu, M.~Catasta, and J.~Leskovec,
  ``Open graph benchmark: Datasets for machine learning on graphs,'' in
  \emph{Adv. Neural Inf. Process. Syst. 33}, 2020, pp. 22\,118--22\,133.

\bibitem{graphsmote2021wsdm}
T.~Zhao, X.~Zhang, and S.~Wang, ``Graphsmote: Imbalanced node classification on
  graphs with graph neural networks,'' in \emph{Proc. 14th {ACM} Int. Conf. Web
  Search and Data Mining}, 2021, pp. 833--841.

\bibitem{gretel2022cikm}
M.~A. Prado{-}Romero and G.~Stilo, ``{GRETEL:} graph counterfactual explanation
  evaluation framework,'' in \emph{Proc. 31st {ACM} Int. Conf. Inf. Knowl.
  Manage.}, 2022, pp. 4389--4393.

\bibitem{causebox2021cikm}
P.~Sheth, U.~Jeong, R.~Guo, H.~Liu, and K.~S. Candan, ``Causebox: {A} causal
  inference toolbox for benchmarkingtreatment effect estimators with machine
  learning methods,'' in \emph{Proc. 30th {ACM} Int. Conf. Inf. Knowl.
  Manage.}, 2021, pp. 4789--4793.

\bibitem{causallearn2023}
Y.~Zheng, B.~Huang, W.~Chen, J.~D. Ramsey, M.~Gong, R.~Cai, S.~Shimizu,
  P.~Spirtes, and K.~Zhang, ``Causal-learn: Causal discovery in python,''
  \emph{CoRR}, vol. abs/2307.16405, 2023.

\bibitem{alphago2021nature}
J.~Jumper, R.~Evans, A.~Pritzel, T.~Green, M.~Figurnov, O.~Ronneberger,
  K.~Tunyasuvunakool, R.~Bates, A.~{\v{Z}}{\'\i}dek, A.~Potapenko
  \emph{et~al.}, ``Highly accurate protein structure prediction with
  alphafold,'' \emph{Nature}, vol. 596, no. 7873, pp. 583--589, 2021.

\bibitem{pinsage2018kdd}
R.~Ying, R.~He, K.~Chen, P.~Eksombatchai, W.~L. Hamilton, and J.~Leskovec,
  ``Graph convolutional neural networks for web-scale recommender systems,'' in
  \emph{Proc. 24th {ACM} {SIGKDD} Int. Conf. Knowl. Discov. Data Mining}, 2018,
  pp. 974--983.

\bibitem{pasca2022www}
W.~Zhang, Y.~Shen, Z.~Lin, Y.~Li, X.~Li, W.~Ouyang, Y.~Tao, Z.~Yang, and
  B.~Cui, ``Pasca: {A} graph neural architecture search system under the
  scalable paradigm,'' in \emph{Proc. ACM Web Conf. 2022}, 2022, pp.
  1817--1828.

\bibitem{scalablegnn_benchmark2022nips}
K.~Duan, Z.~Liu, P.~Wang, W.~Zheng, K.~Zhou, T.~Chen, X.~Hu, and Z.~Wang, ``A
  comprehensive study on large-scale graph training: Benchmarking and
  rethinking,'' in \emph{Adv. Neural Inf. Process. Syst. 35}, 2022, pp.
  5376--5389.

\bibitem{gfmsurvey2023}
J.~Liu, C.~Yang, Z.~Lu, J.~Chen, Y.~Li, M.~Zhang, T.~Bai, Y.~Fang, L.~Sun,
  P.~S. Yu, and C.~Shi, ``Towards graph foundation models: {A} survey and
  beyond,'' \emph{CoRR}, vol. abs/2310.11829, 2023.

\bibitem{causal&llm2023}
E.~Kiciman, R.~Ness, A.~Sharma, and C.~Tan, ``Causal reasoning and large
  language models: Opening a new frontier for causality,'' \emph{CoRR}, vol.
  abs/2305.00050, 2023.

\bibitem{trustllm2023}
Y.~Liu, Y.~Yao, J.~Ton, X.~Zhang, R.~Guo, H.~Cheng, Y.~Klochkov, M.~F. Taufiq,
  and H.~Li, ``Trustworthy llms: a survey and guideline for evaluating large
  language models' alignment,'' \emph{CoRR}, vol. abs/2308.05374, 2023.

\bibitem{birm2022cvpr}
Y.~Lin, H.~Dong, H.~Wang, and T.~Zhang, ``Bayesian invariant risk
  minimization,'' in \emph{2022 IEEE/CVF Conf. Comput. Vision Pattern
  Recognit.}, 2022, pp. 16\,000--16\,009.

\bibitem{sparseirm2022icml}
X.~Zhou, Y.~Lin, W.~Zhang, and T.~Zhang, ``Sparse invariant risk
  minimization,'' in \emph{Proc. 39th Int. Conf. Mach. Learn.}, vol. 162, 2022,
  pp. 27\,222--27\,244.

\bibitem{inv_identifiability2022iclr}
C.~Lu, Y.~Wu, J.~M. Hern{\'{a}}ndez{-}Lobato, and B.~Sch{\"{o}}lkopf,
  ``Invariant causal representation learning for out-of-distribution
  generalization,'' in \emph{10th Int. Conf. Learn. Representations}, 2022.

\bibitem{castle2020nips}
T.~Kyono, Y.~Zhang, and M.~van~der Schaar, ``{CASTLE:} regularization via
  auxiliary causal graph discovery,'' in \emph{Adv. Neural Inf. Process. Syst.
  33}, 2020, pp. 1501--1512.

\bibitem{crcg2024aaai}
H.~Gao, C.~Yao, J.~Li, L.~Si, Y.~Jin, F.~Wu, C.~Zheng, and H.~Liu, ``Rethinking
  causal relationships learning in graph neural networks,'' in \emph{Proc. 38th
  {AAAI} Conf. Artif. Intell.}, 2024, pp. 12\,145--12\,154.

\bibitem{dcsgl2024kbs}
H.~Gao, P.~Qiao, Y.~Jin, F.~Wu, J.~Li, and C.~Zheng, ``Introducing diminutive
  causal structure into graph representation learning,'' \emph{Knowledge-Based
  Syst.}, p. 111592, 2024.

\bibitem{pathcf2019aaai}
S.~Chiappa, ``Path-specific counterfactual fairness,'' in \emph{Proc. 33rd
  {AAAI} Conf. Artif. Intell.}, vol.~33, no.~01, 2019, pp. 7801--7808.

\bibitem{interventionfair2012tkde}
S.~Hajian and J.~Domingo{-}Ferrer, ``A methodology for direct and indirect
  discrimination prevention in data mining,'' \emph{IEEE Trans. Knowl. Data
  Eng.}, vol.~25, no.~7, pp. 1445--1459, 2013.

\bibitem{interventionfair2018aaai}
J.~Zhang and E.~Bareinboim, ``Fairness in decision-making - the causal
  explanation formula,'' in \emph{Proc. 32nd {AAAI} Conf. Artif. Intell.},
  vol.~32, no.~1, 2018, pp. 2037--2045.

\bibitem{fedgnn2022nc}
C.~Wu, F.~Wu, L.~Lyu, T.~Qi, Y.~Huang, and X.~Xie, ``A federated graph neural
  network framework for privacy-preserving personalization,'' \emph{Nature
  Commun.}, vol.~13, no.~1, p. 3091, 2022.

\bibitem{fedgnn4recsys2021arxiv}
C.~Wu, F.~Wu, Y.~Cao, Y.~Huang, and X.~Xie, ``Fedgnn: Federated graph neural
  network for privacy-preserving recommendation,'' \emph{CoRR}, vol.
  abs/2102.04925, 2021.

\bibitem{ce_privacy2023kdd}
V.~Vo, T.~Le, V.~Nguyen, H.~Zhao, E.~V. Bonilla, G.~Haffari, and D.~Q. Phung,
  ``Feature-based learning for diverse and privacy-preserving counterfactual
  explanations,'' in \emph{Proc. 29th {ACM} {SIGKDD} Conf. Knowl. Discov. and
  Data Mining}, 2023, pp. 2211--2222.

\bibitem{ce_manipulate2021nips}
D.~Slack, A.~Hilgard, H.~Lakkaraju, and S.~Singh, ``Counterfactual explanations
  can be manipulated,'' in \emph{Adv. Neural Inf. Process. Syst. 34}, 2021, pp.
  62--75.

\bibitem{robust_fair_ce2023cikm}
Y.~Wang, H.~Qian, Y.~Liu, W.~Guo, and C.~Miao, ``Flexible and robust
  counterfactual explanations with minimal satisfiable perturbations,'' in
  \emph{Proc. 32nd {ACM} Int. Conf. Inf. Knowl. Manage.}, 2023, pp. 2596--2605.

\bibitem{zin2022nips}
Y.~Lin, S.~Zhu, L.~Tan, and P.~Cui, ``{ZIN:} when and how to learn invariance
  without environment partition?'' in \emph{Advances in Neural Information
  Processing Systems 35}, 2022.

\bibitem{stcrl2023kdd}
Y.~Zhao, P.~Deng, J.~Liu, X.~Jia, and J.~Zhang, ``Generative causal
  interpretation model for spatio-temporal representation learning,'' in
  \emph{Proc. 29th {ACM} {SIGKDD} Conf. Knowl. Discov. and Data Mining}, 2023,
  pp. 3537--3548.

\bibitem{drugood2022}
Y.~Ji, L.~Zhang, J.~Wu, B.~Wu, L.~Huang, T.~Xu, Y.~Rong, L.~Li, J.~Ren, D.~Xue,
  H.~Lai, S.~Xu, J.~Feng, W.~Liu, P.~Luo, S.~Zhou, J.~Huang, P.~Zhao, and
  Y.~Bian, ``Drugood: Out-of-distribution {(OOD)} dataset curator and benchmark
  for ai-aided drug discovery - {A} focus on affinity prediction problems with
  noise annotations,'' \emph{CoRR}, vol. abs/2201.09637, 2022.

\bibitem{ood_kinetic2023nips}
Z.~Wang, Y.~Chen, Y.~Duan, W.~Li, B.~Han, J.~Cheng, and H.~Tong, ``Towards
  out-of-distribution generalizable predictions of chemical kinetics
  properties,'' \emph{CoRR}, vol. abs/2310.03152, 2023.

\bibitem{fairgraphbench2024}
X.~Qian, Z.~Guo, J.~Li, H.~Mao, B.~Li, S.~Wang, and Y.~Ma, ``Addressing
  shortcomings in fair graph learning datasets: Towards a new benchmark,''
  \emph{CoRR}, vol. abs/2403.06017, 2024.

\bibitem{oodmetric2021}
H.~Ye, C.~Xie, Y.~Liu, and Z.~Li, ``Out-of-distribution generalization analysis
  via influence function,'' \emph{CoRR}, vol. abs/2101.08521, 2021.

\bibitem{dro2018}
J.~C. Duchi and H.~Namkoong, ``Learning models with uniform performance via
  distributionally robust optimization,'' \emph{CoRR}, vol. abs/1810.08750,
  2018.

\bibitem{sp_metric2012itcs}
C.~Dwork, M.~Hardt, T.~Pitassi, O.~Reingold, and R.~S. Zemel, ``Fairness
  through awareness,'' in \emph{Innov. Theor. Comput. Sci. 2012}, 2012, pp.
  214--226.

\bibitem{eo_metric2016nips}
M.~Hardt, E.~Price, and N.~Srebro, ``Equality of opportunity in supervised
  learning,'' in \emph{Adv. Neural Inf. Process. Syst. 29}, 2016, pp.
  3315--3323.

\bibitem{maccs2022chemistry}
G.~P. Wellawatte, A.~Seshadri, and A.~D. White, ``Model agnostic generation of
  counterfactual explanations for molecules,'' \emph{Chem. Sci.}, vol.~13,
  no.~13, pp. 3697--3705, 2022.

\end{thebibliography}

\newpage

%

\appendices

\section{Notations} \label{app:notations}
Generally, we use bold uppercase letters (e.g., $\mathbf{A}$) to denote matrices, bold lowercase letters (e.g., $\mathbf{a}$) to denote vectors and normal lowercase letters (e.g., $a$) to denote real-valued numbers. 
For a matrix, \eg $\mathbf{A}$, $a_{i,j}$ denotes its $(i, j)$-th entry, $\mathbf{a}_{,j}$ denotes its $j$-th column and $\mathbf{a}_i$ denotes its $i$-th row.
In addition, letters in calligraphy font (e.g., $\mathcal{V}$) denote sets.
More frequently used notations are summarized in \tabref{tab:notations}.

\begin{table}[!ht]\scriptsize
    \centering
    \caption{Notations and definitions or descriptions.}
    \begin{tabular}{l|l}
    \toprule
        \textbf{Notations} & \textbf{Definitions or Descriptions} \\ \midrule \midrule
        $\mathcal{G}$ & The random variable representing a graph. \\ \hline
        $\mathcal{G}^I (\mathcal{G}^V)$ & The invariant (variant) component of graph $\mathcal{G}$. \\ \hline
        $\mathcal{G}_u$ & The random variable representing the ego-graph of node $u$. \\ \hline
        $G$ & A graph instance. \\ \hline
        $\mathbf{X}$ & The node attribute matrix of graph $\mathcal{G}$. \\ \hline
        $\mathbf{H}$ & The hidden representation matrix of graph $\mathcal{G}$. \\ \hline
        $\mathbf{h}$ & The hidden representation vector of a node or a whole graph. \\ \hline
        $\mathbf{h}^{\mathcal{G}}$ & The hidden representation of a whole graph $\mathcal{\mathcal{G}}$. \\ \hline
        $\mathbf{h}^I$ & The invariant representation vector of a node or a whole graph. \\ \hline
        $\mathbf{h}^V$ & The variant representation vector of a node or a whole graph. \\ \hline
        $\mathbf{A}$ & The adjacent matrix of graph $\mathcal{G}$. \\ \hline
        $\mathbf{\overline{A}}$ & The adjacency matrix of the complement of graph $\mathcal{G}$. \\ \hline
        $Y$ & The label of a node or a graph. \\ \hline
        $\mathcal{D}_{tr} (\mathcal{D}_{te})$ & The training (testing) dataset. \\ 
        \midrule \midrule
        $\Phi(\cdot)$ & A GNN encoder. \\ \hline
        $\Phi^I(\cdot)$ & The GNN encoder that generates invariant representation. \\ \hline
        $\Phi^V(\cdot)$ & The GNN encoder that generates variant representation. \\ \hline
        $w(\cdot)$ & The predictor for a downstream task. \\ \hline
        $w^I(\cdot)$ & The predictor fed with invariant representation. \\ \hline
        $w^V(\cdot)$ & The predictor fed with variant representation. \\ 
        \midrule \midrule
        $\mathcal{L}(\cdot, \cdot)$ & The expectation of losses over all training samples. \\ \hline
        $l(\cdot, \cdot)$ & The loss of a training sample. \\ \hline
        $\mathbb{I}(\cdot)$ & The indicator function. \\ \hline
        $\text{MI}(\cdot, \cdot)$ & The mutual information function. \\
        \bottomrule
    \end{tabular}
    \label{tab:notations}
\end{table}

\section{Summary of the Reviewed CIGNNs}
\label{app:work_summary}
We provide a comparison of the reviewed CIGNNs from multiple significant dimensions in \tabref{tab:work_summary}.
\begin{table*}[t]\scriptsize
    \centering
    \caption{Summary of the reviewed CIGNNs in chronological order.}
    \begin{tabular}{l|l|l|l|c}
    \toprule
        \textbf{Method} & \textbf{Trustworthiness Risk} & \textbf{Graph Task} & \textbf{Dataset Domain} & \textbf{Code} \\ \hline
        RC-Explainer~\cite{reinforced_explainer2023tpami} & Explainability & Graph & Molecule, Social Network, Image & \href{https://github.com/xiangwang1223/reinforced_causal_explainer}{Link} \\ \hline
        Gem~\cite{gem2021icml} & Explainability & Graph & Synthetic, Molecule & \href{https://github.com/wanyu- lin/ICML2021-Gem}{Link} \\ \hline
        OBS and DBS~\cite{obsdbs2021kdd} & Explainability & Graph & Brain Network & \href{https://github.com/carlo-abrate/CounterfactualGraphs}{Link} \\ \hline
        DGNN~\cite{debiasedgnn2022tnnls} & OOD & Node & Citation, Knowledge Graph & - \\ \hline
        CF-GNNExplainer~\cite{cfgnnexplainer2022aistat} & Explainability & Node & Synthetic & \href{https://github.com/a-lucic/cf-gnnexplainer}{Link} \\ \hline
        NIFTY~\cite{nifty2021uai} & Fairness & Node & 
        Loan Applications, Criminal Justice, Credit Defaulter
        & - \\ \hline
        RCExplainer~\cite{rcexplainer2021nips} & Explainability & Node, Graph & Synthetic, Molecule & \href{https://github.com/RomanOort/FACTAI/blob/4e3db257cdae8124ee1cef66e7c01258edd6617b/packages/gcn_interpretation/gnn-model-explainer/explainer/explain_rcexplainer.py}{Link} \\ \hline
        MCCNIFTY~\cite{mccnifty2021icdm} & Fairness & Node & Loan Applications, Criminal Justice, Credit Defaulter & - \\ \hline
        GNN-MOExp~\cite{moexplanation2021icdm} & Explainability & Node & Citation, Social Network, Co-purchase, Co-author & - \\ \hline
        GEAR~\cite{gear2022wsdm} & Fairness & Node & Synthetic, Loan Applications, Criminal Justice, Credit Defaulter & - \\ \hline
        EERM~\cite{EERM2022iclr} & OOD & Node & Citation, Social Network, Transaction & \href{https://github.com/qitianwu/GraphOOD-EERM}{Link} \\ \hline
        DIR~\cite{DIR2022iclr} & OOD, Explainability & Graph & Synthetic, Image, Text Sentiment, Molecule & \href{https://github.com/Wuyxin/DIR-GNN}{Link} \\ \hline
        CF$^2$~\cite{cff2022www} & Explainability & Node, Graph & Synthetic, Molecule, Citation & - \\ \hline
        StableGNN~\cite{stablegnn2021} & OOD & Graph & Synthetic, Molecule & \href{https://github.com/googlebaba/StableGNN}{Link} \\ \hline
        OrphicX~\cite{orphicx2022cvpr} & Explainability & Graph & Synthetic, Molecule & \href{https://github.com/WanyuGroup/CVPR2022-OrphicX}{Link} \\ \hline
        BA-GNN~\cite{BAGNN2022icde} & OOD & Node & Citation, Co-Author, Co-Purchase, Web Link & - \\ \hline
        OOD-GNN~\cite{oodgnn2022tkde} & OOD & Graph & Synthetic, Image, Molecule &  - \\ \hline
        CFLP~\cite{cflp2022icml} & OOD & Edge & Citation, Social Network, Drug Discovery &  \href{https://github.com/DM2-ND/CFLP}{Link} \\ \hline
        DSE~\cite{DSE2022} & Explainability & Graph & Synthetic, Image, Text Sentiment &  - \\ \hline
        CAL~\cite{CAL2022kdd} & OOD, Explainability & Graph & Synthetic, Molecule, Social Network, Image &  \href{https://github.com/yongduosui/CAL}{Link} \\ \hline
        GIL~\cite{GIL2022nips} & OOD, Explainability & Graph & Synthetic, Image, Text Sentiment, Molecule &  - \\ \hline
        CIGA~\cite{ciga2022nips} & OOD, Explainability & Graph & Synthetic, Drug Discovery, Image, Text Sentiment &  \href{https://github.com/LFhase/CIGA}{Link} \\ \hline
        CLEAR~\cite{clear2022nips} & Explainability & Node, Graph & Synthetic, Molecule, Social Network &  - \\ \hline
        DisC~\cite{DisC2022nips} & OOD, Explainability & Graph & Image &  \href{https://github.com/googlebaba/DisC}{Link} \\ \hline
        MoleOOD~\cite{moleood2022nips} & OOD & Graph & Molecule &  \href{https://github.com/yangnianzu0515/MoleOOD}{Link} \\ \hline
        RGCL~\cite{RGCL2022icml} & OOD, Explainability & Graph & Molecule, Image & \href{https: //github.com/lsh0520/RGCL}{Link} \\ \hline
        GCFExplainer~\cite{GCFExplainer2023wsdm} & Explainability & Graph & Molecule, Protein &  \href{https://github.com/mertkosan/GCFExplainer}{Link} \\ \hline
        RCGRL~\cite{RCGRL2022} & OOD & Graph & Synthetic, Text Sentiment, Social Network, Molecule &  \href{https://github.com/hang53/RCGRL}{Link} \\ \hline
        CIE~\cite{cie2023cikm} & OOD & Node & Citation, Web Link, Social Network &  \href{https://github.com/Chen-GX/CIE}{Link} \\ \hline
        CAF~\cite{caf2023cikm} & Fairness & Node & Loan Applications, Criminal Justice, Credit Defaulter & \href{https://github.com/TimeLovercc/CAF-GNN}{Link} \\ \hline
        CGC~\cite{cgc2023www} & OOD & Graph & Synthetic, Protein, Molecule, Social Network & \href{https://www.dropbox.com/sh/kyf8p9unkhn0r99/AABd33jFBf jGYIkvIqWpuNwYa?dl=0}{Link} \\ \hline
        LiSA~\cite{lisa2023cvpr} & OOD & Node, Graph & Synthetic, Molecule, Image, Social Network, Citation, Financial & \href{https://github.com/Samyu0304/LiSA}{Link} \\ \hline
        CSA~\cite{csa2023ijcai} & OOD & Node & Web Link, Citation & - \\ \hline
        FLOOD~\cite{flood2023kdd} & OOD & Node & Web Link, Citation &  - \\ \hline
        DCE-RD~\cite{DCERD2023kdd} & OOD, Explainability & Graph & Social Network &  \href{https://github.com/Vicinity111/DCE-RD}{Link} \\ \hline
        CMRL~\cite{cmrl2023kdd} & OOD, Explainability & Graph & Molecule &  \href{https://github.com/Namkyeong/CMRL}{Link} \\ \hline
        iMoLD~\cite{imold2023nips} & OOD & Graph & Molecule &  \href{https://github.com/HICAI-ZJU/iMoLD}{Link} \\ \hline
        GALA~\cite{gala2023nips} & OOD, Explainability & Graph & Drug, Image, Text Sentiment &  \href{https://github.com/LFhase/GALA}{Link} \\ \hline
        LECI~\cite{leci2023nips} & OOD, Explainability & Graph & Synthetic, Image, Text Sentiment, Molecule, Drug  & \href{https://github.com/divelab/LECI}{Link} \\ \hline
        INL~\cite{inl2023tois} & OOD, Explainability & Node & Citation, Co-Purchase, Protein & - \\ \hline
        
        RFCGNN~\cite{rfcgnn2023icdm} & Fairness & Node & Credit Default, Criminal Justice & - \\ \hline
        CI-GNN~\cite{ci-gnn2023}
        & OOD, Explainability & Graph & Synthetic, Brain Disease & \href{https://github.com/ZKZ-Brain/CI-GNN}{Link} \\ \hline
        inMvie~\cite{inmvie2024fortune} & OOD & Node & Synthetic, Citation, Social Network & - \\ \hline
        L2R-GNN~\cite{L2R-GNN2024aaai} & OOD & Graph & Synthetic, Molecule, Social Network & - \\ \hline
        ICL~\cite{icl2024aaai} & OOD, Explainability & Graph & Synthetic, Molecule, Protein, Social Network, Text, Image, Sentiment & \href{https://github.com/haibin65535/ICL}{Link} \\ \hline
        GCIL~\cite{gcil2024aaai} & OOD & Node & Citation, Social Network & \href{https://github.com/BUPT-GAMMA/GCIL}{Link} \\ \hline
        CaNet~\cite{canet2024www} & OOD & Node & Citation & \href{https://github.com/fannie1208/CaNet}{Link}\\ \hline
        Banzhaf~\cite{banzhaf2024www} & Explainability & Node & Synthetic & - \\ \hline
        GraphCFF~\cite{graphcff2024} & Fairness & Node & Synthetic, Loan Applications, Credit Defaulter, Criminal Justice & - \\ \hline
        RC-GNN~\cite{rcgnn2024} & OOD, Explainability & Graph & Synthetic, Molecule & - \\ \hline
        PNSIS~\cite{pnsis2024} & OOD, Explainability & Graph & Synthetic, Molecule & - \\ \hline

    \bottomrule
    \end{tabular}
    \label{tab:work_summary}
\end{table*}


\section{Further Discussions on Varied CIGNNs}
\subsection{Causal Reasoning on Graphs}
\subsubsection{Group-level Causal Effect Estimation on Graphs} \label{app:gte_discuss}
The IV and frontdoor adjustment approaches treat the whole input (sub)graph as a treatment variable and can estimate the treatment effect even when the confounders are unobservable. 
In contrast, while stable learning cannot guarantee its performance when unobservable confounders exist, it can measure the causal effect of each feature (cluster) from high-dimensional features simultaneously, which provides more fine-grained causal knowledge to the GNN.

\subsubsection{Individual-level Causal Effect Estimation on Graphs}
\label{app:ite_discuss}
Causal intervention approaches estimate the individual-level causal effects of graph components on model predictions, facilitated by the capability to repeatedly execute the GNN inference mechanism. 
However, they may produce counterfactual samples that do not reflect realistic graph distributions, limiting their applicability~\cite{gear2022wsdm}. Besides, given the potential gap between the causal effects on model predictions and those on ground-truth labels, using the former to guide GNN training might exacerbate model biases. 

In contrast, matching and deep generative methods measure causal effects among graph components or labels by approximating counterfactual outcomes of hypothetical graph interventions.
Nevertheless, the effectiveness of matching is constrained by the availability of high-quality matched samples and the presence of hidden confounders~\cite{propensity1983biometrika}.
The identifiability of deep generative models remains unresolved under certain data and task assumptions~\cite{graphcff2024}, making the selection of suitable models challenging.

\subsubsection{Graph Counterfactual Explanation Generation}
\label{app:gce_discuss}
Continuous optimization based methods leverage gradient information to iteratively refine counterfactual explanations, which are highly efficient and scalable but can be sensitive to the choice of optimization hyperparameters. Besides, they might converge to local minima and fail to guarantee the minimal perturbation property of the generated GCEs~\cite{cff2022www,ce_manipulate2021nips}. 
On the other hand, heuristic search based methods use a trial-and-error approach to explore the graph space for possible GCEs, which typically are more flexible and have mechanisms to prevent from being stuck in local minima. However, they often require more computational resources and have less predictable performance, as their success heavily depends on the design of the heuristic and its parameters. 
Both methods have their merits and limitations, and the choice between them may depend on the specific requirements of the task, such as the need for scalability or precision. Moreover, further efforts are demanded to improve the reliability of both types of GCE generation methods, such as robustness and fairness~\cite{robust_fair_ce2023cikm}.



\subsection{Causal Representation Learning on Graphs}
\subsubsection{Supervised Causal Representation Learning}
\label{app:super_crl_discuss}
Comparatively, group invariant learning offers theoretical causality assurances under accurate environment inference, with a straightforward optimization scheme. Yet, its effectiveness hinges on the diversity of training environments, which, if inadequate, can lead to spurious correlation leakage~\cite{towards_causal2021,zin2022nips,gala2023nips}. 
Conversely, joint invariant and variant learning approaches utilize the detailed interplay between factors $I$ and $V,$ holding the flexibility to tailor CRL to specific graph generation processes. This could circumvent the limitations seen in group invariant learning approaches. Nonetheless, the variability in assumptions about dependencies among $I$, $V$ and label $Y$ across studies can diminish these models' general applicability. Moreover, the increasing number of variant regularizers such as independence-enforcing ones complicates optimization, with a lack of comprehensive evaluations on their practical variants.

Overall, the prevailing focus in supervised approaches is on learning invariant and variant representations. Despite advancements, the field has yet to extensively explore more fine-grained causal representations, which could deepen understanding of the data generation process and support various downstream tasks~\cite{stcrl2023kdd}.

\section{Open-source GNN Trustworthiness Benchmarks}
\label{app:benchmark}
\subsection{Evaluating OOD Generalizability}
A multitude of real-world and synthetic graph datasets have been employed to assess GNNs across node-level and graph-level tasks. These datasets, encompassing diverse sources of distribution shifts, such as featural and structural diversity, are extensively utilized for evaluating graph OOD methods.
Li~\etal~\cite{oodgraphsurvey2022} provided a comprehensive summary of the popular real-world and synthetic graph datasets, along with their key statistics.
Gui and Li~\etal~\cite{good2022nips} created an advanced graph OOD benchmark, GOOD, based on open-source graph datasets for comprehensive comparison among different graph OOD methods. It contains 6 graph-level datasets and 5 node-level datasets generated by conducting no-shift, covariate shift, and concept shift splitting on existing graph datasets.
Ji and Zhang~\etal~\cite{drugood2022} curated an AI-aided drug discovery benchmark with data environment splitting aligned with biochemistry knowledge, serving as a great testbed for evaluating graph OOD generalization methods.
Wang and Chen~\etal~\cite{ood_kinetic2023nips} developed an OOD kinetic property prediction benchmark that exhibits distribution shifts in the dimension of graph structure, reaction condition and reaction mechanism.
Gao and Yao~\etal~\cite{crcg2024aaai} constructed a synthetic dataset with controllable SCMs for graph-level research. They assumed the existence of causal, confounder and noise factors and generated the graph samples by designing the causal relationships among them.


\subsection{Evaluating Graph Fairness}
The datasets used for graph fairness research are generated to include examples of potential bias and unfairness, such as under-represented groups or imbalanced classes, and require additional considerations beyond traditional graph learning benchmarks. 
Dong~\etal~\cite{fairness4gnn2022survey} summarized the benchmark graph fairness datasets and categorized them into social networks, recommendation-based networks, academic networks, and other types of networks. 
Qian and Guo~\etal~\cite{fairgraphbench2024} curated a collection of synthetic, semi-synthetic, and real-world datasets. They are tailored to consider graph structure utility and bias information, offering flexibility to create data with controllable biases for comprehensive evaluation.

\eat{
Different datasets might support studies on varied fairness notions.
Here we exclusively present datasets related to the evaluation of graph counterfactual fairness in \tabref{tab:fairness_data}.
\begin{table}[h]
    \centering
    \begin{tabular}{c|c}
         &  \\
         & 
    \end{tabular}
    \caption{Widely-used benchmark graph datasets for graph counterfactual fairness research.}
    \label{tab:fairness_data}
\end{table}
}

\subsection{Datasets for Evaluating Graph Explanations}
To evaluate factual and counterfactual graph explainers,
it is important to collect diverse datasets that vary in terms of size, type, structure, and application scenarios.
Moreover, since human interpretation is indispensable for assessing the quality of generated explanations, 
it is preferable that the graph dataset satisfies two criteria, (i) it should be human-understandable and easy to visualize, and (ii) the graph rationales~\cite{DIR2022iclr} are identifiable with expert knowledge, which serve as a valuable approximation to ground-truth explanations and facilitate quantitative evaluation of the explainers.
A series of frequently used datasets spanning over synthetic graphs, sentiment graphs, and molecular graphs for evaluating the quality of factual graph explanations have been thoroughly analyzed in~\cite{xgnnsurvey2022tpami}. They have also been used for assessing graph counterfactual explanations~\cite{gce_survey2022}.
\eat{
We concisely summarize them in \tabref{tab:xgnn_data} and relate them with reviewed literature.
\begin{table}[ht]
    \centering
    \begin{tabular}{c|c}
         &  \\
         & 
    \end{tabular}
    \caption{Widely-used benchmark graph datasets for explainable GNN research.}
    \label{tab:xgnn_data}
\end{table}
}

\section{TGNN Evaluation Metrics Details}
\label{app:metric}
\subsection{Metrics for Evaluating Graph OOD Generalizability}
To assess the OOD generalizability, one straightforward way is to compare accuracy-related measures of the model in testing environments with varied distribution shifts. 
However, in high-stakes applications like criminal justice and financial domains, there may be a preference for assessing the overall stability or robustness of the model's performance across a range of OOD scenarios~\cite{oodmetric2021,stable2022nature,dro2018}.
Therefore, we here list several metrics for a more comprehensive evaluation of the graph OOD generalization methods on multiple test environments. 
\eat{
We use $\text{acc}_k$ to denote any accuracy-related measure of the target method in the $k$th testing environment.
\begin{definition}[Average Accuracy~\cite{stable_2018}]
\begin{align*}
    \overline{\text{Acc}} = \frac{1}{K}\sum_{k=1}^K \text{acc}_k.
\end{align*}
\end{definition}
\begin{definition}[Standard Deviation Accuracy~\cite{stable_2018}]
\begin{align*}
    \text{Acc}_{\text{std}} = \sqrt{\frac{1}{K-1}\sum_{k=1}^K (\text{acc}_k - \overline{\text{Acc}})^2}.
\end{align*}
\end{definition}
\begin{definition}[Worse Case Accuracy~\cite{dro2018}]
\begin{align*}
    \text{Acc}_{\text{worse}} = \min_{k\in[K]}\text{acc}_k.
\end{align*}
\end{definition}
}
Average Accuracy~\cite{stable_2018} measures the average performance in all testing environments. Standard Deviation Accuracy~\cite{stable_2018} measures the performance variation in all testing environments. Worse Case Accuracy~\cite{dro2018} reflects the worst possible outcome a method might produce.
The first two metrics offer a broader perspective on OOD stability, while the last one is favored in applications where extreme performances are unacceptable.


\subsection{Metrics for Evaluating Graph Fairness}
Various metrics have been proposed to evaluate GNNs \wrt{different correlation-based fairness notions}~\cite{fairness4gnn2022survey}.
For instance, statistical parity~\cite{sp_metric2012itcs} measures the disparity in model predictions for populations with different sensitive attributes. Equal opportunity~\cite{eo_metric2016nips} measures such disparity solely within populations with positive labels.
As counterfactual fairness is conceived as a more comprehensive notion than correlation-based fairness notions~\cite{cf2017nips}, 
it is necessary to evaluate GNNs optimized for GCF on these correlation-based fairness metrics~\cite{nifty2021uai,mccnifty2021icdm,gear2022wsdm}.


Furthermore, the GCF notion induces causality-based metrics which complement correlation-based fairness metrics. 
Agarwal~\etal~\cite{nifty2021uai} proposed Unfairness Score, which is defined as the percentage of nodes whose predicted label changes when the sensitive attribute of the node is altered.
Ma~\etal~\cite{gear2022wsdm} proposed a GCF metric serving as a practical counterpart of the GCF notion in \defref{def:gcf}, 
\begin{align}
    \delta_{\text{GCF}} = 
    |P(\hat{Y}_u |do(\mathbf{s}'), \mathbf{X}, \mathbf{A}) - P(\hat{Y}_u|do(\mathbf{s}''), \mathbf{X}, \mathbf{A})|,
\end{align}
where $\mathbf{s}', \mathbf{s}'' \in \{0, 1\}^n$ denote arbitrary values of the sensitive attributes for all nodes.
This metric measures the discrepancy between the interventional distributions of model predictions rather than node representations. Besides, it considers the influences of both a node's and its neighbors' sensitive attributes on model fairness.


\subsection{Metrics for Evaluating Graph Explainability}
\eat{
}
As graph explanations are generated to explain the model behavior, several evaluation metrics have been proposed from the model's standpoint~\cite{xgnnsurvey2022tpami}.

Fidelity~\cite{moexplanation2021icdm,orphicx2022cvpr} measures the change in model prediction when masking the explanation from the original graph.  
Sparsity~\cite{moexplanation2021icdm, orphicx2022cvpr} quantifies the extent to which the explanation disregards insignificant graph components. Stability~\cite{moexplanation2021icdm} is adopted to assess the robustness of an explainer by comparing the generated explanations before and after perturbing the input graph. Contrastivity~\cite{reinforced_explainer2023tpami} measures the differences between explanations for graphs from different classes.
Probability of Sufficiency~\cite{cff2022www} measures the percentage of input nodes or graphs whose explanations are sufficient to maintain the same model predictions.
In graph datasets with identifiable invariant components $I$, such as synthetic or molecule graphs, $I$ can reasonably approximate the ground truth explanation for a well-trained GNN. Accuracy~\cite{gem2021icml,reinforced_explainer2023tpami} is then employed to measure the distance between generated explanations and $I.$
These metrics are applicable to both factual and counterfactual graph explanation methods~\cite{cfgnnexplainer2022aistat,rcexplainer2021nips}

Uniquely, as maintaining similarity with the original input graph is crucial in generating GCEs, it is imperative to employ appropriate metrics for the evaluation of GCEs from this perspective.
Existing similarity/distance measures such as Graph Edit Distance~\cite{obsdbs2021kdd} and Tanimoto Similarity~\cite{maccs2022chemistry} have been adopted.
Besides, Lucic~\etal~\cite{meg2021ijcnn} proposed a customized MEG Similarity which is calculated as a convex combination of Tanimoto similarity and cosine similarity.  
Liu~\etal~\cite{moexplanation2021icdm} defined Counterfactual Relevance to measure the difference between the faithfulness of factual and counterfactual explanations.
Tan~\etal~\cite{cff2022www} proposed Probability of Necessity to quantify the percentage of input nodes or graphs, removing whose explanations can result in a change in the model prediction.
In addition, all the metrics mentioned above can be generalized to situations where multiple GCEs are generated for each input graph by averaging the metric over all GCEs~\cite{clear2022nips}.
\eat{
}
Furthermore, there are metrics proposed to evaluate other perspectives of the GCE beyond its vanilla definition.
Ma~\etal~\cite{clear2022nips} designed a Causality Ratio to measure the proportion of GCEs generated for an input graph that satisfies the domain-specific causality constraints.
Huang and Kosan~\etal~\cite{GCFExplainer2023wsdm} proposed Coverage, Cost and Interpretability, adapting the instance-level GCEs metrics for model-level GCEs.

\end{document}